\documentclass[twoside]{IEEEtran}
\usepackage{graphicx}
\usepackage{booktabs} % for professional tables
\usepackage{verbatim} % for using the {comment} environment
\usepackage{ifthen} % manage which version to show in LaTeX
\newboolean{short_version}
\newboolean{arxiv_version}
\setboolean{arxiv_version}{true} % true for arxiv version 
\setboolean{short_version}{false} % false for full version

\usepackage{hyperref}

\usepackage{balance}

\usepackage{subfiles}
\usepackage{wrapfig} % to wrap text around figures

\usepackage{dcolumn}        % columns alignment on a 'decimal point'
\usepackage{nicefrac}       % nice inline fractions

\usepackage{mathrsfs}
\usepackage{amsmath,amsthm,amssymb}
\usepackage{bm} % bold math symbols, e.g. Greek letters

\usepackage{multirow} % multirow command

\PassOptionsToPackage{dvipsnames}{xcolor} % more colour names, should be loaded before TikZ

\usepackage[caption=false,font=footnotesize]{subfig}
\usepackage{tikz}
\usetikzlibrary{shapes,arrows}
\usepackage{pgfplots}
\usepgfplotslibrary{fillbetween}
\usepackage[draft]{tikzpeople} % figures of people
\usetikzlibrary{calc} % calculations in TikZ
\usetikzlibrary{positioning}
\usetikzlibrary{arrows}
\usetikzlibrary{shapes.multipart}
\usepackage{url}

\usepackage[capitalize]{cleveref} % "clever" references with \cref{} (no "Theorem~" needed)

\newtheorem{lemma}{Lemma}
\newtheorem{theorem}{Theorem}
\newtheorem{definition}{Definition}

\allowdisplaybreaks

\newcommand{\Reals}{\mathbb{R}}

\newcommand{\tn}[1]{\textnormal{#1}}
\newcommand{\vect}[1]{\bm{#1}}
\newcommand{\inv}[1]{#1^{-1}} % inverse  
\newcommand{\const}[1]{\textnormal{\usefont{U}{eur}{m}{n}\selectfont #1}} % Euler font for special constants
\newcommand{\set}[1]{\mathcal{#1}} % set
\newcommand{\code}[1]{\mathcal{#1}} % the code notation
\newcommand{\collect}[1]{\mathscr{#1}} % the collection notation
\newcommand{\Gaussian}[2]{\mathcal{N} (#1, #2)} % the Gaussian distribution
\newcommand{\eqdef}{\triangleq} % definition

\newcommand{\trans}[1]{#1^{\textup{\textsf{\tiny T}}}} % transpose  

\newcommand{\card}[1]{\left|#1\right|}       % cardinality of a set.

\DeclareMathOperator*{\argmax}{arg\,max}  % limits under argmax
\DeclareMathOperator*{\argmin}{arg\,min}

\newcommand{\Exp}{\operatorname{\textnormal{\textsf{E}}}}
\newcommand{\E}[2][]{\Exp_{#1}\left[#2\right]}

\newcommand{\bigE}[2][]{\Exp_{#1}\bigl[#2\bigr]}
\newcommand{\BigE}[2][]{\Exp_{#1}\Bigl[#2\Bigr]}

\newcommand{\Econd}[3][]{\Exp_{#1}\left[#2 \kern0.1em\middle|\kern0.1em #3\right]}
\newcommand{\eEcond}[3][]{\Exp_{#1}[#2 \kern0.1em|\kern0.1em #3]}
\newcommand{\bigEcond}[3][]{\Exp_{#1}\bigl[#2 \kern-0.1em \bigm| \kern-0.1em #3\bigr]}
\newcommand{\BigEcond}[3][]{\Exp_{#1}\Bigl[#2 \kern-0.1em \Bigm| \kern-0.1em #3\Bigr]}
\newcommand{\biggEcond}[3][]{\Exp_{#1}\biggl[#2 \kern-0.1em \biggm| \kern-0.1em #3\biggr]}
\newcommand{\BiggEcond}[3][]{\Exp_{#1}\Biggl[#2 \kern-0.1em \Biggm| \kern-0.1em #3\Biggr]}

\newcommand{\Var}[2][]{\mathop{}\!\mathsf{Var}_{#1}\left[#2\right]}

\newcommand{\HH}{\mathop{}\!\mathsf{H}} % entropy
\newcommand{\HP}[1]{\HH\left(#1\right)} 
\newcommand{\eHP}[1]{\HH(#1)} 
\newcommand{\bigHP}[1]{\HH\bigl(#1\bigr)}

\newcommand{\HPcond}[2]{\HH\left(#1 \kern0.1em\middle|\kern0.1em #2\right)}
\newcommand{\eHPcond}[2]{\HH(#1 \kern0.1em|\kern0.1em #2)} 
\newcommand{\bigHPcond}[2]{\HH\bigl(#1 \kern-0.1em \bigm| \kern-0.1em#2\bigr)}
\newcommand{\BigHPcond}[2]{\HH\Bigl(#1 \kern-0.1em \Bigm| \kern-0.1em#2\Bigr)}
\newcommand{\CE}[2]{\HH\left(#1 \kern0.1em\middle|\kern0.1em #2\right)} 
\newcommand{\eCE}[2]{\HH(#1 \kern0.1em\|\kern0.1em #2)} 
\newcommand{\bigCE}[2]{\HH\bigl(#1 \kern0.1em\bigm\|\kern0.1em #2\bigr)}
\newcommand{\BigCE}[2]{\HH\Bigl(#1 \kern0.1em\BigM\|\kern0.1em #2\Bigr)}
\newcommand{\II}{\mathop{}\!\mathsf{I}}  % mutual information
\newcommand{\MI}[2]{\II\left(#1 \kern0.1em{;}\kern0.1em #2\right)} 
\newcommand{\eMI}[2]{\II(#1 \kern0.1em{;}\kern0.1em #2)} 
\newcommand{\bigMI}[2]{\II\bigl(#1 \kern0.1em{;}\kern0.1em #2\bigr)}
\newcommand{\BigMI}[2]{\II\Bigl(#1 \kern0.1em{;}\kern0.1em #2\Bigr)}
\newcommand{\MIcond}[3]{\II\left(#1 \kern0.1em{;}\kern0.1em #2 \kern0.1em\middle|\kern0.1em #3\right)}
\newcommand{\eMIcond}[3]{\II(#1 \kern0.1em{;}\kern0.1em #2 \kern0.1em|\kern0.1em #3)} 
\newcommand{\bigMIcond}[3]{\II\bigl(#1 \kern0.1em{;}\kern0.1em #2 \kern-0.1em \bigm| \kern-0.1em#3\bigr)}
\newcommand{\BigMIcond}[3]{\II\Bigl(#1 \kern0.1em{;}\kern0.1em #2 \kern-0.1em \Bigm| \kern-0.1em#3\Bigr)}
\newcommand{\eUniform}[1]{\tn{Uniform}(#1)}

\usepackage{todonotes}
\makeatletter
\tikzset{
	database/.style={
		path picture={
			\draw (0, 1.5*\database@segmentheight) circle [x radius=\database@radius,y radius=\database@aspectratio*\database@radius];
			\draw (-\database@radius, 0.5*\database@segmentheight) arc [start angle=180,end angle=360,x radius=\database@radius, y radius=\database@aspectratio*\database@radius];
			\draw (-\database@radius,-0.5*\database@segmentheight) arc [start angle=180,end angle=360,x radius=\database@radius, y radius=\database@aspectratio*\database@radius];
			\draw (-\database@radius,1.5*\database@segmentheight) -- ++(0,-3*\database@segmentheight) arc [start angle=180,end angle=360,x radius=\database@radius, y radius=\database@aspectratio*\database@radius] -- ++(0,3*\database@segmentheight);
		},
		minimum width=2*\database@radius + \pgflinewidth,
		minimum height=3*\database@segmentheight + 2*\database@aspectratio*\database@radius + \pgflinewidth,
	},
	database segment height/.store in=\database@segmentheight,
	database radius/.store in=\database@radius,
	database aspect ratio/.store in=\database@aspectratio,
	database segment height=0.1cm,
	database radius=0.25cm,
	database aspect ratio=0.35,
}
\makeatother

\usepackage[noadjust]{cite}       % Improves presentation of citations, like [3--7].
\ifCLASSINFOpdf
\else
\fi
\usepackage{algorithm}
\usepackage{algorithmic}

\definecolor{darkgreen}{rgb}{0, 0.5, 0}
\usepackage[normalem]{ulem} % for strikethrough text

\begin{document}
% \title{Bare Demo of IEEEtran.cls\\ for IEEE Journals}
\title{Generative Adversarial User Privacy in \\
  Lossy Single-Server Information Retrieval}

\author{% Michael~Shell,~\IEEEmembership{Member,~IEEE,}
%         John~Doe,~\IEEEmembership{Fellow,~OSA,}
%         and~Jane~Doe,~\IEEEmembership{Life~Fellow,~IEEE}% <-this % stops a space
% \thanks{M. Shell was with the Department
% of Electrical and Computer Engineering, Georgia Institute of Technology, Atlanta,
% GA, 30332 USA e-mail: (see http://www.michaelshell.org/contact.html).}% <-this % stops a space
% \thanks{J. Doe and J. Doe are with Anonymous University.}% <-this % stops a space
% \thanks{Manuscript received April 19, 2005; revised August 26, 2015.}
  Chung-Wei~Weng,~% \IEEEmembership{Student Member,~IEEE,}
  Yauhen~Yakimenka,
  Hsuan-Yin~Lin,~\IEEEmembership{Senior~Member,~IEEE},
  Eirik~Rosnes,~\IEEEmembership{Senior~Member,~IEEE},
  and~J{\"o}rg~Kliewer,~\IEEEmembership{Senior~Member,~IEEE}% <-this % stops a space
  \thanks{This work was supported in part by US NSF under Grant CNS-1813942. Also, the research presented in this article has benefited from the Experimental Infrastructure for Exploration of Exascale Computing (eX3), which is financially supported by the Research Council of Norway under contract 270053 (\url{https://www.ex3.simula.no/}). This paper was presented in part at the NeurIPS Workshop on Privacy Preserving Machine Learning - PRIML and PPML Joint Edition, Vancouver, BC, Canada, Dec. 11, 2020.}
  \thanks{C.-W.~Weng, Y.~Yakimenka, H.-Y.~Lin, and E.~Rosnes are with Simula UiB, N-5006 Bergen, Norway (e-mail: chungwei@simula.no, yauhen@simula.no, lin@simula.no, eirikrosnes@simula.no).}% <-this % stops a space
  \thanks{J.~Kliewer is with Helen and John C. Hartmann Department of Electrical and Computer Engineering, New Jersey Institute of Technology, Newark, New Jersey 07102, USA (e-mail: jkliewer@njit.edu).}}

\maketitle

\begin{abstract}
  % The abstract goes here.
  We propose to extend the concept of private information retrieval by allowing for distortion in the retrieval process and relaxing the perfect privacy requirement at the same time. In particular, we study the trade-off between download rate, distortion, and user privacy leakage, and show that in the limit of large file sizes this  trade-off can be captured via a novel information-theoretical formulation for datasets with a known distribution. Moreover, for scenarios where the statistics of the dataset is unknown, we propose a new deep learning framework by leveraging a generative adversarial network approach, which allows the user to learn efficient schemes from the data itself. %, minimizing the download cost. 
  We evaluate the performance of the scheme on a synthetic Gaussian dataset as well as on the MNIST, CIFAR-$10$,  and LSUN datasets. For the MNIST,  CIFAR-$10$,  and LSUN datasets, the data-driven approach significantly outperforms a nonlearning-based scheme which combines source coding with the download of multiple files.
%\jj{ \sout{For the MNIST dataset, the data-driven approach significantly outperforms a non-learning based scheme which combines source coding with multiple file download, while the CIFAR-$10$ and LSUN performance is notably better.}}
\end{abstract}


% \fi
\IEEEpeerreviewmaketitle

\section{Introduction}
\label{sec:introduction}
% for IEEE journal papers produced under \LaTeX\ using
% IEEEtran.cls version 1.8b and later.
% % You must have at least 2 lines in the paragraph with the drop letter
% % (should never be an issue)
% I wish you the best of success.

 

\IEEEPARstart{M}{achine} learning (ML) has been recognized as a game-changer in modern information technology, and various ML techniques are increasingly being utilized for a variety of applications from intrusion detection to image classification. % and to recommending new movies.
Efficient information retrieval (IR) from a single or several servers storing such datasets under a strict user privacy constraint has been extensively studied within the framework of private information retrieval (PIR). In PIR, first introduced by Chor \emph{et al.} \cite{ChorGoldreichKushilevitzSudan95_1}, a user can retrieve an arbitrary file from a dataset without disclosing any information (in an information-theoretical sense) about which file she is interested in to the servers storing the dataset. Typically, the size of the queries is much smaller than the size of a file. Hence, the efficiency of a PIR protocol is usually measured in terms of the download cost, or equivalently, the download (or PIR) rate, neglecting the upload cost of the queries. PIR has been studied extensively over the last decade, see, e.g., \cite{BanawanUlukus18_1,Freij-HollantiGnilkeHollantiKarpuk17_1,KoppartySarafYekhanin11_1,SunJafar17_1,TajeddineGnilkeElRouayheb18_1,Yekhanin10_1} and references therein. 

Recently, there has been several works proposing to relax the perfect privacy condition of PIR in order to improve on the download cost, see, e.g.,~\ifthenelse{\boolean{arxiv_version}}{\cite{LinKumarRosnesGraellAmatYaakobi22_1,LinKumarRosnesGraellAmatYaakobi21_1,SamyAttiaTandonLazos21_1,ZhouGuoTian20_1,QianZhouTianLiu22_1,ToledoDanezisGoldberg16_1}}{\cite{LinKumarRosnesGraellAmatYaakobi22_1,LinKumarRosnesGraellAmatYaakobi21_1,SamyAttiaTandonLazos21_1,ToledoDanezisGoldberg16_1}}. Inspired by this line of research, we propose to simultaneously relax both the perfect privacy condition and the perfect recovery condition, by allowing for some level of distortion in the recovery process of the requested file, in order to achieve even lower download costs (or, equivalently, lower download rates). This scenario is of interest in many real-world applications. For instance, a user may be willing to share the genre of a requested movie to the servers storing it (but not the exact identity of the movie) and may also wish to retrieve the movie under a small level of distortion as long as the retrieved quality is high enough, in order to reduce the required download time.

We concentrate on the practical scenario in which the dataset is stored on a single server, which is in alignment with the current research trend within the PIR literature as it is very difficult to ensure that the servers cannot collude in practice. For instance, typically all the servers are owned by one entity, e.g., Google, and then we cannot assume that they are independent. 

% \begin{wrapfigure}[15]{r}{.4\textwidth}
\begin{figure}[t!]
  \centering
  % \begin{center}
  \scalebox{0.775}{\input{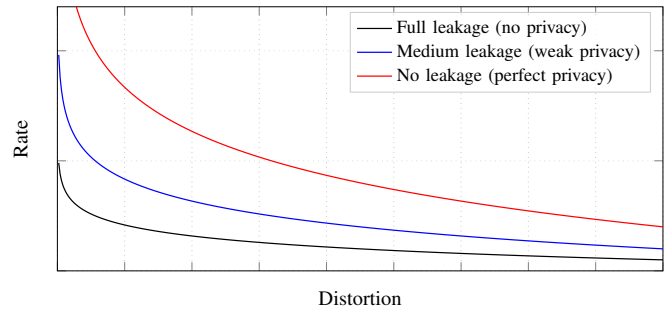}}
  \vspace*{-7mm}
  % \end{center}
  \caption{The rate-distortion trade-off under different privacy levels.}
  \label{fig:RvsD_showcase}
\end{figure}
% \end{wrapfigure}

Our main contributions are summarized as follows.
\begin{itemize}
\item %We concentrate on the practical scenario in which the dataset is stored on a single server, which is in alignment with the current research trend within the PIR literature as it is very difficult to ensure that the servers cannot collude in practice. For instance, typically all the servers are owned by one entity, e.g., Google, and then we cannot assume that they are independent. 
A problem formulation with arbitrary distortion and leakage functions is presented, which establishes a trifold trade-off between download rate, privacy leakage to the server storing the dataset, and distortion in the recovery process for the user.
\item We show  that the optimal rate-distortion-leakage trade-off is convex and this allows us to further show a concise information-theoretical formulation in terms of mutual information (MI) in the limit of large file sizes (see~\cref{thm:RDL-function}).
% \item In the special case of full leakage to the server, the proposed formulation yields the well-known rate-distortion curve.
  The typical behavior of a rate-distortion-leakage trade-off is illustrated in Fig.~\ref{fig:RvsD_showcase}, showing that an increased level of privacy leads to a higher rate-distortion trade-off curve, and hence a sacrifice in either download rate or distortion. Furthermore, a general achievable scheme combining source coding with the download of multiple files is proposed. % for datasets with a known distribution.
\item To overcome the practical limitation of unknown statistical properties of real-world datasets, we consider a data-driven approach leveraging recent advancements in generative adversarial networks (GANs)~\cite{Goodfellow-etal14_1}, which allows a user to learn efficient schemes % (in terms of download rate)
 from the data itself. In our proposed novel GAN-based IR framework, learning can be phrased as a constrained minimax game between a user which desires to keep the identity of the requested file private and a server that tries to infer which file the user is interested in, under both a user distortion and a download rate constraint. \ifthenelse{\boolean{arxiv_version}}{Similar to~\cite{Springenberg16_1}, where a cross-entropy loss function is used as a discriminative classifier for unlabeled or partially labeled data, the server is modeled as a discriminator %in the generalized GAN framework, 
and trained with cross-entropy for labeled data.}{The server is modeled as a discriminator and trained with cross-entropy for labeled data.} 
The data flow can be seen as similar to that of auxiliary classifier GANs~\cite{OdenaOlahShlens17_1}, but with important differences in terms of, e.g., the loss function.
%\item In our proposed novel GAN-based IR framework, learning can be phrased as a constrained minimax game between a user which desires to keep the identity of the requested file private and a server that tries to infer which file the user is interested in, under both a user distortion and a download rate constraint. Similar to~\cite{Springenberg16_1}, where a cross-entropy loss function is used as a discriminative classifier for unlabeled or partially labeled data, the server is modeled as a discriminator in the generalized GAN framework, and also trained with cross-entropy for labeled data. We evaluate the performance of the proposed scheme on a synthetic Gaussian dataset as well as on the MNIST~\cite{LecunBottouBengioHaffner98_1},  CIFAR-$10$~\cite{Krizhevsky09_1}, {\revone and LSUN~\cite{Yu-etal15_1sub}} datasets.
\item We present numerical results for the MNIST~\cite{LecunBottouBengioHaffner98_1}, CIFAR-$10$~\cite{Krizhevsky09_1}, and LSUN~\cite{Yu-etal15_1sub} datasets (the latter comprising  $24$ times as many images as CIFAR-$10$), showing that the data-driven approach significantly outperforms the proposed achievable scheme,\footnote{The code for this work is available at \url{https://github.com/Simula-UiB/GAUP_TIFS22}.} while for a synthetic  Gaussian dataset, where the source statistics is known, it performs close to the proposed achievable scheme using a variant of the generalized Lloyd algorithm~\cite{Lloyd82_1,LindeBuzoGray80_1} for the source code. %{\jj \sout{For  CIFAR-$10$~\cite{Krizhevsky09_1}, the performance of the data-driven approach is mostly comparable to that of the proposed achievable scheme. However, when the download rate is sufficiently low, it clearly outperforms the achievable scheme. For the large-scale LSUN dataset~\cite{Yu-etal15_1sub} (comprising  $24$ times as many images as CIFAR-$10$) we observe a similar gain compared to the compression-based scheme as for CIFAR-$10$.}}
\end{itemize}  
% Moreover,

% \subsection{Organization of Paper}
% \label{sec:organization-paper}

% The remainder of this paper is structured as follows. Section~\ref{sec:preliminaries-SystemModel} presents the notation, the establishment of the rate-distortion-leakage trade-off for single-server IR, and the corresponding generative adversarial formulation. An optimization formulation for the optimal rate-distortion-leakage trade-off in terms of mutual information is also provided in the limit of large file sizes (referred to as Shannon's scheme). In Section~\ref{sec:theoretical_approach}, we present a general achievable scheme when the data statistics is known. The proposed data-driven framework for IR is elaborated further in Section~\ref{sec:data_driven_approach}. In Section~\ref{sec_numerical_results}, the results of the data-driven approach for a synthetic Gaussian dataset, as well as the MNIST and CIFAR-$10$ datasets are presented. The learning performance are also compared to that of the compression-based schemes for all datasets and Shannon's scheme for Gaussian dataset. Finally, Section~\ref{sec:conclusion} concludes the paper. 
% 

\subsection{Related Work}
\label{sec:related-work}

As outlined above, in this work we consider ``information retrieval'' in the sense of PIR, while ``information retrieval'' in the traditional sense used by the \emph{information retrieval community} has a different meaning. In particular, in the traditional sense ``information retrieval'' refers to the problem of providing a list of documents given a query and has a wide range of applications \cite{Baeza-YatesRibeiro-Neto99_1}. In \cite{WangYu-etal17_1}, the authors proposed to iteratively optimize two well-established models of traditional IR; namely generative retrieval focusing on predicting relevant documents given a query and discriminative retrieval focusing on predicting document relevance given a query and document pair. The resulting optimization problem is formulated as a minimax game. Due to the differences in the system model there is no clear connection to our proposed framework, besides the formulation as a minimax game.

Preserving the privacy of sensitive information from publicly released datasets, while providing useful utility is a well-studied problem in the computer science literature. %The notion of privacy in this line of work is different from what is considered in our work, where the goal is to 
In contrast, the goal in this work is to preserve the privacy of requests. In \cite{KiferMachanavajjhala11_1}, differential privacy \cite{DworkMcSherryNissimSmith06_1,Dwork06_1} is used as privacy metric and a fundamental trade-off between privacy and utility for statistical databases is presented. A similar privacy-utility trade-off is also investigated in~\cite{SeifTandonLi19_1}, but for another privacy measure and under an assumption that some prior knowledge about the data is known. Recently, this kind of problem is also studied for synthetic datasets in the ML literature~\cite{StadlerOprisanuTroncoso22_1}, and it has been shown that it is related to rate-distortion theory and has a strong connection to the problem of robust ML~\cite{WangAeronRakinKoike-AkinoMoulin21_1}. Adversarial training is also applied for learning anonymized representations of a dataset, while protecting the private labels from an intermediate representation in~\cite{FeutryPiantanidaBengioDuhamel18_1sub}. In~\cite{Wu-etal18_1}, adversarial training for preserving the privacy of visual recognition is adopted. An adversarial training framework that can simultaneously prevent leakage of private attributes and withstand reconstruction attacks was proposed in~\cite{LiGuoYangSalimChen21_1}. Lastly, we remark here that in our work, in contrast to the literature on adversarial training for privacy, the communication efficiency (in terms of download rate) of an IR scheme is considered an important performance metric. 

Similar data-driven approaches, under the names of generative adversarial privacy~\cite{HuangKairouzChenSankarRajagopal17_1,HuangKairouzChenSankarRajagopal18_1}, privacy-preserving adversarial networks~\cite{TripathyWangIshwar19_1}, and compressive privacy GAN~\cite{TsengWu20_1}, have recently been proposed for learning a privatization mechanism directly from the dataset in order to release it to the public and for generating compressed representations that retain utility while being able to withstand reconstruction attacks. A similar approach was also taken in~\cite{BlauMichaeli19_1} where a trifold trade-off between rate, distortion, and perception in lossy image compression was established.

Relaxing the perfect information-theoretical privacy condition by considering computationally-private IR, where the privacy  relies on an intractability assumption (e.g., the hardness of deciding quadratic residuosity), has been investigated in
several previous works, see, e.g.,~\cite{KushilevitzOstrovsky97_1,KushilevitzOstrovsky00_1,LipmaaPavlyk17_1}. Hence, given infinite computational power, the requested file index can be determined precisely. Moreover, in~\cite{KadheGarciaHeidarzadehElRouayhebSprintson20_1}, it was shown that allowing for side information can also decrease the download cost in single-server PIR. In~\cite{SamyAttiaTandonLazos20_1}, instead of keeping the identity of the requested file private, the authors propose to preserve the privacy of latent attributes  of the user that are dependent on the retrieving requests.
%file induced by the retrieving requests.
In contrast to these previous works, here we propose to relax the perfect reconstruction constraint in order to decrease the download cost. Moreover, to the best of our knowledge, extending PIR to \emph{both} nonperfect privacy and recovery, specifically in a learning-based context by employing generative adversarial models, has not been addressed in the open literature so far.

% \subsection{Subsection Heading Here}
% Subsection text here.

%\IEEEpubidadjcol

% \subsubsection{Subsubsection Heading Here}
% Subsubsection text here.

\section{Preliminaries and System Model}
\label{sec:preliminaries-SystemModel}      

\subsection{Notation}
\label{sec:notation}

We define $[a]\eqdef\{1,2,\ldots,a\}$. 
%by $\mathbb{N}$ the set of all positive integers and  $[a]\eqdef\{1,2,\ldots,a\}$. %  for $a\in\{0\}\cup\mathbb{N}$  %and $[a:b]\eqdef\{a,a+1,\ldots,b\}$ for $a,b\in\{0\}\cup\mathbb{N}$ and $a \leq b$. 
An arbitrary field is denoted by $\mathbb{F}$, while the set of nonnegative real numbers is denoted by $\Reals_{\ge 0}$. Vectors are denoted by bold letters and sets by calligraphic uppercase letters, e.g., $\vect{x}$ and $\set{X}$, respectively. We use uppercase letters for random variables (RVs) (either scalar or vector), e.g., $X$ or $\vect{X}$. For a given index set $\set{I}$, we write $X^{\set{I}}$ to represent $\bigl\{X^{(m)}\colon m\in\set{I}\bigr\}$. $\E[X]{\cdot}$ and $\E[P_X]{\cdot}$ denote expectation with respect to the RV $X$ and the probability mass function (PMF) $P_X$, respectively. %, and the notation $X \sim P_X$ signifies that $X$ is distributed according to $P_X$. 
$\HP{X}$ or $\HP{P_X}$ represents the entropy of $X$, while  %, where $P_{X}(\cdot)$ is the probability mass function (probability vector) of the RV $X$. 
$\eMI{X}{Y}$ denotes the MI between $X$ and $Y$. % The cross-entropy between the distributions $P_X$ and $P_Y$ is denoted by $\bigCE{P_X}{P_Y}$. 
The multivariate Gaussian distribution with mean $\vect \mu$ and covariance matrix $\Sigma$ is denoted as $\mathcal{N}(\vect \mu,\Sigma)$. In particular, if the entries corresponding to this distribution are mutually independent, we have $\Sigma = \sigma^2 I$, for marginal standard deviation $\sigma \in \Reals_{\ge 0}$, where $I$ denotes the identity matrix.
%$\One{\cdot}$ denotes the indicator function, i.e., $\One{\text{statement}}$ equals to $1$ if the statement holds, and $0$ otherwise.
The transpose of a vector is denoted as $\trans{(\cdot)}$, while the gradient of a function $f(x)$ is denoted by $\nabla f(x)$.

\subsection{Single-Server Information Retrieval}
Consider a dataset containing $\const{M}$ files $\vect{X}^{(1)}, \dotsc, \vect{X}^{(\const{M})}$ stored on a single server, where each file
$\vect{X}^{(m)}=\trans{\bigl(X_1^{(m)},\ldots,X_\beta^{(m)}\bigr)}$, $m\in [\const{M}]$, can be seen as a $\beta\times 1$ random vector (according to some probability distribution $P_{\vect{X}^{(m)}}$) over $\mathbb{F}^\beta$, where $\mathbb{F}$ is any field (finite or infinite).\footnote{With some abuse of wording, \emph{dataset} refers to both the set of files $\vect{X}^{(1)}, \dotsc, \vect{X}^{(\const{M})}$ and the training samples for the data-driven approach.}   %Define $\vmtrx{X}^{[\const{M}]} = \{ \vect{X}^{(m)}: m \in [\const{M}] \}$.
Assume that the user wishes to retrieve the $M$-th file, $M\in[\const{M}]$, where, for simplicity, $M$ is assumed to be uniformly distributed over $[\const{M}]$.\footnote{Throughout the paper, we assume for simplicity that $M$ is uniformly distributed and also that the file sizes are  equal and fixed. The distribution $P_{M}$ does not affect the generality of our results, and it is also common to have equal and fixed file sizes in the PIR literature. These assumptions can be lifted, which is referred to as semantic PIR in the literature \cite{VithanaBanawanUlukus22_1}.} 
%æ
%\footnote{The assumption that $M$ is uniformly distributed can be lifted.} 
%
%
%
%
The formal definition of a general single-server IR scheme is as follows.
\begin{definition}
  \label{Def:Mn-IRscheme}
  An IR scheme $\collect{C}$ for a single server  storing $\const{M}$ files consists of:
  % \begin{itemize}
  % \setlength\itemsep{-0.5em}
  % \item 
  (i) a random strategy $\vect{S}$. % whose alphabet is $\set{S}$,
  % \item
  (ii) A deterministic query function $f_{\textnormal{Q}}$ that generates a query $\vect{Q}=f_{\textnormal{Q}}(M,\vect{S})$, for a requested file index $M$, where query $\vect{Q}$ is sent to the server.
  % \begin{displaymath}
  %   f_{\textnormal{Q}}\colon\{1,\ldots,\const{M}\}\times\set{S}\to\set{Q},
  % \end{displaymath}
  % that generates a query $\vect{Q}=f_{\textnormal{Q}}(M,\vect{S})$ with alphabet $\set{Q}$, where query $\vect{Q}$ is sent to the database,
  % \item
  (iii) A deterministic answer function $f_{\textnormal{A}}$
    % \begin{displaymath}
    %   f_{\textnormal{A}}\colon\set{Q}\times\underbrace{\set{X}^{\beta}\times\cdots\times\set{X}^{\beta}}_{\#=\const{M}}\to\set{A}^L,
    % \end{displaymath}
  that returns the answer $\vect{A}=f_{\textnormal{A}}(\vect{Q},\vect{X}^{[\const{M}]})$ back to the user. 
  % \item
  (iv) A deterministic reconstruction function $\hat{\vect{X}}\eqdef f_{\textnormal{\^{X}}}(\vect{A},M,\vect Q)$ giving an estimate of the desired file using the answer from the server together with the requested file index $M$ and the query $\vect Q$. 
    % \begin{displaymath}
    %   \psi\colon\set{A}^L\times\{1,\ldots,\const{M}\}\times\set{S}\to\hat{\set{X}}^{\beta}.
    % \end{displaymath}
    % Thus, $\hat{\vect{X}}^{(M)}\eqdef\psi(\vect{A},M,\vect S)$ is the retrieved file.
  % \end{itemize}  
\end{definition}
%It might seem that using $\vect S$ for the reconstruction of
%$\hat{\vect X}$ could give the user %additional power. However, the
%server does not use $\vect S$ directly to produce $\vect
%A$. %Therefore, using $\vect Q$ is sufficient.
Note that the server does not use $\vect S$ directly to produce $\vect
A$  and thus using $\vect Q$ is sufficient in the reconstruction.

We are interested in designing an IR scheme such that both the user's \emph{utility} and \emph{privacy} are preserved. On the one hand, this scheme should satisfy the condition of retrievability with a distortion measure $d(\cdot,\cdot)$, i.e.,
\begin{IEEEeqnarray}{c}
  \E[M,\vect{Q},\vect{X}^{[\const{M}]}]{d\bigl(\vect{X}^{(M)},\hat{\vect{X}}\bigr)}\leq\const{D},
  \label{eq:def_distortion}
\end{IEEEeqnarray}
where $\const{D}$ is a given distortion constraint and $d\bigl(\vect{X}^{(M)},\hat{\vect{X}}\bigr) = \nicefrac{1}{\beta} \sum_{i=1}^{\beta} d_i\bigl(X^{(M)}_i,\hat{X}_i\bigr)$, where $d_i: \mathbb{F} \times \mathbb{F} \rightarrow  \Reals_{\ge 0}$ is a per-symbol distortion measure that is \emph{translation invariant}, i.e., $d_i(x+z,\hat{x}+z)=d_i(x,\hat{x})$. For simplicity, we let $d_1=\cdots=d_\beta=d_{\textnormal{sym}}$. %\footnote{\eirik Here, we will assume that the distortion measures $d_1$ through $d_\beta$ are the same, and we denote them all, with some abuse of notation, by $d$.}}  
On the other hand, the user would like to preserve her privacy with the query function, in the sense that the server should not be able to fully determine  the identity $M$ of the requested file. The server receives the query $\vect{Q}$, and the leakage is measured in terms of a leakage metric $\rho(P_{\vect{Q}|M})$. The query function  should be designed such that
\begin{IEEEeqnarray}{c}
  \rho(P_{\vect{Q}|M})\leq\const{L},\label{eq:def_leakage}
\end{IEEEeqnarray}
where $\const{L}$ is the maximum allowed leakage.\footnote{Note that the leakage metric is a function of the probability distribution of $\vect Q|M$ and not the particular values. In other words, re-labeling queries in some bijective way does not change the leakage. For example, if instead of sending integers $1$, $2$, \dots, the user starts sending strings ``one'', ``two'', \dots, the leakage does not change.}
  
Moreover, it is worth mentioning that, unlike the setting of classical PIR, where perfect retrievability is ensured for every file, i.e., $\bigE[\vect{Q},\vect{X}^{[\const{M}]}]{d\bigl(\vect{X}^{(m)},\hat{\vect{X}}\bigr)}=0$ for all $m \in [\const{M}]$, the distortions for different files need not to be equal. In other words, it is possible to have $\bigE[\vect{Q},\vect{X}^{[\const{M}]}]{d\bigl(\vect{X}^{(m)},\hat{\vect{X}}\bigr)} \neq \bigE[\vect{Q},\vect{X}^{[\const{M}]}]{d\bigl(\vect{X}^{(m')},\hat{\vect{X}}\bigr)}$ for $m \neq m'$, and \eqref{eq:def_distortion} can be expressed as
% \begin{displaymath}
\begin{IEEEeqnarray*}{c}
  % \E[M,\vect{Q}]{d(\vect{X}^{(M)},\hat{\vect{X}})}& = &
  \E[M]{\Econd[\vect{Q},\vect{X}^{[\const{M}]}]{d\bigl(\vect{X}^{(M)},\hat{\vect{X}}\bigr)}{M=m}}
  \leq\const{D}.
\end{IEEEeqnarray*}  
% \end{displaymath}

Given a query function $f_{\textnormal{Q}}$ and an answer function $f_{\textnormal{A}}$ of an IR scheme, we measure its efficiency in terms of the download rate (in bits per symbol) defined as
\begin{IEEEeqnarray}{rCl}
  \IEEEeqnarraymulticol{3}{l}{%
    \const{R}(f_{\textnormal{Q}},f_{\textnormal{A}})}\nonumber\\*\quad%
  & \eqdef &{\frac{\eEcond[\vect{Q},\vect{A}]{\ell(\vect{A})}{\vect{Q}}}{\beta}=\frac{1}{\beta}\sum_{\vect{q}}P_{\vect{Q}}(\vect{q})\eEcond[\vect{A}]{\ell(\vect{A})}{\vect{Q}=\vect{q}}},\IEEEeqnarraynumspace\label{eq:def_rate}
\end{IEEEeqnarray}
where $\ell(\vect{A}=\vect{a})\mid \vect{Q}=\vect{q}$ is the length of the answer $\vect{a}$ for query $\vect{q}$.\footnote{Note that we define the download rate in accordance with the rate-distortion literature, i.e., as the fraction between the expected download cost and the requested file size $\beta$, while in the PIR literature the download rate is defined as the inverse fraction, i.e., as the fraction between the requested file size and the expected download cost.}
% encoded by the coding scheme.
% Further in the paper, the length $\ell_{\vect{q}}(\vect{a})$ is a constant (i.e., independent of $\vect{a}$ and $\vect{q}$) that is chosen according to the desired rate. 

% \begin{wrapfigure}[15]{r}{.68\textwidth}
\begin{figure}[t]
  \centering
  \scalebox{0.68}{
  \begin{tikzpicture}[squarednode/.style={rectangle, draw=black, very thick, minimum size=10mm}]
    \node[alice,%monitor,
    minimum size=.85cm,anchor=south] (User) at (0,0) {};
    
    \node[database,database radius=0.5cm,database segment height=0.25cm,anchor=south] (DBlabel) at (8,0) {};
    
    \node [below=.2cm of User, align=center] (UserLabel) {User};

    \draw[thick,densely dashed] (-.75,3.4) rectangle (4.55,-2.5);
        \node at (1.9,-3) {User};
    \draw[thick,densely dashed] (7,3.4) rectangle (11.15,-2.5);
        \node at (9.075,-3) {Server};

    \node[squarednode]      (fMhatblock) [above=0.9cm of DBlabel] {$f_{\textnormal{\^{M}}}$};
    
    \node[squarednode]      (fQblock) [left=3.2cm of fMhatblock] {$f_{\textnormal{Q}}$};
    
    \node[squarednode]      (fAblock) [below=1.0cm of DBlabel] {$f_{\textnormal{A}}$};
    
    \node[squarednode]      (fXhatblock) [left=3.2cm of fAblock] {$f_{\textnormal{\^{X}}}$};
    
    \node[right=1.1cm of fMhatblock,align=center] (svrInfer) {$\hat{M}$};
    
    \node[left=3cm of fXhatblock,align=center] (Mhatlabel) {$\hat{\vect X}$};
    
    \draw[->] (fMhatblock.east) -- (svrInfer.west);
	
    \node[squarednode] [left=2.75cm of fQblock, align=center] (noiseblock) {noise\\gen.};
    
    \path[->] (noiseblock.east) edge[] node[fill=white,anchor=center,pos=0.5] {$\vect S$} (fQblock.west);
    
%    \draw[->] (noiseblock.south) -- node [right=.5mm] {$\vect S$} (fXhatblock.north);		

	\draw[->] (fXhatblock.west) -- (Mhatlabel.east);
%    \path[->] (fXhatblock.west) edge[bend left] node[fill=white,anchor=center,pos=0.5] {$\hat{\bm X}$} (UserLabel.north east);
    
    \path[->] (User.east) edge[] node[fill=white,anchor=center,pos=0.5] {$M$} (fQblock.south west);
    
    \path[->] (User.south east) edge[] node[fill=white,anchor=center,pos=0.5] {$M$} (fXhatblock.north west);
    
    \path[->] (fQblock.east) edge[] node[fill=white,anchor=center,pos=0.5] {$\vect Q$}  (fMhatblock.west);
    
    \path[->] (fQblock.south east) edge[] node[fill=white,anchor=center,pos=0.5] {$\vect Q$}  (fAblock.north west);
	
    \path[->] (fAblock.west) edge[] node[fill=white,anchor=center,pos=0.5] {$\vect A$}  (fXhatblock.east);
    
    \path[->] (fQblock.south) edge[] node[fill=white,anchor=center,pos=0.5] {$\vect Q$}  (fXhatblock.north);			
    
    \node [above=1mm of DBlabel, align=center] (DBLabel) {DB};
    
    \draw[->] (DBlabel) -- (fAblock);					
    
    % \node [left=0.1cm of User.west] {$M$};
    \node [right=0.1cm of DBlabel.east] {$\vect{X}^{[\const M]}\begin{cases}
        \vect{X}^{(1)}
        \\
        \,\;\vdots
        \\
        \vect{X}^{(\const M)}
      \end{cases}$};
  \end{tikzpicture}}
  %\vspace{-2ex}
\caption{IR scheme for an arbitrary dataset stored on a single server.% Here, NG denotes a noise generator for $\vect{S}$.
}
\label{fig:DNN-model}
\end{figure}
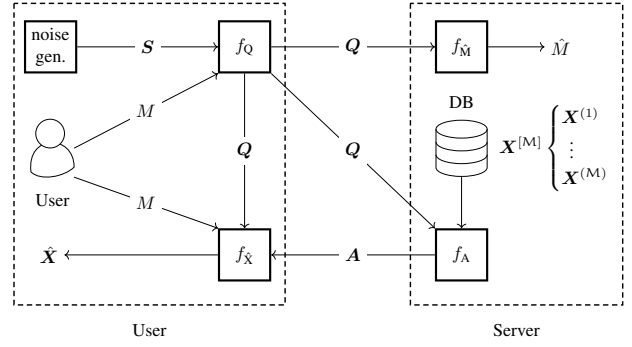
% \end{wrapfigure}

\vspace{-1ex}
%\subsection{Generative Adversarial Model for the Honest-But-Curious Database}
\subsection{Problem Formulation}
\label{sec:gen-adversarial-model}

We consider a model that protects the privacy of the user against a honest-but-curious server. The IR model above guarantees that a user can retrieve an arbitrary file stored on a single server (with some distortion), while the server can only partially infer the identity of the requested file. In this work, we will study the trifold trade-off between download rate, expected distortion, and privacy leakage to the server for an IR scheme. In particular, for a given constraint $\const{D}$ on the expected distortion $\E[M,\vect{Q},\vect{X}^{[\const{M}]}]{d\bigl(\vect{X}^{(M)},f_{\textnormal{\^{X}}}(\vect{A},M,\vect{Q})\bigr)}$ for the retrieval of the requested file indexed by $M$ and for a given constraint $\const{L}$ on the leakage to the server $\rho(P_{\vect{Q}|M})$, the goal is to minimize the download rate $\const{R}(f_{\textnormal{Q}},f_{\textnormal{A}})$. This can be formulated as the constrained optimization problem
% \begin{subequations}  
\begin{IEEEeqnarray*}{rCl}
  \min_{(f_{\textnormal{Q}},f_{\textnormal{A}},f_{\textnormal{\^{X}}})}& &\quad \const{R}(f_{\textnormal{Q}},f_{\textnormal{A}})
  \\
  \text{subject to }& &\quad
  \bigE[M,\vect{Q},\vect{X}^{[\const{M}]}]{d\bigl(\vect{X}^{(M)},\,\hat{\vect X}\bigr)}\leq\const{D},\,
  % \label{eq:distortion-cnstr}
  \\[1mm]
  & &\quad
  \rho(P_{\vect{Q}|M})\leq\const{L},\,\hat{\vect{X}} = f_{\textnormal{\^{X}}}(\vect{A},M,\vect Q),
  % \label{eq:DL-cnstr}
  \\
  &&\quad\vect Q = f_{\textnormal Q}(M, \vect S),\,\vect{A}=f_{\textnormal{A}}(\vect{Q},\vect{X}^{[\const{M}]}).
  \IEEEyesnumber\IEEEeqnarraynumspace\label{eq:optimization_RDL}
\end{IEEEeqnarray*}
% \end{subequations}
To further capture the information leakage for training, we define $\hat{M}\eqdef f_{\textnormal{\^{M}}}(\vect{Q})$ to be the server's inference of the user's requested file index $M$ from the query $\vect{Q}$ using a decision decoder $f_{\textnormal{\^{M}}}$. The proposed model is illustrated in Fig.~\ref{fig:DNN-model}.\footnote{In case the files are of different sizes, they can be dividing into different classes based on their sizes, and we would need a separate query function $f_{\textnormal{Q}}$ and decoding function $f_{\textnormal{\^X}}$ for each class,  while only a single answer function $f_{\textnormal{A}}$ and adversary function $f_{\textnormal{\^M}}$ would be required. Moreover, the distortion and leakage constraints in \eqref{eq:def_distortion} and \eqref{eq:def_leakage}, respectively,  and also the rate in \eqref{eq:def_rate} would be averages over the classes of files.} 
%
%Then, the distortion constraint in (1) can be modified to being a constraint on the average distortion over the classes of files. In a similar manner, the leakage constraint in (2) will be a constraint on the average leakage over the classes of files and also the rate in (3) would be an average rate.
%
%
%
%In order to incorporate the different classes of files, we would need to train a query generation network for each class of files and also a separate decoding network for each class (the size of the output layer for the i-th network would be βi, the size of the files in the i-th class), while only a single answer network and adversary network would be required.
%
%
The leakage on the identity of the requested file can also be quantified by a loss function denoted as $f_{\textnormal{Loss}}(M,\hat{M})$. Hence, the expected loss with respect to $M$ and $\vect{Q}$ is
% \begin{displaymath}
\begin{IEEEeqnarray*}{rCl}
  \IEEEeqnarraymulticol{3}{l}{%
    \mathsf{J}(f_{\textnormal{Q}},f_{\textnormal{\^{M}}})}\nonumber\\*\quad%
  & \triangleq &\BigE[M,\vect{Q}]{f_{\textnormal{Loss}}(M,\hat{M})}=\BigE[M,\vect{Q}]{f_{\textnormal{Loss}}\bigl(M,f_{\textnormal{\^{M}}}(\vect{Q})\bigr)}.
  \label{eq:expected-loss_DB}
\end{IEEEeqnarray*}
% \end{displaymath}
Here, the leakage metric $\rho(P_{\vect{Q}|M})$ in~\eqref{eq:def_leakage} is connected to the expected loss $\mathsf{J}(f_{\textnormal{Q}},f_{\textnormal{\^{M}}})$ as $\rho(P_{\vect{Q}|M}) = \max_{f_{\textnormal{\^{M}}}} \mathsf{J}(f_{\textnormal{Q}},f_{\textnormal{\^{M}}})$. Note that the main objective is to minimize the user's leakage to the server, which is equivalent to minimizing $\max_{f_{\textnormal{\^{M}}}} \mathsf{J}(f_{\textnormal{Q}},f_{\textnormal{\^{M}}})$, while the goal of the server is the opposite, i.e., of making the user's loss as large as possible or, equivalently, of maximizing $\mathsf{J}(f_{\textnormal{Q}},f_{\textnormal{\^{M}}})$.

\subsection{Generative Adversarial Approach}

The inference of the server can also be modeled in a generative adversarial fashion~\cite{HuangKairouzChenSankarRajagopal17_1,HuangKairouzChenSankarRajagopal18_1,TripathyWangIshwar19_1,TsengWu20_1}. %The generative adversarial model for the database's inference is formulated as follows.
In particular, the \emph{leakage-distortion} trade-off for any fixed download rate constraint $\const{R}$ is formally described as follows. Consider a family of IR schemes with query generators $f_{\textnormal{Q}}$, answer functions $f_{\textnormal{A}}$, and with download rate $\const{R}(f_{\textnormal{Q}},f_{\textnormal{A}})$ at most $\const{R}$. The goal of the server is to maximize the expected loss $\mathsf{J}(f_{\textnormal{Q}},f_{\textnormal{\^{M}}})$ of the user by designing the decision function $f_{\textnormal{\^{M}}}$.
% , which is equivalent to maximizing the negative of the expected loss
In contrast, the user would like to design a scheme with $f_{\textnormal{Q}}$, $f_{\textnormal{A}}$, and $f_{\textnormal{\^{X}}}$ such that the download rate $\const{R}(f_{\textnormal{Q}},f_{\textnormal{A}}) \leq \const{R}$ and such that the maximum expected loss $\max_{f_{\textnormal{\^{M}}}} \mathsf{J}(f_{\textnormal{Q}},f_{\textnormal{\^{M}}})$  is minimized, while preserving the utility of the scheme, i.e.,  the user can still retrieve the requested file with an expected distortion smaller than a prescribed $\const{D}$. This leads to the constrained minimax optimization problem
%
%
%In particular, we can formulate the problem as a minimax optimization problem as follows:
\begin{subequations}
  \label{eq:GANconstrained}
  % \begin{align}
  \begin{IEEEeqnarray}{rCl}
    \min_{(f_{\textnormal{Q}},f_{\textnormal{A}},f_{\textnormal{\^{X}}})}\max_{f_{\textnormal{\^{M}}}}& &\,\, \mathsf{J}(f_{\textnormal{Q}},f_{\textnormal{\^{M}}})\label{eq:minmax}
    \\
    \text{subject to }& &\,\,
    \bigE[M,\vect{Q},\vect{X}^{[\const{M}]}]{d\bigl(\vect{X}^{(M)}, \hat{\vect X}\bigr)}\leq\const{D}, \notag \\ %\,
    % \label{eq:distortion-constraints}\\[1mm]
    % \\[1mm]
    % & 
    % \quad
    & &\,\,
    \const{R}(f_{\textnormal{Q}},f_{\textnormal{A}})\leq\const{R},\label{eq:GANconstrained_b} \IEEEeqnarraynumspace
    \\
    & &\,\, \hat{\vect{X}} = f_{\textnormal{\^{X}}}(\vect{A},M,\vect Q),\, % \quad
    \vect Q = f_{\textnormal Q}(M, \vect S),
    \nonumber\\
    & &\,\,\,\vect{A}=f_{\textnormal{A}}(\vect{Q},\vect{X}^{[\const{M}]}).\IEEEeqnarraynumspace
% \end{align}
\end{IEEEeqnarray}
\end{subequations}
Note that the minimax formulation in \eqref{eq:minmax} can be written in a GAN form 
\cite{Goodfellow-etal14_1}  %as %according to
%Note that \eqref{eq:minmax}
%can also be written as the classical GAN problem
%\cite{Goodfellow-etal14_1} according to
%\[
%\max_{f_{\textnormal{\^{M}}}} \min_{f_{\bm{Q}}} \E[\bm{Q}\sim P_{\bm{Q}}]{f_{\textnormal{\^{M}}}(\bm{Q})}- \E[M\sim P_{M}]{f_{\textnormal{\^{M}}}(f_{\bm{Q}}(M,\vect{S}))}
%\]
in which $f_{\textnormal{\^{M}}}$ plays the role of the \emph{discriminator} and 
%$(f_{\textnormal{Q}}, f_{\textnormal{A}}, f_{\textnormal{\^{X}}})$ 
$f_{\textnormal{Q}}$ plays the role of the \emph{generator}.  Thus, the machinery of GANs can be used to determine the leakage-distortion trade-off.
In doing so, \eqref{eq:GANconstrained} is first reformulated as the unconstrained optimization problem 
% \begin{equation}
\begin{IEEEeqnarray}{rCl}
  &&\min_{(f_{\textnormal{Q}},f_{\textnormal{A}},f_{\textnormal{\^{X}}})}\max_{f_{\textnormal{\^{M}}}}\Bigl[\mathsf{J}(f_{\textnormal{Q}},f_{\textnormal{\^{M}}})+\eta_1\bigE[M,\vect{Q},\vect{X}^{[\const{M}]}]{d\bigl(\vect{X}^{(M)},\hat{\vect{X}}\bigr)}\nonumber\\
  &&\qquad\qquad\qquad +\>\eta_2 \const{R}(f_{\textnormal{Q}},f_{\textnormal{A}})\Bigr],
  \label{eq:formulation-for-training}\IEEEeqnarraynumspace
\end{IEEEeqnarray}
% \end{equation}
where $\eta_1$ and $\eta_2$ are tuning parameters, and $\vect Q$ and $\vect A$ are according to \cref{Def:Mn-IRscheme}. The minimax game in \eqref{eq:formulation-for-training} with soft decision decoding, i.e., with the log-loss function in \eqref{eq:log_loss_function}, will be the basis for training, as described below in \cref{sec:data_driven_approach}.

Note that the formulations in \eqref{eq:GANconstrained} and \eqref{eq:optimization_RDL} are equivalent in the sense that they give raise to the same rate-distortion-leakage trade-off region. However, the second formulation in \eqref{eq:GANconstrained}  is more amenable to learning as minimizing the rate  as in \eqref{eq:optimization_RDL} is difficult in a data-driven approach, while on the other hand, the first formulation in \eqref{eq:optimization_RDL} is more amenable to theoretical analysis. The  reason the formulation in \eqref{eq:GANconstrained}  is more amenable to learning is that we can easily fix a rate constraint $\const R$ by selecting the size of the output layer of the answer network and the number of quantizing levels $\kappa \geq 2$ used. See \cref{sec:learning} below for further details.

\section{Rate-Distortion-Leakage Trade-off}
\label{sec:theoretical-results}
% In this section, we study the rate-distortion-leakage trade-off for single-server IR.

In this section, we first show that the set  of achievable rate-distortion-leakage triples $(\const{R},\const{D},\const{L})$ from \eqref{eq:optimization_RDL} is a convex set. Then, we consider two different loss functions $f_\textnormal{Loss}(m,\hat{M})$, the $0$-$1$ loss and the  log-loss. Next, we derive an expression for the optimal download rate $\const{R}(f_{\textnormal{Q}},f_{\textnormal{A}})$ for datasets with $\const{M}$ files where $\bigl\{X_i^{(1)},\ldots,X_i^{(\const{M})}\bigr\}_{i=1}^{\beta}=\bigl\{X_i^{[\const{M}]}\bigr\}_{i=1}^\beta$ are independent and identically distributed (i.i.d.), and a known distribution. Finally, we  present a general achievable scheme that can be applied to any dataset.

\begin{lemma}
\label{lem:timesharing}
The set of achievable (feasible) rate-distortion-leakage triples $(\const{R},\const{D},\const{L})$ from \eqref{eq:optimization_RDL} is a convex set, for any leakage metric $\rho$ that is convex in $P_{\vect{Q}|M}$.
\end{lemma}
\begin{IEEEproof}
  See Appendix~\ref{sec:proof_Lemma1}.
\end{IEEEproof}
We remark that the convexity definition of $\rho(\cdot)$ is according to the convexity property of MI~\cite[Ch.~2]{CoverThomas06_1}, i.e., the metric is called \emph{convex} if $\rho(\lambda P_{\vect{Q}_1|M}+(1-\lambda)P_{\vect{Q}_0|M})\leq\lambda\rho(P_{\vect{Q}_1|M})+(1-\lambda)\rho(P_{\vect{Q}_0|M})$ for any $0\leq\lambda\leq 1$, where $P_{\vect{Q}_i|M}$, $i\in\{0,1\}$, are both defined over the same alphabet $[\const{M}]\times\set{Q}$. % Moreover, Lemma~\ref{lem:timesharing} also holds for any loss function $f_{\textnormal{Loss}}$, since any $\rho(P_{\vect{Q}|M})$ that can be expressed via $\mathsf{J}(f_{\textnormal{Q}},f_{\textnormal{\^{M}}})$ is always convex in $P_{\vect{Q}|M}$.

\subsection{Loss Functions}

%\subsubsection{Hard Decision Decoding}
%\label{sec:hard-decision-decoding}
Note that if $f_\textnormal{Loss}(m,\hat{M})$ is the $0$-$1$ loss function \cite{NguyenScott13_1} it can be easily shown, following a similar argument as in \cite[Sec.~2.2]{HuangKairouzChenSankarRajagopal17_1}, that  the optimal inference strategy for the server is the maximum aposteriori probability (MAP) decoder, and that the privacy metric $\rho(P_{\vect{Q}|M})$ in this case equals the server's \emph{inference accuracy} $\Pr(M=\hat{M})$. This case is referred to as hard decision decoding.

%\subsubsection{Soft Decision Decoding}
%\label{sec:soft-decis-decoding}
%
In contrast to hard decision decoding where the server  guesses exactly one file index from the query, we can also consider a soft decision decoding rule for the server. In this case, %$f_{\textnormal{\^{M}}}(\vect{Q})$ can be seen as a distribution over $[\const{M}]$, i.e., 
%{\lin $f_{\textnormal{\^{M}}}(\vect{Q})$ is randomly distributed according to a PMF $F_{\hat{M}}(\cdot|\vect{Q})$.} % , 
likelihoods for all file indices are computed by the server, i.e., the server can infer a PMF $F_{\hat{M}}(\cdot|\vect{Q})$, % , 
where $\sum_{m}F_{\hat{M}}(m|\vect{Q})=1$.
Let $f_{\textnormal{Loss}}(m,\hat{M})$ be the log-loss function defined as  
\begin{equation}
  \label{eq:log_loss_function}
  f_{\textnormal{Loss}}(m,\hat{M}) = \eHP{M}+\log{F_{\hat{M}}(m|\vect{Q})},
\end{equation}
 i.e., the loss is zero when the server's guess is equally likely over $[\const{M}]$. Then, the expected loss function is equal to
 % \begin{align}
 \begin{IEEEeqnarray}{rCl}
   \IEEEeqnarraymulticol{3}{l}{%
     \mathsf{J}(f_{\textnormal{Q}},f_{\textnormal{\^{M}}})}\nonumber\\*%
   & = & \eHP{M}-\sum_{\vect{q}}\sum_{m} P_{M,\vect{Q}}(m,\vect{q})\log{\frac{1}{F_{\hat{M}}(m|\vect{q})}}
   \nonumber\\
   & = &\eHP{M}-\sum_{\vect{q}}P_{\vect{Q}}(\vect{q})\Bigl(-\sum_{m}P_{M|\vect{Q}}(m|\vect{q})\log F_{\hat{M}}(m|\vect{q})\Bigr)
   \nonumber\\
   & = &\eHP{M}-\sum_{\vect{q}}P_{\vect{Q}}(\vect{q})\bigCE{P_{M|\vect{Q}=\vect{q}}(\cdot)}{F_{\hat{M}}(\cdot|\vect{q})}
   \IEEEeqnarraynumspace\label{eq:link_cross-entropy}\\
   & \leq &\eHP{M}-\sum_{\vect{q}}P_{\vect{Q}}(\vect{q})\bigHP{P_{M|\vect{Q}=\vect{q}}(\cdot)}
   \IEEEeqnarraynumspace\label{eq:use_cross-entropy}\\
   & = &\eHP{M}-\sum_{\vect{q}}P_{\vect{Q}}(\vect{q})\bigHPcond{M}{\vect{Q}=\vect{q}},\nonumber
   % \end{align}       
 \end{IEEEeqnarray}
 where $\eCE{P_X}{P_Y}$ denotes the cross-entropy between the distributions $P_X$ and $P_Y$, and~\eqref{eq:use_cross-entropy} follows since for any two distributions $P_{X}(\cdot)$ and $P_{Y}(\cdot)$, we have $\bigCE{P_X(\cdot)}{P_Y(\cdot)}\geq \bigHP{P_X(\cdot)}$. Moreover, observe that
 % \begin{IEEEeqnarray*}{c}
 \[ \max_{f_{\textnormal{\^{M}}}}\bigl(\mathsf{J}(f_{\textnormal{Q}},f_{\textnormal{\^{M}}})\bigr)\leq \HP{M}-\bigHPcond{M}{\vect{Q}}=\MI{M}{\vect{Q}},
   % \end{IEEEeqnarray*}
 \]
and from \eqref{eq:use_cross-entropy}, it follows that equality holds if $P_{M|\vect{Q}=\vect{q}}(\cdot)=F_{\hat{M}}(\cdot|\vect{q})$. Therefore, under the log-loss function, the leakage to the server is measured in terms of MI, i.e.,  $\rho(P_{\vect{Q}|M})=\MI{M}{\vect{Q}}$, and the user wishes to minimize the MI leakage. %Furthermore, \eqref{eq:link_cross-entropy} gives an explicit reason to use empirical categorical cross-entropy as loss function for the data-driven approach in \cref{sec:data_driven_approach}.  
% However, if $f_{\textnormal{Loss}}(m,\hat{M})$ is the log-loss function defined as  
% \begin{equation}
%   \label{eq:log_loss_function}
%   f_{\textnormal{Loss}}(m,\hat{M}) = \eHP{M}+\log{F_{\hat{M}}(m|\vect{Q})},
% \end{equation}
% then the leakage to the server is measured in terms of MI, i.e., $\rho(P_{\vect{Q}|M})=\MI{M}{\vect{Q}}$. 
This case is referred to as soft decision decoding. Moreover, % the detailed derivation, which can be found in Appendix~\ref{sec:soft-decis-decoding},
this gives an explicit reason (see~\eqref{eq:link_cross-entropy}) to use empirical categorical cross-entropy as loss function for the data-driven approach in \cref{sec:data_driven_approach}.

Finally, note that it can be seen that Lemma~\ref{lem:timesharing} also holds for any loss function $f_{\textnormal{Loss}}$, since any $\rho(P_{\vect{Q}|M})$ that can be expressed via  an expectation as $\mathsf{J}(f_{\textnormal{Q}},f_{\textnormal{\^{M}}})$ is always convex in $P_{\vect{Q}|M}$. Thus, in the rest of paper, we will always assume that the metric $\rho(\cdot)$ is convex in its argument.

\subsection{Optimal Download Rate for Data With Known Distribution}

\begin{theorem}
  \label{thm:RDL-function}
  Assume that $\bigl\{X_i^{[\const{M}]}\bigr\}_{i=1}^{\beta}$ are i.i.d.\ and that for any $i\in[\beta]$, the RVs $\bigl\{X_i^{(m)}\bigr\}_{m=1}^{\const{M}}$ are distributed according to a prototype joint PMF $P_{X^{(1)},\ldots,X^{(\const{M})}}$. Then, the download rate given in \eqref{eq:optimization_RDL} equals %to the following information rate-distortion-leakage function defined as
  \begin{IEEEeqnarray}{rCl}
      \const{R}(\const{D},\const{L}) %\nonumber\\*\quad%
    & \eqdef &\min_{P_{\vect{Q}|M},P_{\hat{X}^{[\const{M}]}|X^{[\const{M}]},\vect{Q}}\in\set{F}(\const{D},\const{L})}\bigMIcond{X^{[\const{M}]}}{\hat{X}^{[\const{M}]}}{\vect{Q}},
    \IEEEeqnarraynumspace\label{eq:RDL-function}
  \end{IEEEeqnarray}
  as $\beta\to\infty$, where
  \begin{IEEEeqnarray*}{rCl}
    \set{F}(\const{D},\const{L})& \eqdef &\biggl\{P_{\vect{Q}|M}, P_{\hat{X}^{[\const{M}]}|X^{[\const{M}]},\vect{Q}}\colon\nonumber\\
    &&\!\!\!\!\!\!\!\!\!\!\!\!\bigE[M,\vect{Q},\vect{X}^{[\const{M}]}]{d_{\textnormal{sym}}\bigl(X^{(M)},\hat{X}^{(\const{M})}\bigr)}\leq\const{D},\,\rho(P_{\vect{Q}|M})\leq\const{L}\biggr\}\IEEEeqnarraynumspace\label{eq:RDL-constraints}
  \end{IEEEeqnarray*}
is the set of feasible distributions $P_{\vect{Q}|M}$ and $P_{\hat{X}^{[\const{M}]}|X^{[\const{M}]},\vect{Q}}$ for which the joint distribution $P_{M,\vect{Q},X^{[\const{M}]},\hat{X}^{[\const{M}]}}$ satisfies the leakage and distortion constraints of $\set{F}(\const{D},\const{L})$. % {\eirik and where $\hat{X}^{[\const{M}]}$ are  prototype RVs of the reconstructed symbols.}
\end{theorem}
The proof that the rate $\const{R}$ is bounded from below by $\const{R}(\const{D},\const{L})$, for any $\beta$, is given in~Appendix~\ref{sec:proof-theorem1}.
%While the achievability proof is not provided, we do give the
%following intuition and relative technique %from information theory.
We provide the following intuition based on information-theoretical considerations. Note that our single-server IR scheme can be seen as a source coding problem with side information, where the server and the user act as an encoder and a decoder, respectively, and where the query can be seen as a controllable side information known to both the encoder and decoder in order to enforce the privacy condition. This is similar to the well-known Wyner-Ziv problem~\cite{WynerZiv76_2}. Hence, a random-coding based scheme can be used to achieve $\const{R}(\const{D},\const{L})$ as $\beta\to\infty$. Further, it can be shown that the alphabet size of the designed queries can be restricted to $\const{M}+3$, which follows from applying Carath{\'{e}}odory’s theorem~\cite[Thm.~15.3.5]{CoverThomas06_1}. Note, however, that finding a closed-form expression for $\const{R}(\const{D},\const{L})$ in our case is intractable. % even for {\eirik files with $\bigl\{X_i^{[\const{M}]}\bigr\}_{i=1}^\beta$ being i.i.d.}. 
Hence, we provide numerical results for Gaussian data in~\cref{sec_numerical_results}, which verify, by combining the achievable scheme proposed below  in~\cref{sec:theoretical_approach} with a convexifying approach, that \eqref{eq:RDL-function} is indeed achievable.

\subsection{Achievable Schemes}
\label{sec:theoretical_approach}

In this subsection, we present a general achievable scheme for an arbitrary number of files $\const{M}$ and hard decision leakage (or accuracy) $\const L \in \{1,\nicefrac 12, \nicefrac 13, \dotsc, \nicefrac 1{\const M}\}$. 

%The scheme is inspired by 
%the methods of lossy data compression and the capacity-achieving single-server weakly-private %information retrieval scheme presented in \cite[Sec.~VI]{LinKumarRosnesGraellAmatYaakobi21_1}.
%, which is also presented in \cite[Sec.~VI]{YakimenkaLinRosnesKliewer20_1sub}. 
%This general IR scheme works for any dataset stored on a single server (finite or infinite).

The construction is based on the following fact. If $\const L = 1$, we have the problem of lossy compression: the user explicitly tells the server what she needs to download, and the server sends the requested data compressed with some pre-agreed method. One example of such compression is quantization. This scheme is designed against a maximum likelihood decoder that provides the best inference for the server. Assume that the user wishes to retrieve the $M$-th file and to ensure a leakage of $\const L = \nicefrac 1{\const N}$, $1 \le \const N \le \const M$. 
%This scheme is designed against a strong theoretical maximum-likelihood (ML) decoder for the server. 
%We describe the steps of the scheme as follows.

\begin{itemize}

\item The user  first selects a lossy source coding scheme $\code{C}$ to encode (separately or together) $\const N$ files of size $\beta$ each. Assume that the source coding scheme achieves an average compression size (in bits) of $\log_2 |\mathcal{C}|$ and expected normalized distortion  $\const{D}$ (normalized by $\const{N} \beta$).%, respectively. %\footnote{Note that here, the rate $\const R$ is calculated per $\const N \beta$ symbols.}

\item The user's query is designed to exactly request $\const{N}$ files with indices $M_1,\ldots,M_\const{N}$ such that $M\in\{M_1,\ldots,M_\const{N}\}$, where the $\const{N}-1$ nondesired indices are chosen uniformly at random. These $\const N-1$ file indices are added in order to ``trick'' the server and hide the real file index $M$ of interest.

\item After receiving the queries sent by the user, the server compresses all  $\const{N}$ files by using the source coding scheme $\code{C}$, and transmits the answer back to the user.

%\item Since the source coding scheme is used to encode each of the $\const{N}$ files including the requested file, the user is guaranteed to reconstruct the requested file with a distortion level $\const{D}$ from the answer.
\item The user decompresses the answer using the selected source coding scheme, and the $\const{N}-1$ nondesired reconstructed files are discarded.
% by the user, while the desired reconstructed file is kept.
\end{itemize}

The rate of this IR scheme is $\const{R} = \log_2 |\mathcal{C}| / \beta$, and the expected distortion is  $\const{D}$ and the leakage is $\const{L}$. The performance of the scheme is strongly dependent on the used source coding scheme $\code{C}$. If the distributions of the files differ significantly, a separate encoder for each subset $\{M_1, M_2, \dotsc, M_{\const N}\}$ will achieve better results compared to a single encoder for all $\const N$-subsets. Using \cref{lem:timesharing}, a scheme for any leakage $\nicefrac{1}{\const{M}} \leq \const{L} \leq 1$ (not only reciprocal of an integer) can be constructed.

From the general scheme above, two specific schemes can be constructed based on the source coding scheme used. The first scheme is constructed by selecting a good source coding scheme for a finite $\beta$, e.g., a scheme based on quantization. In particular, for Gaussian data we use a variant of the generalized Lloyd algorithm~\cite{Lloyd82_1,LindeBuzoGray80_1} in order to construct a good quantization scheme. In the sequel, we will refer to this scheme as the compression-based scheme (cf.~Appendix~\ref{sec:achievable_schemes}). The second scheme, referred to in the sequel as Shannon's scheme, assumes $\beta \to \infty$ and uses the well-known rate-distortion function from information theory for the source coding scheme \cite{Gray73_1}. In~\cref{fig:quant-schemes}(a) below, combined with a convexifying approach described in Appendix~\ref{sec:shannon_GaussianData}, we plot the performance of both schemes for a synthetic Gaussian dataset. It is worth mentioning that by numerically solving~\eqref{eq:RDL-function} for the considered Gaussian data, the lower bound and the approximation values of Shannon's scheme are actually quite close, which shows that Shannon's scheme performs very close to the information-theoretical optimum for  the considered  Gaussian data. % xxx: How to say about "i.i.d." Gaussian dataset?

%that the converse bound is tight for the i.i.d.\ Gaussian case.

%We remark that this theoretical scheme achieves the download rate $\const{N}\const{R}$, distortion $\const{D}$, and leakage $\const{L}=\nicefrac{1}{N}$. The performance of this scheme is strongly dependent on the used source coding scheme $\code{C}$.

\section{Data-Driven Approach}
\label{sec:data_driven_approach}

In this section, we describe % in detail
our new data-driven framework for constructing an efficient IR scheme for downloading an arbitrary file from an arbitrary dataset stored on a single server. The four functions in Fig.~\ref{fig:DNN-model} are represented as deep neural networks and we assume, for now, 
%
%The data flow diagram for the proposed framework is illustrated in Fig.~\ref{fig:DNN-model} and consists of four neural networks, two at the user side, denoted by NNQ and NND, and two at the server side, denoted by NNS and NNA. 
that they  have already been trained. %, and they are used as black boxes in the description below.

The user wishes to retrieve the  $M$-th file $\vect X^{(M)}$ and encodes $M$ as a one-hot $\vect Y = (Y_1, Y_2, \ldots , Y_{\const{M}}) \in \{0,1\}^{\const{M}}$, where $Y_j=1$ if $j=M$, and $Y_j=0$ otherwise.
% \[
% Y_j = \begin{cases}
% 	1 & \text{if $j = M$},\\
% 	0 & \text{otherwise}.
% \end{cases}
% \]
Next, the user generates a ``noise'' vector $\vect S = (S_1, S_2, \ldots, S_{\const{M}})$, where $S_1, S_2,\ldots, S_{\const{M}}$ are i.i.d.\ according to the standard Gaussian distribution $\mathcal N(0, 1)$.
%\[
%	S_i \sim \mathcal N(0, 1), \quad i=1,2,\ldots,\const{M}.
%\]
%The length of the query is $\const{M}$.
The concatenation $(S_1, S_2, \ldots, S_{\const{M}}, Y_1, Y_2,\ldots, Y_{\const{M}})$ is the input to a deep neural network, representing the function $f_{\textnormal{Q}}$, for query generation. This network produces the query $\vect Q$ that is sent to the server. The intuition behind this neural network is to hide the value of $M$. %Finally, the query $\vect Q$ is sent to the server.
%The intuition is as follows.
The server's answer is produced by feeding the stored data $\vect X^{[\const{M}]}$ and the received query $\vect Q$ into a deep neural network, representing the function $f_{\textnormal{A}}$, for answer construction. The deep neural network produces the answer vector $\vect A$  that is sent back to the user. The user then feeds $\vect A$ and $(S_1, \ldots, S_{\const{M}}, Y_1, \ldots, Y_{\const{M}})$  into a deep neural network for decoding, representing the function $f_{\textnormal{\^{X}}}$, to produce the estimate $\hat{\vect X}$ of the requested file $\vect X^{(M)}$.

% We use squared-error as distortion measure in neural network training:
% \[
% 	d(X_{i}^{(m)},\hat X_{i}^{(m)}) = \frac{1}{\beta} \sum_{i=1}^{\beta} \left( X_{i}^{(m)} - \hat X_{i}^{(m)} \right)^2.
% \]

On the server side, a deep neural network, representing the function $f_{\textnormal{\^{M}}}$, is used to guess the identity of the requested file. The input to the  network is  the query vector $\vect Q$ and the output  (using softmax) is a distribution-like vector
%
% $\const{M}$ real numbers which are transformed by a softmax function to a 
%\begin{gather*}
	$\vect{W} = \vect{W}(\vect{Q})=(W_1, W_2, \ldots, W_{\const{M}})$, where 
	$\sum_{j=1}^{\const{M}} W_j = 1$,
	%\quad
	$0 \le W_j \le 1$,
%\end{gather*}
and where $W_j$ can be interpreted as ``with probability/likelihood $W_j$, the user's requested file index $M$ is equal to $j$.'' The server's estimate of $M$ %the user requested file index 
is then
%\[
	$\hat{M} = \argmax_{j \in [\const{M}]} W_j$.
%\]
%and the accuracy the server's guess is 
%\[
%	L = \Pr [M = \hat{M}].
%\]
%Note that the server can always just randomly guess $M$ (or just always output some constant value, e.g., $\hat{M}=1$). This gives a lower bound for the accuracy $1/\const{M}$.

\begin{algorithm}[tb!] % xxx: fix algorithm notation
  \caption{Training algorithm for generative adversarial user privacy}      
  \label{alg:training}
  % \SetKwFunction{CompEraPat}{ComputeErasurePatternList}
  % \SetKwFunction{CompEraMat}{ComputeMatrix}
  % \SetKwFunction{CompInfSet}{ComputeInformationSetList}
  \begin{algorithmic}[1]
    % \SetKwInOut{Input}{Input}
    % \SetKwInOut{Output}{Output}
    % \SetAlgoLined
    % \DontPrintSemicolon
    % Some LaTeX compilers require you to use \dontprintsemicolon instead
    \STATE {\bfseries Input:} Number of training samples $\const{n}$, training samples $\bigl\{\bigl(\vect{x}^{(1)}(l),\ldots,\vect{x}^{(\const{M})}(l)\bigr)\bigr\}_{l\in[\const{n}]}$, number of training iterations $\const{T}$, size of minibatch for stochastic gradient descent  $\const{b}$, and initial tuning parameter $\eta_{\textnormal{initial}}$
    \STATE {\bfseries Output:} $f_{\textnormal{Q}}$, $f_{\textnormal{A}}$, $f_{\textnormal{\^X}}$, and $f_{\textnormal{\^M}}$
    
    % $\const{n} \leftarrow$ Number of training samples in each class\\
    % $\const{b} \leftarrow$ Size of minibatch \\
    % $\eta \leftarrow$ Tuning parameter \\
    \STATE $t \leftarrow 1$,\,
    %\\
    %\STATE 
    $\eta \leftarrow \eta_{\textnormal{initial}}$\\
    \STATE Initialize the neural networks representing $f_{\textnormal{Q}}$,  $f_{\textnormal{A}}$,
    $f_{\textnormal{\^{X}}}$,  $f_{\textnormal{\^{M}}}$\\
    \STATE Initialize  $\vect{q}^{(m)}_l$ and $\vect{a}^{(m)}_l$, $l \in [\const{n}]$, $m \in [\const{M}]$\\ 
    
    % \For{\textnormal{number of training iterations}}{
    \WHILE{$t \leq \const{T}$}      
      \FOR{$m \in [\const{M}]$}
      %\STATE Generate an index minibatch ${\lin\set{B}^{(m)}}\subseteq{[\const{n}]}$  of size $\const{b}$%corresponding to the $m$-th file,  ${\lin\ecard{\set{B}^{(m)}}} = \const{b}$
      %\\ 
      \STATE Generate $\const{b}$ noise samples  $\{\vect{s}_b\}_{b \in [\const{b}]}$ from %the %distribution 
      $\mathcal{N}(\vect{0},I)$\\
      
      \FOR{$b \in [\const{b}]$}
      \STATE $\vect{q}^{(m)}_b  \leftarrow f_\textnormal{Q}(m,\vect{s}_{b})$ \\
      \ENDFOR
      \ENDFOR      
      \STATE Update $f_{\textnormal{\^{M}}}$ by stochastic gradient descent using %descending its stochastic gradient: 
      \\
      \STATE $\nabla \left(\sum_{m \in [\const{M}]}  \sum_{b \in [\const{b}]} \frac{1}{\const{b} \cdot \const{M} }  \log w_{m}\bigl(\vect{q}^{(m)}_b\bigr)\right)$ %f_{\textnormal{Q}}(l,s_{k \times \const{M} + l}))$
      
      \FOR{$m \in [\const{M}]$}
      \STATE Generate an index minibatch $\set{B}^{(m)} \subseteq{[\const{n}]}$  of size $\const{b}$ %corresponding to the $m$-th file,  $|{\lin\set{B}^{(m)}}| = \const{b}$
      \\
      \STATE Generate $\const{b}$ noise samples  $\{\vect{s}_b\}_{b \in \set{B}^{(m)}}$ from %the %distribution 
      $\mathcal{N}(\vect{0},I)$
      \\  
      %\STATE Generate a minibatch of $\const{b}$ training samples $\bigl\{{\lin\vect{X}^{(m)}(b)}\colon b \in {\lin\set{B}^{(m)}} \bigr\}$ % \vect{X}_{2}^{(m)},\ldots,\vect{X}_{\const{b}}^{(m)} \}$         
      
      % \{ \vect{X}_{0}^{(2)}, \vect{X}_{1}^{(2)},\ldots,\vect{X}_{m-1}^{(2)} \}, \ldots, \{ \vect{X}_{0}^{(\const{M})}, \vect{X}_{1}^{(\const{M})},\ldots,\vect{X}_{m-1}^{(\const{M})} \} \}$       
      
      \FOR{$b \in \set{B}^{(m)}$}
      \STATE $\vect{q}^{(m)}_b  \leftarrow f_\textnormal{Q}(m,\vect{s}_{b})$
      \STATE
      $\vect{a}^{(m)}_b \leftarrow f_{\textnormal{A}}{\bigl(\vect{q}^{(m)}_b,\vect{x}^{(1)}(b),\ldots,\vect{x}^{(\const{M})}(b)\bigr)}$
      \ENDFOR
      \ENDFOR
      
      %\FOR{$m \in [\const{M}]$}      
      %\FOR{$b \in {\lin\set{B}^{(m)}}$}
      % Let $\vect{a}_{m,b} \leftarrow f_{\textnormal{A}}(\vect{q}_{m,b},\vect{X}_b^{([\const{M}])})$, $\forall\, b \in {\lin\set{B}^{(m)}}$, $\forall\, m \in \const{M}$
      %\STATE
      %$\vect{a}_{m,b} \leftarrow f_{\textnormal{A}}{\lin\bigl(\vect{q}_{m,b},\vect{X}^{(1)}(b),\ldots,\vect{X}^{(\const{M})}(b)\bigr)}$
      %\ENDFOR
      %\ENDFOR
      % Generate $\const{b}$ noise samples $\{ \vect{s}_{1},\vect{s}_{2},\ldots, \vect{s}_{\const{b}} \}$ from the distribution $\mathcal{N}(\vect{0},I)$\\
      \STATE Update $f_{\textnormal{Q}}$, $f_{\textnormal{A}}$,  $f_{\textnormal{\^{X}}}$ by stochastic gradient descent using
      \\
      \STATE $\nabla \Bigl( \sum_{m \in [\const{M}]}\sum_{b \in \set{B}^{(m)}}\frac{1}{\const{b} \cdot \const{M}}\Bigl[\log w_{m}\bigl(\vect{q}^{(m)}_b\bigr)$\\$\quad\quad\quad\quad\quad\quad+\,\eta \cdot d\bigl(\vect{x}^{(m)}(b), f_{\textnormal{\^{X}}}\bigl(\vect{a}^{(m)}_b,m,\vect{q}^{(m)}_b\bigr) \bigr)\Bigr] \Bigr)$
      \\       
      % $\sum_{k=0}^{\const{b}-1}  \frac{1}{m \times \const{M}}\sum_{l=1}^{\const{M}}\left[
      %   \sum_{j=1}^{\const{M}} Y_{k,l,j} \log f_{\textnormal{\^{M}},j} (f_{\textnormal{Q}}(l,s_{k \times \const{M} + l})) + \eta \cdot d \left(\vect X_{k}^{(l)}, f_{\textnormal{\^{X}}}(f_{\textnormal{A}}(f_{\textnormal{Q}}(l,s_{k \times \const{M} + l}),X_{k}^{(l)}),l,s_{k \times \const{M} + l}) \right) \right]$ 
      \STATE Update the tuning parameter $\eta$
      \\
      \STATE $t \leftarrow t+1$
    \ENDWHILE

    \STATE {\bfseries Return:} $f_{\textnormal{Q}}$, $f_{\textnormal{A}}$, $f_{\textnormal{\^X}}$, and $f_{\textnormal{\^M}}$
  \end{algorithmic}
\end{algorithm}        

\subsection{Learning Algorithm}
\label{sec:learning}

%Let  $\const{n}$ denote the number of training samples, $M_l$ the $l$-th requested file index, $\vect{X}^{(M_l)}
%=\bigl(X_{1}^{(M_l)},\ldots,X_{\beta}^{(M_l)}\bigr)
%$ the $l$-th requested file, and $\hat{\vect{X}}_l$  the user reconstructed estimate of the $l$-th requested file.
%Then, the distortion between $\mtrx{X}$ and $\hat{\mtrx{X}}$ for the Gaussian case is
%\begin{displaymath}
%  d(\mtrx{X},\hat{\mtrx{X}})
% \eqdef
%  \frac{1}{\const{n}}\sum_{l=1}^{\const{n}} d_{\mathrm{SE}}\left( \vect X^{(m_l)}, \hat{\vect X}_l \right).
%\end{displaymath}
%
%Moreover, if the server's prediction is $\hat{m}_l$ for the $l$-th sample, we estimate the accuracy  by 
%\begin{equation}
%\hat{L} = \frac{\sum_{l=1}^{\const{n}}\mathbbm{1} \{  m_l = \hat{m}_l\} }{\const{n}}.
%\end{equation}
% where $\mathbbm{1}$ is indicator function.
Training the deep neural networks representing the functions $f_{\textnormal{Q}}$, $f_{\textnormal{A}}$, $f_{\textnormal{\^{X}}}$, and $f_{\textnormal{\^{M}}}$ is done following~\eqref{eq:formulation-for-training} and~\eqref{eq:link_cross-entropy} by first fixing a download rate constraint $\const{R}$ and then solving the minimax optimization problem 
% \begin{equation}
\begin{IEEEeqnarray}{rCl}
  \min_{(f_{\textnormal{Q}}, f_{\textnormal{A}}, f_{\textnormal{\^{X}}})}\max_{f_{\textnormal{\^{M}}}}\frac{1}{\const{n} \cdot \const{M}}\sum_{m=1}^{\const{M}}\sum_{l=1}^{\const{n}}&&\Bigl[
  -f_{\textnormal{XE-Loss}}\bigl(\vect{y}^{(m)}(l),\vect{w} \bigr)\nonumber\\
  &&\qquad+\>\eta\cdot d\bigl(\vect{x}^{(m)}(l),\hat{\vect{x}}\bigr)
  %\frac{1}{\beta}\sum_{i=1}^\beta \bigl(x^{(m_j)}_{i}-\hat x^{(m_j)}_{j,i}\bigr)^2
  %+\eta  %\log f_{m_j}
  \Bigr],
  % \end{equation}
  \IEEEeqnarraynumspace\label{eq:training}
\end{IEEEeqnarray}
where 
% \[
$f_{\textnormal{XE-Loss}}\bigl(\vect{y}^{(m)}(l),\vect{w}\bigr) = - \sum_{j=1}^{\const{M}} y^{(m)}_j(l) \log w_{j} = -\log w_{m}$ 
% \]
measures the categorical cross-entropy between $\vect{y}^{(m)}(l) = \bigl(y^{(m)}_{1}(l),\ldots,y^{(m)}_{\const{M}}(l)\bigr)$ and the corresponding  $\vect{w} = (w_{1},\ldots,w_{\const{M}})$ at the output of the adversary  network, and  $\const{n}$ is the number of training samples. Here, $\bigl(\vect{x}^{(1)}(l),\ldots,\vect{x}^{(\const{M})}(l)\bigr)$ %{\lin $\vect{x}^{({\eirik m})}(l)=\bigl(x_{1}^{({\eirik m})}(l),\ldots,x_{\beta}^{({\eirik m})}(l)\bigr)$} 
denotes the {$l$-th} training sample, $l\in[\const{n}]$, and $\vect{y}^{(m)}(l)$  is the  corresponding $\vect{y}$-vector  when requesting the $m$-th file. %and {\lin $\hat{\vect{X}}(l)=\bigl(\hat{X}_1(l),\ldots,\hat{X}_{\beta}(l)\bigr)$} the corresponding user reconstructed estimate.}
The parameter $\eta$ is a trade-off coefficient that is slightly increased in every epoch (typically by a value in the range from $0.00005$ to $0.0005$, depending on the actual learning scenario). The initial value of $\eta$ is typically picked in the range from $0.1$ to $2$. The rate is computed based on the dimension of the answer network's output and the corresponding quantizer levels like~\cite{BlauMichaeli19_1,Mentzer-etal18_1}. In particular, the quantization in the forward direction (but after the sigmoid activation of the output layer) is  done  by nearest-neighbor assignment with a uniformly distributed noise added in order to improve the training, i.e., the $i$-th quantized output during training  is $a^\mathsf{f}_i = \argmin_{\ell \in \mathcal{L}} \lVert \tilde{a}_i - \ell \rVert_2 + \eUniform{[\nicefrac{-1}{2(\kappa-1)},\nicefrac{1}{2(\kappa-1)}]}$, where $\mathcal{L} = \{\ell_1,\ldots,\ell_\kappa \}$ is a \emph{fixed} set of $\kappa \geq 2$ linearly equally spaced  quantization points  %evenly spaced 
in the output range $[0, 1]$ of the sigmoid function (i.e., %$\ell_i = \nicefrac{(2i-1)}{2\kappa}$ for $i \in [\kappa]$), 
$\ell_i = \nicefrac{(i-1)}{(\kappa-1)}$ for $i \in [\kappa]$),
$\tilde{a}_i$ denotes the $i$-th output of the sigmoid activated output layer, $\eUniform{\mathcal{X}}$ denotes a RV that is uniformly distributed over the set $\mathcal{X}$, and $\lVert\cdot\rVert_p$ is the $\ell^p$-norm. Here, the width of the added uniformly distributed noise is set to be the same as the distance between two quantization points ($\nicefrac{1}{(\kappa-1)}$)\ifthenelse{\boolean{arxiv_version}}{~\cite{BalleLaparraSimoncelli17_1}.}{.} In the testing phase, no noise is added to the quantized output. In order to compute gradients in the backward pass,  we use the differential ``soft'' assignment % xxx: discussion
\begin{equation*}
a_i^\mathsf{b} = \sum_{j=1}^\kappa \frac{\mathrm{e}^{- \lVert \tilde{a}_i - \ell_j \rVert_1}}{\sum_{l=1}^\kappa \mathrm{e}^{- \lVert \tilde{a}_i - \ell_l \rVert_1}} \ell_j \approx a_i^\mathsf{f}
\end{equation*}
as an approximation to the nearest-neighbor assignment with noise, or $a_i^\mathsf{f}$. 
The quantization scheme described above follows \cite[App.~E]{BlauMichaeli19_1} (see also \cite{Mentzer-etal18_1}). %Hence, by changing $\kappa$ or the dimension of the sigmoid activated output layer of the answer network, different rates and be achieved and the rate constraint of (5b) can be easily enforced %when optimizing the expected loss as formulated in (5a). 

To fix the rate, we fix the dimension of the answer network's output layer and also the corresponding number of  quantizing levels $\kappa$, which gives an upper bound on the operational rate in \eqref{eq:def_rate}, and the rate constraint of \eqref{eq:GANconstrained_b} can be easily enforced. To vary the fixed rate, we vary the dimension of the answer network's output layer  %(we found that having $\kappa=2$ levels  works best for the rates we considered)}. 
with $\kappa=2$. %, as in \cite{BlauMichaeli19_1}}}. 
The solution to the minimax optimization problem in~\eqref{eq:training} is found using an iterative algorithm employing minibatch stochastic gradient descent outlined in Algorithm~\ref{alg:training}, which is similar to~\cite[Alg.~1]{HuangKairouzChenSankarRajagopal17_1}. In particular, the solution is found by first maximizing  the objective function of~\eqref{eq:training} to determine the optimal $f_{\textnormal{\^{M}}}$ for a fixed initial triple $(f_{\textnormal{Q}}, f_{\textnormal{A}}, f_{\textnormal{\^{X}}})$. Next, the optimal triple $(f_{\textnormal{Q}}, f_{\textnormal{A}}, f_{\textnormal{\^{X}}})$ is found for the given  $f_{\textnormal{\^{M}}}$ from the previous step by minimizing the same objective function. This iterative process is continued until convergence or the maximum number of iterations is exceeded. Note that, although training is done based on a log-loss function (see~\eqref{eq:log_loss_function}), we evaluate the performance more intuitively using accuracy, i.e., $\Pr(M=\hat{M})$, in Section~\ref{sec_numerical_results} below.

\section{Numerical Results}
\label{sec_numerical_results}

We demonstrate the application of our proposed data-driven approach to a synthetic Gaussian dataset and also to the MNIST, CIFAR-$10$, and LSUN datasets, showing that guaranteeing a certain privacy level leads to a higher rate-distortion trade-off curve, and hence a sacrifice in either download rate or distortion. We refer the reader to Appendix~\ref{sec:Details_learning_alg} for further details of the learning. The LSUN dataset is considered a large-scale dataset (comprising $24$ times as many images as CIFAR-$10$), and in order to expedite the learning, we consider a downsized version comprising $64 \times 64$ pixels color  images. We also remark here that training with very large file sizes  is not the main target application area for this framework. However, in case we do have very large files, the files can be split into smaller blocks and then we can train a scheme for the smaller blocks as outlined in \cite[Sec.~VI-A]{YakimenkaLinRosnesKliewer22_1}, e.g., a $128 \times 128$ pixels image can be split into $4$ separate smaller $64 \times 64$ pixels sub-images. The resulting overall scheme has the same rate, distortion, and leakage as the small scheme  trained on blocks considered as independent images. %(assuming the smaller images have the same distribution).} %Note that in this case the user needs to generate the query only once and can reuse it for all instances of the small scheme.}
We also compare the data-driven approach with both the compression-based scheme (for all datasets) and Shannon's scheme (for the Gaussian dataset) of \cref{sec:theoretical_approach}.
%
%
%to construct efficient IR schemes for the MNIST dataset, where a user
%intends to download a specific digit from this dataset under
%nonperfect reconstruction and privacy constraints.
%
%
%Results for CIFAR-$10$ are presented in the appendix, as well as a discussion of the query upload costs and the cost of transferring the actual answer network to the server in comparison to the download cost of the answers.
%
%
%
%
%
%\subsection{Gaussian Dataset}
%
The Gaussian dataset consists of $\const{M}=4$ files, each of dimension $\beta = 3$ and drawn independently according to $\mathcal N(\vect \mu^{(m)}, \sigma^2 I)$, where  
%\begin{IEEEeqnarray*}{llCll}
	$\vect \mu^{(1)} = \trans{(3, 3, 3)}$, 
	$\vect \mu^{(2)} = \trans{(3, -3, -3)}$, 
	$\vect \mu^{(3)} = \trans{(-3, 3, -3)}$, 
	$\vect \mu^{(4)} = \trans{(-3, -3, 3)}$, 
	and $\sigma = 3$. 
%\end{IEEEeqnarray*}
 %$d_{\mathrm{SE}} \colon \Reals^\beta \times \Reals^\beta \to \Reals_{\ge 0}$:
%\begin{displaymath}
%  d_{\mathrm{SE}}(\vect X^{(m)}, \hat {\vect X}) = \frac 1\beta \sum_{i=1}^\beta \left( X_i^{(m)} - \hat X_i \right)^2.
%\end{displaymath}
%We also demonstrate the application of our proposed data-driven
%approach to construct efficient IR schemes for the MNIST dataset, where a user
%intends to download a specific digit from this dataset under
%nonperfect reconstruction and privacy constraints. 
 %$1000000$ ($250000$ for each of the $\const M = 4$ files)}
 %
We pre-generated $1000000$ ($100000$) data points for training  (testing), $250000$ ($25000$) for each of the $\const M = 4$ files. For MNIST, CIFAR-$10$, and LSUN, $\const{M}=10$ (corresponding to the number of digits/classes),  the training set is comprised of respectively $\const{n}=6000$, $5000$, and $120000$ images of size $28 \times 28$, $32 \times 32$, and $64 \times 64$ pixels for each digit/class, and testing is performed on randomly chosen images from $1000$ test images from each digit/class for MNIST and CIFAR-$10$ and from $4000$ test images from each class for LSUN. The distortion between a requested file $\vect{X}^{(m)}$ and its estimate $\hat{\vect{X}}$ is measured as the per-symbol squared error. 
%
%
% The CIFAR-$10$ dataset consists of $\const{n}=5000$ and size $32 \times 32$ pixels $3$ channels training images for each classes. There are $10$ classes in CIFAR-$10$ dataset and there are $1000$ testing images for each class.

\begin{figure*}[tb!]
\centering
\subfloat[]{
  % \resizebox{0.33\textwidth}{0.3\textwidth}{
  \scalebox{0.9}{
    % This file was created by tikzplotlib v0.9.1.
\begin{tikzpicture}

\definecolor{mplblue}{rgb}{0.12156862745098,0.466666666666667,0.705882352941177}
\definecolor{mplorange}{rgb}{1,0.498039215686275,0.0549019607843137}
\definecolor{mplgreen}{rgb}{0.172549019607843,0.627450980392157,0.172549019607843}

\pgfplotsset{every tick label/.append style={font=\tiny}}

\pgfplotsset{
	/pgfplots/xlabel near ticks/.style={
		/pgfplots/every axis x label/.style={
			at={(ticklabel cs:0.5)},anchor=near ticklabel
		}
	},
	/pgfplots/ylabel near ticks/.style={
		/pgfplots/every axis y label/.style={
			at={(ticklabel cs:0.5)},rotate=90,anchor=near ticklabel}
	}
}

\begin{axis}[
legend style={cells={align=left},font=\small}, % required for multiline legend entries
tick label style={/pgf/number format/fixed},
width=0.5\textwidth,
height=0.3\textheight,
legend cell align={left},
legend style={fill opacity=0.8, draw opacity=1, text opacity=1, draw=white!80!black},
%tick align=outside,
%tick pos=both,
grid style={gray,opacity=0.5,dotted},
xlabel={Per-symbol squared error distortion},
xlabel near ticks,
xlabel style={font=\small},
xmajorgrids,
xmin=-0.05, xmax=6.75,
xtick style={color=black},
%<<<<<<< HEAD
%ylabel={Leakage $\const{L}$},
%=======
%ylabel={Leakage $\const L$},
ylabel={Accuracy},
ylabel near ticks,
ylabel style={font=\small},
%>>>>>>> b97d8d44fc3ff11aab3ff5ce4b9d2b6554f83bbb
ymajorgrids,
ymin=0.23, ymax=1.02,
%ytick style={color=black}
]

%\addlegendimage{empty legend};
%\addlegendentry{$\const{R}=2$};

\addplot [semithick, mplblue, mark=*, mark options={solid}]
table [x=Distortion, y=Accuracy, col sep=comma] {gaussian6bitsData};
% table [x=Distortion, y=Accuracy, col sep=comma] {../simulation-results_Mark/gaussian6bitsData};
%table {%
%	2.371161 1
%	2.716265 0.750977
%	4.388816 0.506836
%	6.653297 0.250977
%};
\addlegendentry{data-driven, $\const{R}=2$} %approach}

%\addplot [thick, mplblue, dashed]
%table {%
%	1.05734 1
%	1.34394279925933 0.909090909090909
%	1.58277846530877 0.833333z333333333
%	1.83548909725673 0.769230769230769
%	2.12833213863291 0.714285714285714
%	2.38212944115893 0.666666666666667
%	2.6042020808692 0.625
%	2.80014852767237 0.588235294117647
%	2.97432314705298 0.555555555555556
%	3.13016359597246 0.526315789473684
%	3.27042 0.5
%	3.54325714285714 0.476190476190476
%	3.79129090909091 0.454545454545455
%	4.01775652173913 0.434782608695652
%	4.22535 0.416666666666667
%	4.416336 0.4
%	4.59263076923077 0.384615384615385
%	4.75586666666667 0.37037037037037
%	4.90744285714286 0.357142857142857
%	5.04856551724138 0.344827586206897
%	5.18028 0.333333333333333
%	5.30349677419355 0.32258064516129
%	5.4190125 0.3125
%	5.52752727272727 0.303030303030303
%	5.62965882352941 0.294117647058824
%	5.72595428571429 0.285714285714286
%	5.8169 0.277777777777778
%	5.90292972972973 0.27027027027027
%	5.98443157894737 0.263157894736842
%	6.06175384615385 0.256410256410256
%	6.13521 0.25
%};
%\addlegendentry{theoretical approach}

\addplot [thick, mplblue, dashed]
table {
	0.918464 1.0
	2.97739  0.5
	4.39277  0.3333
	5.32161  0.25
	%0.923018 1.0
	%2.98608 0.5
	%4.39148 0.33333
	%5.33113 0.25
};
\addlegendentry{compression-based, $\const{R}=2$}% scheme}

\addplot [thick, mplblue, dotted]
table {%
  0.5625	1
0.63093025	0.975
0.70185587	0.95
0.77508607	0.925
0.85044671	0.9
0.927779	0.875
1.0069382	0.85
1.0877922	0.825
1.1702205	0.8
1.2541132	0.775
1.3393696	0.75
1.4258979	0.725
1.5136135	0.7
1.6024394	0.675
1.6923045	0.65
1.7831435	0.625
1.8748963	0.6
1.9675076	0.575
2.0609262	0.55
2.1551049	0.525
2.25	0.5
2.25	0.5
2.3158266	0.49166667
2.3816912	0.48333333
2.4475911	0.475
2.513524	0.46666667
2.579488	0.45833333
2.645481	0.45
2.7115013	0.44166667
2.7775473	0.43333333
2.8436175	0.425
2.9097105	0.41666667
2.9758249	0.40833333
3.0419597	0.4
3.1081137	0.39166667
3.1742859	0.38333333
3.2404754	0.375
3.3066812	0.36666667
3.3729026	0.35833333
3.4391387	0.35
3.5053888	0.34166667
3.5716524	0.33333333
3.5716524	0.33333333
3.61772	0.32916667
3.663836	0.325
3.7099977	0.32083333
3.756203	0.31666667
3.8024496	0.3125
3.8487357	0.30833333
3.8950592	0.30416667
3.9414184	0.3
3.9878116	0.29583333
4.0342374	0.29166667
4.0806943	0.2875
4.1271808	0.28333333
4.1736957	0.27916667
4.2202377	0.275
4.2668057	0.27083333
4.3133987	0.26666667
4.3600156	0.2625
4.4066553	0.25833333
4.4533171	0.25416667
4.5	0.25
	% 0.5625 1
	% 0.723749005130305 0.909090909090909
	% 0.892913091732112 0.833333333333333
	% 1.06658720363157 0.769230769230769
	% 1.24210070288304 0.714285714285714
	% 1.41741118113173 0.666666666666667
	% 1.59099025766973 0.625
	% 1.76171899632197 0.588235294117647
	% 1.92879896141941 0.555555555555556
	% 2.09167954876891 0.526315789473684
	% 2.25 0.5
	% 2.40354380433241 0.476190476190476
	% 2.55220317493979 0.454545454545455
	% 2.69595154466291 0.434782608695652
	% 2.83482236226346 0.416666666666667
	% 2.96889279923901 0.4
	% 3.09827126518711 0.384615384615385
	% 3.22308786843591 0.37037037037037
	% 3.34348715055813 0.357142857142857
	% 3.45962257483406 0.344827586206897
	% 3.57165236692845 0.333333333333333
	% 3.67973639767705 0.32258064516129
	% 3.78403386864172 0.3125
	% 3.88470161563437 0.303030303030303
	% 3.98189288742148 0.294117647058824
	% 4.07575648918758 0.285714285714286
	% 4.16643620529281 0.277777777777778
	% 4.25407043512514 0.27027027027027
	% 4.33879199074123 0.263157894736842
	% 4.42072801652069 0.256410256410256
	% 4.5 0.25
};
\addlegendentry{Shannon's, $\const{R}_\mathsf{inf}=2$}% scheme}

%%%%%%%%%%%%%%%%%%%%%%%%%%%%%%%%%%%%%%%%%%%%%%%%%%%%%%%%%%%%%%%%

%\addlegendimage{empty legend};
%\addlegendentry{%\\%
%  $\const{R}=4$};

\addplot [semithick, mplorange, mark=*, mark options={solid}]
table [x=distortion, y=accuracy, col sep=comma] {GaussianR4};
% table [x=distortion, y=accuracy, col sep=comma] {../simulation-results_Mark/GaussianR4};
%table {%
%	2.371161 1
%	2.716265 0.750977
%	4.388816 0.506836
%	6.653297 0.250977
%};
\addlegendentry{data-driven, $\const{R}=4$}% approach}

\addplot [thick, mplorange, dashed]
table {%
	0.0720998 1
	0.827231  0.5
	1.88223   0.3333
	2.83216   0.25
	%0.0803191 1
	%0.863182 0.5
	%1.94346 0.33
	%2.89764 0.25
};
\addlegendentry{compression-based, $\const{R}=4$}% scheme}

%\addplot [thick, mplorange, dashed]
%table {%
%	0.0855101 1
%	0.309544347518014 0.750977
%	1.02864192292402 0.506836
%	3.26000568148815 0.250977
%};
%\addlegendentry{quantization-based scheme}

\addplot [thick, mplorange, dotted]
table {%
0.03515625	1
0.044998746	0.975
0.056440832	0.95
0.06954871	0.925
0.084375	0.9
0.10095982	0.875
0.11933193	0.85
0.13950995	0.825
0.16150352	0.8
0.1853144	0.775
0.2109375	0.75
0.23836189	0.725
0.2675716	0.7
0.29854646	0.675
0.33126277	0.65
0.36569391	0.625
0.40181089	0.6
0.43958286	0.575
0.47897746	0.55
0.51996126	0.525
0.5625	0.5
0.5625	0.5
0.59886709	0.49166667
0.63605545	0.48333333
0.67403594	0.475
0.71278044	0.46666667
0.75226191	0.45833333
0.79245429	0.45
0.83333254	0.44166667
0.87487258	0.43333333
0.9170513	0.425
0.95984649	0.41666667
1.0032368	0.40833333
1.0472018	0.4
1.0917219	0.39166667
1.1367781	0.38333333
1.1823523	0.375
1.2284274	0.36666667
1.2749864	0.35833333
1.3220135	0.35
1.3694934	0.34166667
1.4174112	0.33333333
1.4174112	0.33333333
1.4576106	0.32916667
1.4979945	0.325
1.5385561	0.32083333
1.5792888	0.31666667
1.6201865	0.3125
1.6612433	0.30833333
1.7024535	0.30416667
1.7438118	0.3
1.7853129	0.29583333
1.8269522	0.29166667
1.8687247	0.2875
1.9106262	0.28333333
1.9526522	0.27916667
1.9947987	0.275
2.0370618	0.27083333
2.0794378	0.26666667
2.1219229	0.2625
2.1645139	0.25833333
2.2072073	0.25416667
2.25	0.25

	% 0.03515625 1
	% 0.0582014024919008 0.909090909090909
	% 0.0885881988207333 0.833333333333333
	% 0.126400918105623 0.769230769230769
	% 0.171423795122504 0.714285714285714
	% 0.223228272933028 0.666666666666667
	% 0.28125 0.625
	% 0.344850424666856 0.588235294117647
	% 0.413362825952512 0.555555555555556
	% 0.486124814970903 0.526315789473684
	% 0.5625 0.5
	% 0.641891424371632 0.476190476190476
	% 0.723749005130305 0.454545454545455
	% 0.807572747907816 0.434782608695652
	% 0.892913091732112 0.416666666666667
	% 0.97936938370814 0.4
	% 1.06658720363157 0.384615384615385
	% 1.15425504529542 0.37037037037037
	% 1.24210070288304 0.357142857142857
	% 1.32988759558905 0.344827586206897
	% 1.41741118113173 0.333333333333333
	% 1.50449555070992 0.32258064516129
	% 1.59099025766973 0.3125
	% 1.67676740472358 0.303030303030303
	% 1.76171899632197 0.294117647058824
	% 1.84575455101719 0.285714285714286
	% 1.92879896141941 0.277777777777778
	% 2.01079058522286 0.27027027027027
	% 2.09167954876891 0.263157894736842
	% 2.17142624400566 0.256410256410256
	% 2.25 0.25
};
\addlegendentry{Shannon's, $\const{R}_\mathsf{inf}=4$}% scheme};
\end{axis}

\end{tikzpicture}}}
\hfill
\subfloat[]{
  % \resizebox{0.33\textwidth}{0.3\textwidth}{
  \scalebox{0.9}{
    % This file was created by tikzplotlib v0.9.1.
\begin{tikzpicture}

\definecolor{mplblue}{rgb}{0.12156862745098,0.466666666666667,0.705882352941177}
\definecolor{mplorange}{rgb}{1,0.498039215686275,0.0549019607843137}
\definecolor{mplgreen}{rgb}{0.172549019607843,0.627450980392157,0.172549019607843}

\pgfplotsset{every tick label/.append style={font=\tiny}}

\pgfplotsset{
	/pgfplots/xlabel near ticks/.style={
		/pgfplots/every axis x label/.style={
			at={(ticklabel cs:0.5)},anchor=near ticklabel
		}
	},
	/pgfplots/ylabel near ticks/.style={
		/pgfplots/every axis y label/.style={
			at={(ticklabel cs:0.5)},rotate=90,anchor=near ticklabel}
	}
}

\begin{axis}[
legend style={cells={align=left},font=\small}, % required for multiline legend entries
tick label style={/pgf/number format/fixed},
width=0.5\textwidth,
height=0.3\textheight,
legend cell align={left},
legend style={fill opacity=0.8, draw opacity=1, text opacity=1, draw=white!80!black},
%tick align=outside,
%tick pos=both,
grid style={gray,opacity=0.5,dotted},
xlabel={Per-symbol squared error distortion},
xlabel near ticks,
xlabel style={font=\small},
xmajorgrids,
xmin=0.000, xmax=0.35,
xtick style={color=black},
%<<<<<<< HEAD
%ylabel={Leakage $\const{L}$},
%=======
%ylabel={Leakage $\const L$},
ylabel={Accuracy},
ylabel near ticks,
ylabel style={font=\small},
%>>>>>>> b97d8d44fc3ff11aab3ff5ce4b9d2b6554f83bbb
ymajorgrids,
ymin=0.08, ymax=1.02,
%ytick style={color=black}
]

%\addlegendimage{empty legend};
%\addlegendentry{$\const{R}=2$};

\addplot [semithick, mplblue, mark=*, mark options={solid}]
table {%
0.033812   1.0
0.06305125 0.506100
0.069303885 0.406100
0.08373307 0.3097
0.098677 0.2157
0.127022 0.1098
};
\addlegendentry{data-driven, $\const{R}=\nicefrac{1}{8}$}

\addplot [thick, mplblue, dashed]
table {
0.215 1.0
0.265 0.5
0.293 0.333
0.308 0.25
0.317 0.2
0.322 0.167
0.327 0.143
0.330 0.125
0.337 0.111
0.343 0.1
};
\addlegendentry{compression-based (U), $\const{R}=\nicefrac{1}{8}$}
%[0.215 0.265 0.293 0.308 0.317 0.322 0.327 0.33  0.337 0.343]

\addplot [thick, mplblue, dashdotted]
table {
	0.163 1.0
	0.204 0.5
	0.243 0.333
	0.264 0.25
	0.277 0.2
	0.286 0.167
	0.295 0.143
	0.302 0.125
	0.308 0.111
	0.312 0.1
};
\addlegendentry{compression-based (NU), $\const{R}=\nicefrac{1}{8}$}

\addplot [semithick, mplorange, mark=*, mark options={solid}]
%table [x=Distortion, y=Accuracy, col sep=comma] {./simulation-results_Mark/MNISTr05};
table {%
0.018732 1.00000
0.033018 0.505000
0.038432 0.409700
0.048860 0.306100
0.062314 0.215000
0.096884 0.107600
};
% old data:
% table {%
% 0.040750127 1.00000
% 0.044609535 0.6282
% 0.05332412 0.5396
% 0.08577931 0.2067
% 0.09441076 0.1061
% };
\addlegendentry{data-driven, $\const{R}=\nicefrac{1}{2}$}

\addplot [thick, mplorange, dashed]
table {
0.116 1.0
0.165 0.5
0.192 0.333
0.215 0.25
0.235 0.2
0.248 0.167
0.258 0.143
0.265 0.125
0.274 0.111
0.282 0.1
};
\addlegendentry{compression-based (U), $\const{R}=\nicefrac{1}{2}$}
%[0.116 0.165 0.192 0.215 0.235 0.248 0.258 0.265 0.274 0.282]

\addplot [thick, mplorange, dashdotted]
table {
0.071 1.0
0.104 0.5
0.143 0.333
0.163 0.25
0.179 0.2
0.190 0.167
0.198 0.143
0.204 0.125
0.217 0.111
0.227 0.1
};
\addlegendentry{compression-based (NU), $\const{R}=\nicefrac{1}{2}$}

\end{axis}

\end{tikzpicture}}}
\\
% \subfloat[]{
% \resizebox{0.33\width}{0.33\width}{
% \scalebox{0.45}{
% \input{figs/MNIST_reconstruction_ex.tikz}}}
\subfloat[]{
  \scalebox{0.9}{
    % \resizebox{0.33\textwidth}{0.3\textwidth}{
    % This file was created by tikzplotlib v0.9.1.
\begin{tikzpicture}

\definecolor{mplblue}{rgb}{0.12156862745098,0.466666666666667,0.705882352941177}
\definecolor{mplorange}{rgb}{1,0.498039215686275,0.0549019607843137}
\definecolor{mplgreen}{rgb}{0.172549019607843,0.627450980392157,0.172549019607843}

\pgfplotsset{every tick label/.append style={font=\tiny}}

\pgfplotsset{
	/pgfplots/xlabel near ticks/.style={
		/pgfplots/every axis x label/.style={
			at={(ticklabel cs:0.5)},anchor=near ticklabel
		}
	},
	/pgfplots/ylabel near ticks/.style={
		/pgfplots/every axis y label/.style={
			at={(ticklabel cs:0.5)},rotate=90,anchor=near ticklabel}
	}
}

\begin{axis}[
	legend style={cells={align=left},font=\small}, % required for multiline legend entries
	tick label style={/pgf/number format/fixed},
	width=0.5\textwidth,
	height=0.3\textheight,
	legend cell align={left},
	legend style={fill opacity=0.8, draw opacity=1, text opacity=1, draw=white!80!black},
	%tick align=outside,
	%tick pos=both,
	grid style={gray,opacity=0.5,dotted},
	xlabel={Per-symbol squared error distortion},
	xlabel near ticks,
	xlabel style={font=\small},
	xmajorgrids,
	xmin=0.03, xmax=0.12,
	xtick style={color=black},
	ylabel={Accuracy},
	ylabel near ticks,
	ylabel style={font=\small},
	ymajorgrids,
	ymin=0.08, ymax=1.02,
	%ytick style={color=black}
]

	\addplot [semithick, mplblue, mark=*, mark options={solid}]
	%table [x=distortion, y=accuracy, col sep=comma] {../simulation-results_Mark/GaussianR4};
	table {%
		0.045450684 1.0
		0.06480756 0.509000
		0.070240214 0.412000
		0.076252 0.3018
		0.084259 0.222
		0.108109    0.10480
	};
	\addlegendentry{data-driven, $\const{R}=\nicefrac{1}{8}$}% approach}
	
	\addplot [thick, mplblue, dashed]
	table {
		0.088 1
		0.11  0.5
		0.124 0.333333333333333
		0.135 0.25
		0.142 0.2
		0.15  0.166666666666667
		0.157 0.142857142857143
		0.181 0.125
		0.201 0.111111111111111
		0.218 0.1
%		0.05223818530533129 1.0
%		0.0711938278546713 0.5
%		0.0927862663518839 0.25
%		0.1151183439502435 0.125
	};
	\addlegendentry{compression-based, $\const{R}=\nicefrac{1}{8}$}% scheme}
	
	\addplot [semithick, mplorange, mark=*, mark options={solid}]
	%table [x=Distortion, y=Accuracy, col sep=comma] cifar10};
	table {%
	0.03520954 1.0
	0.05172279 0.501400
	0.056397907 0.396300
	0.063558854 0.303
	0.073629 0.2069
	0.090171 0.1047
	};
	\addlegendentry{data-driven, $\const{R}=\nicefrac{1}{4}$}

	\addplot [thick, mplorange, dashed]
	table {
		0.066 1
		0.088 0.5
		0.1   0.333333333333333
		0.11  0.25
		0.116 0.2
		0.124 0.166666666666667
		0.13  0.142857142857143
		0.135 0.125
		0.139 0.111111111111111
		0.142 0.1
%		0.040839095900615144 1.0
%		0.05223818530533129 0.5
%		0.07119382785467128 0.25
%		0.0927862663518839 0.125
	};
	\addlegendentry{compression-based, $\const{R}=\nicefrac{1}{4}$}
\end{axis}

\end{tikzpicture}}}
\hfill
\subfloat[]{
  \scalebox{0.9}{
    % \resizebox{0.33\textwidth}{0.3\textwidth}{
    % This file was created with tikzplotlib v0.10.1.
\begin{tikzpicture}

\definecolor{mplblue}{rgb}{0.12156862745098,0.466666666666667,0.705882352941177}
\definecolor{mplorange}{rgb}{1,0.498039215686275,0.0549019607843137}
\definecolor{mplgreen}{rgb}{0.172549019607843,0.627450980392157,0.172549019607843}

\definecolor{darkgray176}{RGB}{176,176,176}
\definecolor{lightgray204}{RGB}{204,204,204}
\definecolor{steelblue31119180}{RGB}{31,119,180}

\pgfplotsset{every tick label/.append style={font=\tiny}}

\pgfplotsset{
	/pgfplots/xlabel near ticks/.style={
		/pgfplots/every axis x label/.style={
			at={(ticklabel cs:0.5)},anchor=near ticklabel
		}
	},
	/pgfplots/ylabel near ticks/.style={
		/pgfplots/every axis y label/.style={
			at={(ticklabel cs:0.5)},rotate=90,anchor=near ticklabel}
	}
}

\begin{axis}[
	legend style={cells={align=left},font=\small}, % required for multiline legend entries
	tick label style={/pgf/number format/fixed},
	width=0.5\textwidth,
	height=0.3\textheight,
	legend cell align={left},
	legend style={fill opacity=0.8, draw opacity=1, text opacity=1, draw=white!80!black},
	%tick align=outside,
	%tick pos=both,
	grid style={gray,opacity=0.5,dotted},
	xlabel={Per-symbol squared error distortion},
	xlabel near ticks,
	xlabel style={font=\small},
	xmajorgrids,
	xmin=0.03, xmax=0.14,
	xtick style={color=black},
	ylabel={Accuracy},
	ylabel near ticks,
	ylabel style={font=\small},
	ymajorgrids,
	ymin=0.08, ymax=1.02,
]

\addplot [semithick, mplgreen, mark=*, mark options={solid}]
table {%
0.03989621810615063 1
0.058301795 0.499300
0.066923015 0.401944
0.07393489 0.309837
0.082486905 0.201506
0.09233253 0.099781
};
\addlegendentry{data-driven (image splitting), $\const{R}=\nicefrac{1}{8}$}

\addplot [semithick, mplblue, dashed]
table {%
0.079 1
0.098 0.5
0.109 0.333333333333333
0.119 0.25
0.125 0.2
0.131 0.166666666666667
0.135 0.142857142857143
0.139 0.125
0.142 0.111111111111111
0.144 0.1
%0.05 1
%0.066 0.5
%0.078 0.333333333333333
%0.084 0.25
%0.091 0.2
%0.096 0.166666666666667
%0.099 0.142857142857143
%0.102 0.125
%0.107 0.111111111111111
%0.111 0.1
};
\addlegendentry{compression-based, $\const{R}=\nicefrac{1}{8}$}

\addplot [semithick, mplorange, mark=*, mark options={solid}]
table {%
0.0318731 1
0.04564578 0.511525
0.048925053 0.408100
0.053736743 0.31595
0.06654087 0.205200
0.08022395 0.099550
};
\addlegendentry{data-driven, $\const{R}=\nicefrac{1}{4}$}

\addplot [semithick, black, mark=*, mark options={solid}]
table {%
0.031858117692172526 1.0
0.045678403 0.500231
0.0500329 0.401619
0.05774249 0.294925
0.06709902 0.200175
0.080573 0.099969
};
\addlegendentry{data-driven (image splitting), $\const{R}=\nicefrac{1}{4}$}

\addplot [semithick, mplorange, dashed]
table {%
0.062 1
0.079 0.5
0.089 0.333333333333333
0.098 0.25
0.103 0.2
0.109 0.166666666666667
0.115 0.142857142857143
0.119 0.125
0.122 0.111111111111111
0.125 0.1	
%0.037 1
%0.05 0.5
%0.061 0.333333333333333
%0.066 0.25
%0.073 0.2
%0.078 0.166666666666667
%0.081 0.142857142857143
%0.084 0.125
%0.088 0.111111111111111
%0.091 0.1
};
\addlegendentry{compression-based, $\const{R}=\nicefrac{1}{4}$}

\end{axis}

\end{tikzpicture}}}
 
\caption{Accuracy versus per-symbol squared error distortion for both the data-driven approach and the schemes from \cref{sec:theoretical_approach}. (a) Synthetic Gaussian dataset. (b) MNIST (U and NU stand for uniform and nonuniform quantization, respectively, cf. Appendix~\ref{app:mnist-compression-based}). (c) CIFAR-$10$. (d) LSUN.}
\label{fig:quant-schemes}

% \medskip
% \todo[inline]{Mark: update the points of LSUN dataset.}
\end{figure*}
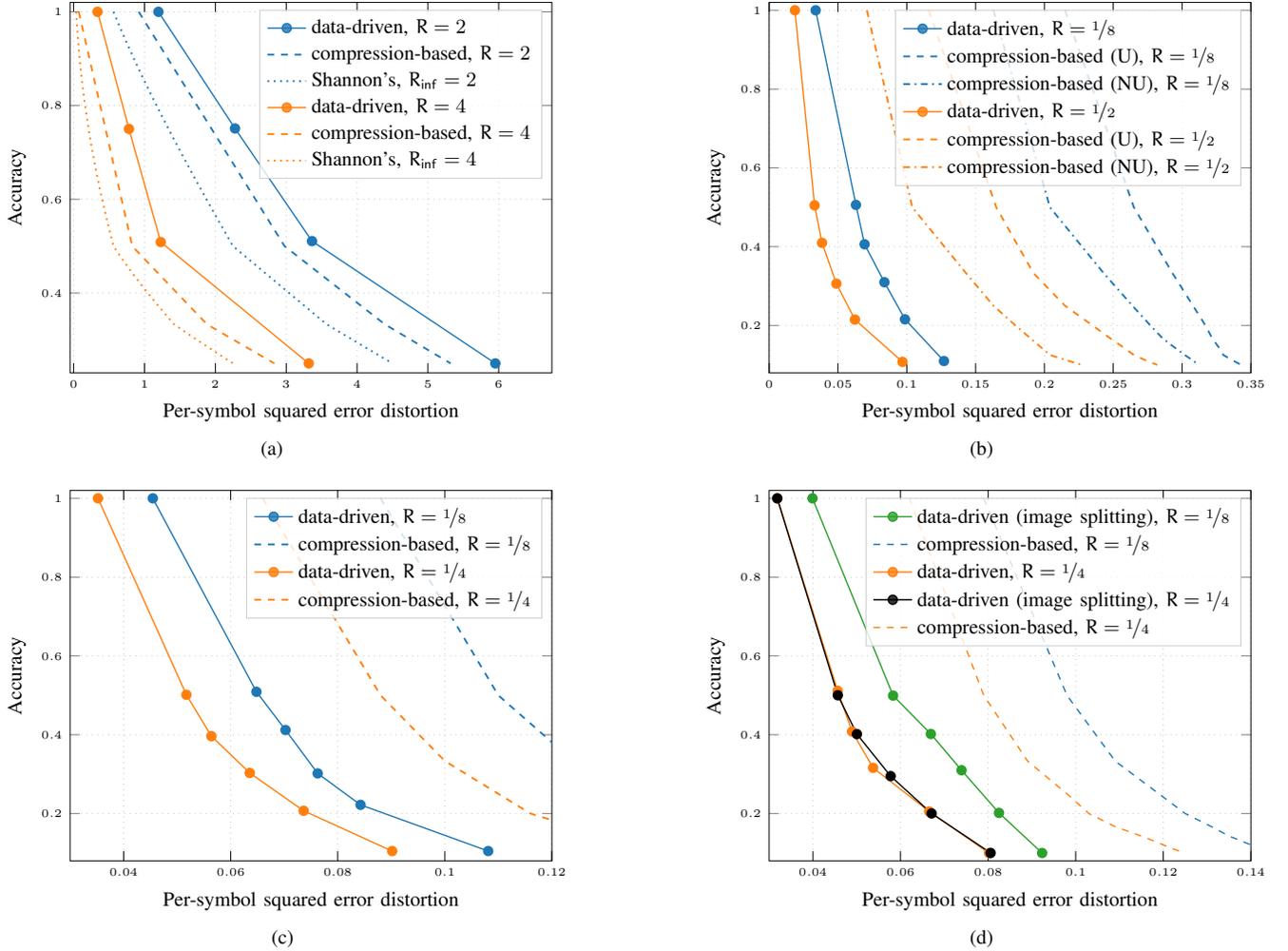

In \cref{fig:quant-schemes}, we plot the accuracy, or $\Pr(M=\hat{M})$, as function of per-symbol squared error distortion for different download rates $\const{R}$ for the data-driven approach (solid curves) and the compression-based scheme (dashed and dash-dotted curves). %The operational download rate $\const{R}_\mathsf{op}$ is defined as the ratio between number of downloaded bits and the number of symbols in a raw file from the dataset. 
For MNIST a symbol is a gray-scale pixel,  for CIFAR-$10$ and LSUN a pixel of each color, while for the Gaussian dataset it is a one-dimensional Gaussian RV.
% while for the Gaussian dataset it is defined as the ratio $\dots$.}
As a comparison, for the Gaussian dataset, we also plot the performance of Shannon's scheme  (dotted curves), assuming $\beta \to \infty$, as outlined in \cref{sec:theoretical_approach}. For Shannon's scheme, we use the \emph{information rate}, denoted by  $\const{R}_\mathsf{inf}$ and defined as $\eHPcond{\vect{A}}{\vect{Q}} / \beta$. It is well-known that the information rate is a true lower bound on the \emph{operational} rate from \eqref{eq:def_rate} (cf.~\cite{CoverThomas06_1}). %{\eirik Although $\const{R}_\mathsf{inf}$ is a lower bound on $\const{R}$, their numerical values are close, e.g., for the Gaussian dataset and  $\const{D}=3.36$, $\const{R}=2$ while $\const{R}_\mathsf{inf} \approx 1.99$.} % (see the appendix for a complete comparison).}

As expected, one can have a higher privacy level (i.e., smaller leakage) for a given distortion at the expense of a higher download rate. For the MNIST dataset, for both $\const{R}=\nicefrac{1}{2}$ and $\const{R}=\nicefrac{1}{8}$,  the data-driven approach significantly outperforms the compression-based scheme, while 
for the Gaussian dataset it performs close to the compression-based scheme using a variant of the generalized Lloyd algorithm~\cite{Lloyd82_1,LindeBuzoGray80_1} for the source code. This should not come as a surprise, as for the Gaussian dataset, the probabilistic model is simple and known precisely. In particular, we believe that the generalized Lloyd algorithm provides close-to-optimal compression (cf.~\cite{SabinGray86_1}). Note that as the exact operational rate for the data-driven curves could potentially be lower, the actual gap to Shannon's scheme could in fact be smaller. For MNIST we combined JPEG-like compression (including discrete-cosine transform), run-length encoding, and other entropy coding techniques for the compression-based scheme (cf.~\cite[Sec.~8.2]{Bocharova10_1}). However, due to a very small size of images, the constant overhead (e.g., for storing the Huffman codebook) turned out to be unacceptably high. It is a well-known phenomenon that sophisticated compression methods work well only on files of medium and large size. We thus opted for scalar quantization (see Appendix~\ref{app:mnist-compression-based} for details). \cref{fig:mnist-reconstruction-examples} depicts an example of the reconstructed digits for three levels of distortion $\const{D}$ (with respect to the original digits displayed in the figure) and accuracy $\const{L}$ for a download rate of $\const{R}=\nicefrac{1}{2}$ for the MNIST dataset. These reconstructions were obtained by the data-driven approach.  %\footnote{\eirik{Note that the distortion values are with respect to the original digits displayed in the figure and not averaged over the MNIST dataset.}}  % CIFAR-$10$.  %In the case of CIFA

%For CIFAR-$10$, as for MNIST, we have considered   two different download rates, $\const{R}=\nicefrac{1}{4}$ and $\nicefrac{1}{8}$.
For CIFAR-$10$,  we have considered the  two  download rates $\const{R}=\nicefrac{1}{4}$ and $\const{R}=\nicefrac{1}{8}$.
%In order to show the broader applicability of the data-driven approach, we have also applied it to CIFAR-$10$ with two different download rates, namely $\const{R}=\nicefrac{1}{4}$ and $\nicefrac{1}{8}$. 
The compression-based scheme for CIFAR-$10$ is rather similar to the scheme for MNIST (see Appendix~\ref{app:cifar10-compression-based}).
%\begin{wrapfigure}[12]{r}{.30\textwidth}
%	\begin{center}
%		\scalebox{0.45}{\input{figs/MNIST_reconstruction_ex.tikz}}
%	\end{center}
%	%\caption{MNIST, $\const{R}=\nicefrac 12$ bits per pixel.}
%	\label{fig:mnist-curves}	
%\end{wrapfigure}
The data-driven approach, again, outperforms the compression-based scheme for both rates considered with a gap similar to the one for MNIST.
%\sout{Compared to MNIST, the performance of the data-driven approach for CIFAR-$10$ is close to that of the compression-based scheme for  $\const{R}=\nicefrac{1}{4}$, while for  $\const{R}=\nicefrac{1}{8}$ it outperforms the compression-based scheme. Reasons for this might be  the fact that the number of training images for CIFAR-$10$ is lower and that the dimension of the images is around $4$ times larger than for MNIST. Hence, training is more difficult (due to the curse of dimensionality). Moreover, the images of CIFAR-$10$ are more structured (and with $3$ separate color channels) which may make the training harder.} }

As for CIFAR-$10$, we have considered $\const{R}=\nicefrac{1}{4}$ and  $\const{R}=\nicefrac{1}{8}$ for LSUN. Moreover, we have considered the same compression-based scheme as for  CIFAR-$10$ (see Appendix~\ref{app:cifar10-compression-based}).  For $\const{R}=\nicefrac{1}{4}$, as can be seen from \cref{fig:quant-schemes}(d), the data-driven approach  outperforms the compression-based scheme for all considered distortion constraints, with a performance gain that is comparable to that of the two previous image datasets. For comparisons, we have also shown the performance of the data-driven approach with image splitting  mentioned above, where the images are split  into $4$ separate smaller $32 \times 32$ pixels sub-images. The resulting $32 \times 32$ pixels sub-images, treated as independent images, are used to train the same neural network architecture  (see \cref{tab:img_datasets_arch} in Appendix~\ref{sec:Details_learning_alg} for further details) as for CIFAR-$10$. As can be seen from the figure, the splitting approach performs almost as well as training on the original images. This shows that image splitting is a powerful approach when having larger image sizes and also that the neural network architecture is robust across different classes of fixed-size images. For $\const{R}=\nicefrac{1}{8}$, the data-driven approach with image splitting outperforms the compression-based scheme, as for $\const{R}=\nicefrac{1}{4}$. We also note that the approach of splitting a $64 \times 64$ pixels image into $4$ separate $32 \times 32$ pixels sub-images does not change much for the compression-based scheme, as the scheme splits the images into blocks anyway. The only potential difference would be for extremely low rates (e.g., $1$ bit per whole image) when blocks of sizes larger than $32 \times 32$ pixels are required.

\begin{figure}[tb!]
  % \centering
  % \begin{minipage}{0.4\textwidth}
  \centering
  \scalebox{0.5}{\input{figs/MNIST_reconstruction_ex.tikz}}
  \caption{MNIST, $\const{R}=\nicefrac 12$ bits per pixel.}
  \label{fig:mnist-reconstruction-examples}
\end{figure}
% \end{minipage}\hfill
% \begin{minipage}{0.55\textwidth}
%   \centering
%   \end{minipage}
% \end{figure*}

So far only the download cost of the proposed schemes has been
  considered, neglecting both the query upload cost and the cost of
  distributing the trained neural networks. First, the cost of
  distributing the trained networks can be neglected since training is
  usually done beforehand on a dedicated training server and can hence
  be seen as a one-time cost. This cost vanishes as the protocol can  in principle run for a very long time serving a large number of users while the dataset grows continuously. Second, the query upload cost, i.e., $\HP{\vect{Q}}$ (in bits), is in most cases much smaller than the download cost of the answers. As we will show below, this is also the case here for MNIST, CIFAR-$10$, and LSUN.  For completeness, the one-time cost (in bits) of distributing the query and decoder  networks to the user is $465479 \times 32$, $1259135 \times 32$, $744882 \times 32$, and $9532703 \times 32$ bits for the Gaussian  ($\const{R}=2$), MNIST  ($\const{R}=\nicefrac{1}{2}$), CIFAR-$10$ ($\const{R}=\nicefrac{1}{8}$), and LSUN ($\const{R}=\nicefrac{1}{4}$) datasets, respectively, while the corresponding one-time cost of distributing the answer generation network to the server is $532230 \times 32$  ($\const{R}=2$), $7336888 \times 32$ ($\const{R}=\nicefrac{1}{2}$), $55871908 \times 32$ ($\const{R}=\nicefrac{1}{8}$),  and $47357034 \times 32$ ($\const{R}=\nicefrac{1}{4}$) bits, respectively. These numbers are the sizes of the corresponding neural networks as reported by TensorFlow times  $32$  (the neural networks' edge weights are represented as $32$-bits floats).
  
% \end{center}
%
%neglecting both the query upload cost and the cost of downloading the actual query and decoder networks to the user (assuming training is down at the server) is that in most cases this cost is much smaller than the download cost of the answers. As we will show below, this is also the case here. }
%
The number of neurons of the output layer of the query network is at most equal to the number of files $\const{M}$. Hence, the upload cost is at most $32 \cdot  \const{M}$ (assuming $32$-bits floats). For MNIST, this yields at most $32 \cdot 10 = 320$ bits, while the download cost is $28 \cdot 28 \cdot \nicefrac{1}{2} = 392$ bits (assuming $\const{R}= \nicefrac{1}{2}$), which is higher. Downloading the entire dataset requires an overall communication cost of $28
\cdot  28 \cdot 8 \cdot 10 = 62720$ bits uncompressed, which is significantly
higher. This number can be further reduced using lossless compression. The average losslessly compressed (with the LZMA algorithm \cite[Sec.~3.24]{Salomon07_1}) sizes for
different digits range from $1358$ (digit $1$) to $2151$ (digit $8$), which sum up to $19007$ bits for all $10$ digits. In other words, a one instance of the dataset ($10$ files) can be downloaded as $19007$ bits (on average), which is again significantly
higher than the download cost. Analogously, one instance of the CIFAR-$10$ dataset ($10$ files) has the size of $32 \cdot 32 \cdot 3 \cdot 8 \cdot 10 = 245760$ bits, but can be on average compressed into $197207$ bits, while one instance of the  LSUN dataset, with an original size of $64 \cdot 64 \cdot 3 \cdot 8 \cdot 10=983040$ bits, has a mean compressed size of $721726$ bits.
%For the Gaussian dataset, the upload cost is $32 \cdot 4 =
%128$ bits, while the download cost is $32 \cdot 3 \cdot \const{R}$
%bits where $\const{R}=2$ or $4$. 
In general, the upload cost scales
linearly with the number of files and is independent of the file size,
while the download cost increases with the file size. In most cases, the
file size  is much larger than the number of files which is the
 standard argument for not considering the upload cost. In the embedded table below we summarize the different costs (in bits) for the Gaussian, MNIST, CIFAR-$10$, and LSUN datasets and compare with the cost (in bits) of downloading the entire dataset using lossless source coding. The upload cost is at most equal to the number of neurons of the query network's output layer times $32$, with equality for a uniform query distribution.

  \begin{center}
 % \resizebox{0.5\textwidth}{1.75cm}{
   \scalebox{0.80}{
     \begin{tabular}{@{}lccc@{}}
       \toprule 
       & Download cost   & Upload cost  & Download dataset \\
       \midrule
       Gaussian & $6\, (2), 12\, (4)$ & $\leq 128$ &   $384$\\
%       MNIST  & $392\,  \bigl(\nicefrac{1}{2}\bigr)$   & $320$   & $17825$ \\
        MNIST  & $98\,  \bigl(\nicefrac{1}{8}\bigr)$, $392\, \bigl(\nicefrac{1}{2}\bigr)$  & $\leq 160$   & $19007$ \\
       CIFAR-$10$   & $384\, \bigl(\nicefrac{1}{8}\bigr)$, $768\, \bigl(\nicefrac{1}{4}\bigr)$  & $\leq 160$  & $197207$\\
%       CIFAR-$10$    &   & $160$  & $197208$\\
       LSUN &  $1536\, \bigl(\nicefrac{1}{8}\bigr)$, $3072\, \bigl(\nicefrac{1}{4}\bigr)$ & $\leq 160$ & $721726$ \\
       \bottomrule
     \end{tabular}}
 \end{center}
 % \end{wrapfigure}

\subsection{ML-Based Framework for Privacy Leakage}
\label{sec:ml-based-framework_privacy}
 
Since the derivative of $\Pr(M = \hat{M})$ is not well-defined, using an ML-based training approach to provide information-theoretical privacy guarantees has been substantially discussed in~\cite{HuangKairouzChenSankarRajagopal17_1,TripathyWangIshwar19_1,TsengWu20_1}. In this subsection, in order to assess the quality of training using an MI-based loss function, we tabulate in~\cref{tab:Monte-Carlo_Gaussianan,tab:Monte-Carlo_MNIST-CIFAR10} the estimates of the inference accuracy of a MAP adversary for $M$, denoted by $\const{L}_\mathsf{MAP}$, operating directly on the queries from the query network for different datasets and for different levels of per-symbol squared error distortion $\const{D}$. The results for the Gaussian dataset are tabulated in \cref{tab:Monte-Carlo_Gaussianan}, while those for the MNIST, CIFAR-$10$, and LSUN  datasets can be found in \cref{tab:Monte-Carlo_MNIST-CIFAR10}.\footnote{In the special case of full leakage, the problem reduces to the classical rate-distortion problem, and to improve performance, we only considered an encoder (for compression of the requested file),  corresponding to $f_{\textnormal{A}}$, and a decoder (for decompression at the user side), corresponding to $f_{\textnormal{\^{X}}}$. Hence, $\const{L}=\const{L}_\mathsf{MAP}=1$ in this special case.} 
The estimates have been found by Monte-Carlo simulation of the networks, i.e., by repeatedly generating queries and estimating the probability density functions of the RVs ${\vect{Q}\mid M}$. The distortion values are averaged over  the datasets as opposed to the distortion values reported in \cref{fig:mnist-reconstruction-examples}. For comparison we also show the inference accuracy $\const{L}$ of the adversary network in~\cref{tab:Monte-Carlo_Gaussianan,tab:Monte-Carlo_MNIST-CIFAR10}. As can be seen by comparing the second and third lines of the tables, the inference accuracy of the adversary network is close to the inference accuracy of a MAP adversary. These results indicate that the ML-based model $f_{\textnormal{\^{M}}}$ is trained quite well and that the proposed ML-based framework is able to provide meaningful privacy guarantees. %In addition, the entropy of the answer $\vect{A}$ is also estimated from the answer network in a similar way for the Gaussian dataset in~\cref{tab:Monte-Carlo_Gaussianan}. It can be seen that $\eHP{\vect{A}}$ is close to $\const{R} \cdot \beta$, which gives an indication that the performance of the answer neural network is close to being optimal. %an optimal bound. % the last statement need to be checked
 % For CIFAR-$10$ with accuracy $\const{L}=0.396$ and distortion $\const{D}=0.056$, the MAP accuracy is $\const{L}_{\mathsf{MAP}}=0.411$, which shows that also for CIFAR-$10$ the adversary is trained quite well.
 
\begin{table}[t!]
  \centering
  \caption{Estimated MAP adversary accuracy %and answer entropy 
    for the Gaussian dataset with different values of $(\const{R},\const{D},\const{L})$.}
  % \begin{center}
  \vspace{-2ex}
  \scalebox{1}{
      \begin{tabular}{@{}cccc@{}}
        \toprule 
        \multicolumn{4}{c}{Gaussian dataset and $\const{R} =2$} \\
        \midrule
        $\const{D}$ &  $5.95$ & $3.36$  & $2.28$  \\ %& $1.20$ \\
        $\const{L}_\mathsf{MAP}$ &  $0.25$ & $0.53$ &  $0.75$  \\%& $1.0$ \\
        $\const{L}$ &  $0.25$ & $0.51$  & $0.75$ \\  %& $1.0$\\
        % $\eHP{\vect{A}}$   &  $5.96$ & $5.97$  & $5.90$   & $5.88$ \\
        \midrule
        \multicolumn{4}{c}{Gaussian dataset and $\const{R}=4$} \\
        \midrule
        $\const{D}$ &  $3.32$ & $1.23$ & $0.78$  \\%& $0.34$ \\
        $\const{L}_\mathsf{MAP}$ & $0.25$ & $0.51$ &  $0.75$  \\%& $1.0$ \\
        $\const{L}$ & $0.25$ & $0.51$ & $0.75$  \\%& $1.0$\\
        % $\eHP{\vect{A}}$ & $11.95$  & $11.38$ &   $10.37$   & $10.22$ \\
        \bottomrule
      \end{tabular}}% \hspace{1cm}
  % \end{center}%
  \label{tab:Monte-Carlo_Gaussianan}
\end{table} 
 
%\begin{table}[htbp!]
%  \centering
%  \caption{Estimated MAP adversary accuracy and answer entropy for the Gaussian dataset with different values of $(\const{R},\const{D},\const{L})$.}
%  % \begin{center}
%  {\other \scalebox{1}{
%    \begin{tabular}{ccccc}
%      \toprule 
%      \multicolumn{5}{c}{Gaussian dataset and $\const{R} =2$} \\
%      \midrule
%      $\const{D}$ &  $6.60$ & $5.46$ & $4.35$  & $1.16$ \\
%      $\const{L}_\mathsf{MAP}$ &  $0.26$ & $0.38$ & $0.50$  & $1.00$ \\
%      $\const{L}$ &  $0.25$ & $0.37$ &  $0.50$  & $1.00$\\
%      $\eHP{\vect{A}}$   &  $5.83$ & $5.85$ & $5.90$   & $5.89$ \\
%      \bottomrule
%    \end{tabular}}}% \hspace{1cm}
%  % \end{center}%
%  \\[1mm]
%  % \begin{center}
%  {\other\scalebox{1}{
%    \begin{tabular}{ccccc}
%      \toprule 
%      \multicolumn{5}{c}{Gaussian dataset and $\const{R}=4$} \\
%      \midrule
%      $\const{D}$ &  $3.93$ & $1.15$ & $0.72$  & $0.21$ \\
%      $\const{L}_\mathsf{MAP}$ & $0.27$ & $0.53$ &  $0.76$  & $1.00$ \\
%      $\const{L}$ & $0.26$ & $0.51$ & $0.76$  & $1.00$\\
%      $\eHP{\vect{A}}$ & $11.82$  & $11.56$ &   $11.45$   & $10.86$ \\
%      \bottomrule
%     \end{tabular}}}% \hspace{1cm}
%   % \end{center}%
%  \label{tab:Monte-Carlo_Gaussianan}
%\end{table} 
 % , 6.6048975, 4.3540325, 4.079384 1.1623315
 % ,  0.2587646875, 0.50356515625, 0.6769415625, 1.0
 % , 0.2501, 0.50341, 0.67677 1.0
 % , 5.834754309165664, 5.902184310767963, 5.915056621846864 , 5.8940076978267255
 % 
 % As shown above, this  assumption is valid for MNIST and also for  more complex datasets like CIFAR-10 (since the image size is bigger than for MNIST).}

\begin{table}[htbp!]
  \centering
  \caption{Estimated MAP adversary accuracy for the MNIST, CIFAR-$10$, and LSUN datasets with different values of $(\const{R},\const{D},\const{L})$.}
  % \begin{center}
  \vspace{-2ex}
  \scalebox{1}{
    \begin{tabular}{@{}cccccc@{}}
      \toprule 
      \multicolumn{6}{c}{MNIST dataset and $\const{R}=\nicefrac{1}{2}$} \\
      \midrule
      $\const{D}$              & $0.097$ & $0.062$ & $0.049$ & $0.038$& $0.033$  \\ % & $0.019$ \\
      $\const{L}_\mathsf{MAP}$ & $0.113$ & $0.220$ & $0.315$ & $0.413$ & $0.509$  \\ %& $1.0$ \\
      $\const{L}$              & $0.108$ & $0.215$ & $0.306$ & $0.410$ & $0.505$  \\ %& $1.0$ \\
      \midrule
        \multicolumn{6}{c}{MNIST dataset and $\const{R}=\nicefrac{1}{8}$} \\
        \midrule
        % $\const{D}$              & $0.12707311$ & $0.0986624$ & $0.083711185$ & $0.06933266$ & $0.06306841$ & $0.033812$
        % \\
        % $\const{L}_\mathsf{MAP}$ & $0.1644205$ & $0.222759$  & $0.31947775$ & $0.41746925$ & $0.5151475$ & $1.0$
        % \\
        % $\const{L}$              & $0.109800$ & $0.215700$  & $0.309700$ & $0.406100$ & $0.5061$ & $1.0$
        % \\
        $\const{D}$              & $0.127$ & $0.099$ & $0.084$ & $0.070$ & $0.063$ \\ % & $0.034$
        %\\
        $\const{L}_\mathsf{MAP}$ & $0.131$ & $0.223$  & $0.320$ & $0.417$ & $0.515$ \\ %& $1.0$
        %\\
        $\const{L}$              & $0.110$ & $0.216$  & $0.310$ & $0.406$ & $0.506$ \\ %& $1.0$
        %\\
		\midrule
      \multicolumn{6}{c}{CIFAR-$10$ dataset and $\const{R}=\nicefrac{1}{4}$} \\
      \midrule
      $\const{D}$              & $0.090$ & $0.074$ & $0.064$ & $0.056$ & $0.052$ \\ %& $0.035$ \\
      $\const{L}_\mathsf{MAP}$ &$0.116$   & $0.207$ & $0.318$  & $0.411$  & $0.513$ \\ % & $1.0$ \\
      $\const{L}$              & $0.105$ & $0.207$ & $0.303$  & $0.396$  & $0.501$ \\ % & $1.0$\\
		\midrule
      \multicolumn{6}{c}{CIFAR-$10$ dataset and $\const{R}=\nicefrac{1}{8}$} \\
      \midrule
      $\const{D}$              & $0.108$ & $0.084$ & $0.076$ & $0.070$ & $0.065$ \\ %& $0.045$ \\
      $\const{L}_\mathsf{MAP}$ & $0.118$ & $0.232$  & $0.313$ & $0.427$ & $0.515$ \\ %& $1.0$ \\
      $\const{L}$              & $0.105$ & $0.222$  & $0.302$ & $0.412$ & $0.509$ \\ %&  $1.0$\\
		\midrule
      \multicolumn{6}{c}{LSUN dataset with $\const{R}=\nicefrac{1}{4}$} \\
      \midrule
      % $\const{D}$              & $0.11921329$ & $0.09450728$ & $0.08838331$ & $0.079583146$ & $0.075070225$ & $0.0453445$
      % \\
      % $\const{L}_\mathsf{MAP}$ & $0.146140175$ & $0.2168672$  & $0.3171594$ & $0.4160461$ & $0.52719725$ & $1.0$
      % \\
      % $\const{L}$              & $0.106200$ & $0.204800$  & $0.301025$ & $0.398850$ & $0.513225$ & $1.0$
      % \\revthree
      $\const{D}$              &  $0.080$ &  $0.067$ &  $0.054$ &  $0.049$ &  $0.046$ \\ %&  {\revthree$0.032$}
      %\\
      $\const{L}_\mathsf{MAP}$ & $0.100$ & $0.226$  & $0.325$ & $0.419$ & $0.527$ \\ %& {\revthree$1.0$}
      %\\
      $\const{L}$              & $0.100$ & $0.205$  & $0.316$ & $0.408$ & $0.512$ \\ %& {\revthree$1.0$}\\
      \midrule
            \multicolumn{6}{c}{LSUN dataset with $\const{R}=\nicefrac{1}{4}$, image splitting} \\
      \midrule
      $\const{D}$              &  $0.081$ &  $0.067$ &  $0.058$ &  $0.050$ &  $0.046$ \\ %&  {\revthree$0.032$}
      %\\
      $\const{L}_\mathsf{MAP}$ & $0.100$ & $0.220$  & $0.317$ & $0.415$ & $0.521$ \\ %& {\revthree$1.0$}
      %\\
      $\const{L}$              & $0.100$ & $0.200$  & $0.295$ & $0.402$ & $0.500$ \\ %& {\revthree$1.0$}
      %\\
       \midrule
             \multicolumn{6}{c}{LSUN dataset with $\const{R}=\nicefrac{1}{8}$, image splitting} \\
      \midrule
      $\const{D}$              &  $0.092$ &  $0.083$ &  $0.074$ &  $0.067$ &  $0.058$ \\ %&  {\revthree$0.040$}
      %\\
      $\const{L}_\mathsf{MAP}$ & $0.100$ & $0.216$  & $0.329$ & $0.416$ & $0.521$ \\ %&  {\revthree$1.0$}
      %\\
      $\const{L}$              & $0.100$ & $0.202$  & $0.310$ & $0.402$ & $0.499$ \\ %&  {\revthree$1.0$}
      %\\
      \bottomrule
    \end{tabular}}
  % \end{center}
  \label{tab:Monte-Carlo_MNIST-CIFAR10}
  % \medskip
  % \todo[inline]{Mark: Update the table entries for LSUN with images of size $64\times 64$.}
\end{table}

\subsection{Heat Map Representation of the Answer}
\label{sec:heatmap}

 In order to gain insight about the learned data-driven schemes, we consider CIFAR-$10$ and analyze the output of the \emph{first} network  of the answer generation function (see Appendix~\ref{sec:Details_learning_alg}) % (cf.~\cref{tab:tableCIFAR10arch} in the supplementary material)
 through a heat map. % that reflects the contribution of each stored image in the answer, for a given requested image (further details on the construction of the heat map are given in Appendix~\ref{sec:heatmap}).
 The heat map reflects the contribution of each stored image in the answer, for a given requested image (corresponding to a given row of the heat map). Denote by $\delta^{(1)}, \dotsc, \delta^{(\const{M})}$ the outputs from the $\const{M}$ neurons following the softmax activation functions of the first network of the answer generation function. The second network of the answer generation function extracts the features of the dataset, denoted by $\vect{Z}^{(1)}, \dotsc, \vect{Z}^{(\const{M})}$, which are fed as input to the third network of the answer generation function together with $\delta^{(1)}, \dotsc, \delta^{(\const{M})}$ and the queries (which are fed into the fourth layer). In the second layer (of the third network), the vector $\bigl( \delta^{(1)}, \dotsc, \delta^{(\const{M})} \bigr)$ is multiplied elementwise  with the feature vector $\bigl( \vect{Z}^{(1)}, \dotsc, \vect{Z}^{(\const{M})} \bigr)$, producing the vector $\bigl(\delta^{(1)} \vect{Z}^{(1)}, \dotsc,  \delta^{(\const{M})} \vect{Z}^{(\const{M})} \bigr)$, which is subsequently combined with the queries in the fifth layer  in order to produce the final answer. Now, if $\delta^{(m)}$, for some $m \in [\const{M}]$, is close to zero, then the answer will not depend much on the file indexed by $m$, and it follows that the heat map indeed reflects which files contribute the most to the generated answer.

\begin{figure}[tb!]
  \centering
  \scalebox{0.70}{% This file was created by tikzplotlib v0.9.5.
\begin{tikzpicture}

\begin{axis}[
colorbar,
colorbar style={ytick={0.1,0.2,0.3,0.4,0.5},yticklabels={0.1,0.2,0.3,0.4,0.5},ylabel={}},
colormap={mymap}{[1pt]
  rgb(0pt)=(0,0,0.5);
  rgb(22pt)=(0,0,1);
  rgb(25pt)=(0,0,1);
  rgb(68pt)=(0,0.86,1);
  rgb(70pt)=(0,0.9,0.967741935483871);
  rgb(75pt)=(0.0806451612903226,1,0.887096774193548);
  rgb(128pt)=(0.935483870967742,1,0.0322580645161291);
  rgb(130pt)=(0.967741935483871,0.962962962962963,0);
  rgb(132pt)=(1,0.925925925925926,0);
  rgb(178pt)=(1,0.0740740740740741,0);
  rgb(182pt)=(0.909090909090909,0,0);
  rgb(200pt)=(0.5,0,0)
},
point meta max=0.508195824623108,
point meta min=1.83008663533712e-06,
tick align=outside,
tick pos=both,
x grid style={white!69.0196078431373!black},
xlabel={\large Answer retrieve index},
xmin=0.5, xmax=10.5,
xtick distance=1,
xtick style={color=black},
xticklabel style = {rotate=45.0},
xlabel near ticks,
y dir=reverse,
y grid style={white!69.0196078431373!black},
ylabel={\large Requested file index, $M$},
ymin=0.5, ymax=10.5,
ytick distance=1,
ytick style={color=black}
]
\addplot graphics [includegraphics cmd=\pgfimage,xmin=0.5, xmax=10.5, ymin=10.5, ymax=0.5] {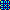};
\end{axis}

\end{tikzpicture}}
  \vspace{-2ex}
  \caption{Heat map with CIFAR-$10$ for leakage $\const{L}=0.303$, distortion $\const{D}=0.064$, and rate $\const R=\nicefrac 14$.}
  \label{fig:scheme-interpretation-diagram}	
\end{figure}

The heat map is shown in \cref{fig:scheme-interpretation-diagram} for $(\const{L},\const{D},\const R)=(0.303,0.064,\nicefrac{1}{4})$, %\ifthenelse{\boolean{short_version}}{,}{ and in Appendix~\ref{sec:another-heatmap} (see~\cref{fig:L04heatmap} for $(\const{L},\const{D},\const R)=(0.396,0.056,\nicefrac{1}{4})$,} 
 where each row corresponds to a requested image. The answer retrieve index  (the label of the $x$-axis) refers to the index of the softmax output values from the first network of the answer function, while the color reflects the actual softmax value in the sense that a warmer color  indicates a higher value.
As an example, consider rows $1$, $6$, and $10$ (corresponding to $M=1$, $6$, and $10$). According to the heat map, the corresponding answers are functions of the files $\vect X^{(1)}$, $\vect X^{(6)}$, and $\vect X^{(10)}$, meaning that the server can infer that the user is requesting one of these files, but not exactly which one, giving an accuracy of $\nicefrac {1}{3}$. %The same observation applies to the remaining file indices (or rows of the heatmap). %For instance, for $M \in \{2, 3, 5, 7, 9\}$, the answer is a function of the files $\{2, 3, 5, 7, 9\}$ and  the server has accuracy $\const L=\nicefrac {1}{5}$. For $M \in \{4, 8\}$, the answer depends only on the file $\{4, 8\}$ and thus the server has accuracy $\const L=\nicefrac {1}{2}$.
By looking at the remaining rows, the overall (average) accuracy becomes $\const{L}=\nicefrac{3}{10} \cdot \nicefrac 13 + \nicefrac{5}{10} \cdot \nicefrac 15 + \nicefrac{2}{10} \cdot \nicefrac 12 = \nicefrac{3}{10}$. 
Note that the scheme in this case is similar to the compression-based scheme from \cref{sec:theoretical_approach} and resembles time-sharing of three different schemes with $\const  L = \nicefrac 13$, $\nicefrac 15$, and $\nicefrac 12$, respectively. Also, averaging the expected distortion values of the schemes for a given requested index gives exactly $\const{D}=0.064$ (see \cref{lem:timesharing}). %However, the network may learn the file statistics to combine the file to get better distortion.}

\ifthenelse{\boolean{short_version}}{}{ %

In \cref{fig:L04heatmap}, the heat map with CIFAR-$10$ for $(\const{L},\const{D},\const R)=(0.396,0.056,\nicefrac{1}{4})$ is shown. As an example, consider rows $1$, $4$, and $8$ (corresponding to $M=1$, $4$, and $8$). According to the heat map, the corresponding answers are functions of the files $\vect X^{(1)}$, $\vect X^{(4)}$, and $\vect X^{(8)}$, meaning that the server can infer that the user is requesting one of these files, but not exactly which one, giving an accuracy of $\nicefrac {1}{3}$. For $M=2$ (corresponding to the second row), however, the answer depends only on the second file,  meaning that the server can infer that the user is requesting the second file, giving an accuracy of $1$.  
By looking at the remaining rows, the overall (average) accuracy becomes $\const{L}=\nicefrac{3}{10} \cdot \nicefrac 13 + \nicefrac{3}{10} \cdot \nicefrac 13 + \nicefrac{3}{10} \cdot \nicefrac 13  +\nicefrac{1}{10} \cdot 1 = \nicefrac{4}{10}$.  %However, the network may learn the file statistics to combine the file to get better distortion.}

%\begin{wrapfigure}[23]{r}{.6\textwidth}
\begin{figure}[tb!]
   \centering
  %\begin{center}
    % \input{figs/cifarA04resamp.tikz}
    \scalebox{0.7}{% This file was created by tikzplotlib v0.9.5.
\begin{tikzpicture}

\begin{axis}[
colorbar,
colorbar style={ytick={0.2,0.4,0.6,0.8},yticklabels={0.2,0.4,0.6,0.8},ylabel={}},
colormap={mymap}{[1pt]
  rgb(0pt)=(0,0,0.5);
  rgb(22pt)=(0,0,1);
  rgb(25pt)=(0,0,1);
  rgb(68pt)=(0,0.86,1);
  rgb(70pt)=(0,0.9,0.967741935483871);
  rgb(75pt)=(0.0806451612903226,1,0.887096774193548);
  rgb(128pt)=(0.935483870967742,1,0.0322580645161291);
  rgb(130pt)=(0.967741935483871,0.962962962962963,0);
  rgb(132pt)=(1,0.925925925925926,0);
  rgb(178pt)=(1,0.0740740740740741,0);
  rgb(182pt)=(0.909090909090909,0,0);
  rgb(200pt)=(0.5,0,0)
},
point meta max=0.993950191736221,
point meta min=2.19032319570545e-08,
tick align=outside,
tick pos=both,
x grid style={white!69.0196078431373!black},
xlabel={\large Answer retrieve index},
xmin=0.5, xmax=10.5,
xtick distance=1,
xtick style={color=black},
xticklabel style = {rotate=45.0},
xlabel near ticks,
y dir=reverse,
y grid style={white!69.0196078431373!black},
ylabel={\large Requested file index, $M$},
ymin=0.5, ymax=10.5,
ytick distance=1,
ytick style={color=black}
]
\addplot graphics [includegraphics cmd=\pgfimage,xmin=0.5, xmax=10.5, ymin=10.5, ymax=0.5] {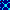};
\end{axis}

\end{tikzpicture}}
  %\end{center}
  \vspace{-2ex}
  \caption{Heat map with CIFAR-$10$ for leakage $\const{L}=0.396$, distortion $\const{D}=0.056$, and rate $\const R=\nicefrac 14$.}
  \label{fig:L04heatmap}
\end{figure}
}
%\end{wrapfigure}

%\begin{wrapfigure}[12]{r}{.30\textwidth}
%	\begin{center}
%		\scalebox{0.45}{\input{figs/MNIST_reconstruction_ex.tikz}}
%	\end{center}
%	%\caption{MNIST, $\const{R}=\nicefrac 12$ bits per pixel.}
%	\label{fig:mnist-curves}	
%\end{wrapfigure}

%\begin{wrapfigure}[12]{r}{.50\textwidth}
%	\begin{center}
%		\scalebox{0.66}{\input{figs/scheme-interpretation-diagram.tikz}}
%	\end{center}
%	\caption{}
%	\label{fig:scheme-interpretation-diagram}	
%\end{wrapfigure}

\subsection{Discussion on the Generative Adversarial Training}
\label{sec:discussion_GANs}

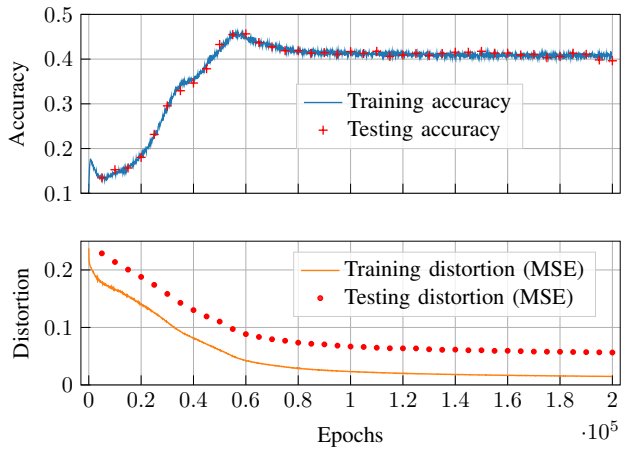
\begin{figure}[tb!]
  \centering
  % \begin{center}
  %   \input{figs/cifarA04resamp.tikz}
  \scalebox{0.85}{% This file was created by tikzplotlib v0.9.5.
\begin{tikzpicture}

\definecolor{color0}{rgb}{0.12156862745098,0.466666666666667,0.705882352941177}
\definecolor{color1}{rgb}{1,0.498039215686275,0.0549019607843137}

% let both axes use the same layers
\pgfplotsset{set layers}
%%%%%%%%%%%%%%%%%%%%%%%%%%%%%%%%%%%%%%
% ordinate (y-axis) for accuracies
%%%%%%%%%%%%%%%%%%%%%%%%%%%%%%%%%%%%%%
\begin{axis}[
	scale only axis, % used for having two ordinates
	xmin=-3000, xmax=203000,
        scaled x ticks = false,
        xticklabels=\empty,
	ylabel near ticks,yticklabel pos=left,
        y=7cm,
        % axis x line=none, % used for having two ordinates
%	tick align=outside,
	tick pos=left,
	x grid style={white!69.0196078431373!black},
	xmajorgrids,
	xtick style={color=black},
	y grid style={white!69.0196078431373!black},
	ymajorgrids,
	ymin=0.1,ymax=0.5,
	ytick style={color=black},
	yticklabel style={
		/pgf/number format/fixed%,
%		/pgf/number format/precision=5,
%		/pgf/number format/fixed zerofill
	},
	ylabel={Accuracy},
        legend cell align={left},
	legend style={fill opacity=0.8, draw opacity=1, text opacity=1, at={(0.82,0.43)}, anchor=east, draw=white!80!black},
]
	\addplot [semithick, color0]
	table {figs/cifarA04resamp_train_acc.data};\label{plt:train-acc}
	
	\addplot [semithick, red, mark=+, mark size=2, mark options={solid}, only marks]
	table {figs/cifarA04resamp_test_acc.data};\label{plt:test-acc}

        % legends for accuracies
	\addlegendimage{/pgfplots/refstyle=plt:train-acc}
	\addlegendentry{Training accuracy}

        \addlegendimage{/pgfplots/refstyle=plt:test-acc}
	\addlegendentry{Testing accuracy}
        
\end{axis}

% \end{tikzpicture}

% \begin{tikzpicture}

% \definecolor{color0}{rgb}{0.12156862745098,0.466666666666667,0.705882352941177}
% \definecolor{color1}{rgb}{1,0.498039215686275,0.0549019607843137}

% let both axes use the same layers
% \pgfplotsset{set layers}
%%%%%%%%%%%%%%%%%%%%%%%%%%%%%%%%%%%%%%
% ordinate (y-axis) for distortions
%%%%%%%%%%%%%%%%%%%%%%%%%%%%%%%%%%%%%%
\begin{axis}[
        scale only axis, % used for having two ordinates
        yshift=-3.0cm,
	xmin=-3000, xmax=203000,
	ylabel near ticks,yticklabel pos=left,
        y=9cm,
	% axis x line=none, % used for having two ordinates
	% tick align=outside,
	tick pos=left,
	x grid style={white!69.0196078431373!black},
	xlabel={Epochs},
	xmajorgrids,
	xtick style={color=black},
	y grid style={white!69.0196078431373!black},
	ymajorgrids,
	ymin=0,ymax=0.25,
	ytick style={color=black},
        %extra y ticks={0.25},
	% yticklabel style={
	% 	/pgf/number format/fixed%,
		% /pgf/number format/precision=5,
		% /pgf/number format/fixed zerofill
	% },
	ylabel={Distortion},	
	legend cell align={left},
	legend style={fill opacity=0.8, draw opacity=1, text opacity=1, at={(0.955,0.7)}, anchor=east, draw=white!80!black},
 ]
	
	\addplot [semithick, color1]
	table {figs/cifarA04resamp_train_dist.data};\label{plt:train-dist}	

	\addplot [semithick, red, mark=*, mark size=1, mark options={solid}, only marks]
	table {figs/cifarA04resamp_test_dist.data};\label{plt:test-dist}

	legends for distortions	
	\addlegendimage{/pgfplots/refstyle=plt:train-dist}
	\addlegendentry{Training distortion (MSE)}
	
	\addlegendimage{/pgfplots/refstyle=plt:test-dist}
	\addlegendentry{Testing distortion (MSE)}
\end{axis}

\end{tikzpicture}}
  % \end{center}
  \vspace*{2.25cm}
  \caption{Learning curves with CIFAR-$10$ for leakage $\const{L}=0.396$, distortion $\const{D}=0.056$, and rate $\const R=\nicefrac 14$.}
  \label{fig:acc04LearnCurve}
\end{figure}

To investigate the well-known overfitting problem in ML \cite{GoodfellowBengioCourville16_1}, we show some learning curves with the CIFAR-$10$ dataset (for leakage $\const{L}=0.396$, distortion $\const{D}=0.056$, and rate $\const R=\nicefrac 14$)  in Fig.~\ref{fig:acc04LearnCurve}. The training distortion and training adversary accuracy curves are averaged over a sliding window of length $100$ and then sampled every $100$ epoch. The testing distortion and testing adversary accuracy curves are sampled every $5000$ epoch. Note that the distortion is measured in terms of mean squared error (MSE). If overfitting happened, then the testing distortion would  increase while the training distortion would decrease with the number of epochs. This does not happen in our case. In fact, the testing distortion curve is decreasing and close to the training distortion curve. Hence, the level of overfitting is negligible. Moreover, training and testing accuracy values are close.

%\begin{wrapfigure}[23]{r}{.6\textwidth}
%\end{wrapfigure}

%However, our generator produces a query which is not real in world, and thus mode collapse is not a problem for our scheme because we do not aim to have diversity in the query generation.

% \begin{center}

The mode collapse issue in GANs is well-known~\cite{SalimanGoodfellow-etal16_1,Goodfellow17_1}. Mode collapse means that the generator finds some weak point of the discriminator and keeps producing this mode of output to trick the discriminator. This may happen when the generator tries to fool the discriminator trained on a real dataset and causes problems because a GAN model is typically required to have diversity in its output. 
Techniques such as minibatch discrimination~\cite{SalimanGoodfellow-etal16_1} can be applied to prevent mode collapse. However, note that in our setup mode collapse is less relevant  (and may even be beneficial in some cases)  as we in general do not need to have diversity in the query generation. In fact, a scheme for the MNIST (or CIFAR-$10$ or LSUN) dataset producing a deterministic query is a valid scheme with MAP adversary accuracy $\const{L}_{\mathsf{MAP}}=\nicefrac{1}{10}$, while a scheme giving a one-to-one correspondence between $M$ and $\vect{Q} \mid M$ is also valid and gives MAP adversary accuracy $\const{L}_{\mathsf{MAP}}=1$. %Moreover, from the heat maps in \cref{fig:scheme-interpretation-diagram,fig:L04heatmap}, it is apparent that there is not full diversity in the query generation, since in general all files do not contribute to the answer. This can be seen as some kind of mode collapse, but as mentioned above this is not a major concern (and may even be beneficial) in our step-up.} We remark that for intermediate values of the accuracy the behavior of the distribution of $\vect{Q} \mid M$ (for both MNIST and CIFAR-$10$) shows no signs of mode collapse as the number of epochs increases.}
In order to investigate  mode collapse in more detail for intermediate values of the accuracy,  we have estimated % the mean $\E[]{\vect{Q} \mid M}$ and
the variance $\Var{\vect{Q}\mid M}$ % of $\vect{Q} \mid M$
for CIFAR-$10$ for leakage $\const{L}=0.396$, distortion $\const{D}=0.056$, and rate $\const{R}= \nicefrac{1}{4}$. The estimated values are from $10^6$ samples and are tabulated in \cref{tab:query-variences_cifar10}. Here, $\vect{Q}\mid M$ is a $5$-dimensional vector since there are $5$ neurons in the output layer of the query generator network in~\ifthenelse{\boolean{short_version}}{\cite[Tab.~VI]{WengYakimenkaLinRosnesKliewer20_1sub}.}{\cref{tab:img_datasets_arch}.} % Table~\ref{tab:tableCIFAR10arch}.
As can be seen from the table, the query network generates queries with some diversity. Moreover, from the MAP accuracy values $\const{L}_{\mathsf{MAP}}$ in~\cref{tab:Monte-Carlo_MNIST-CIFAR10} (for CIFAR-$10$ with $(\const{L},\const{D},\const{R})=(0.396,0.056,\nicefrac{1}{4})$ it is $0.411$), it follows that the discriminator is trained quite well and do not appear to suffer from degeneration in the query generation  training due to mode collapse.

% Hence, in summary, we believe that out training does not suffer from degeneration.

\begin{table}[t!]
  \centering
  \caption{Query variances for a given requested file index for CIFAR-$10$ with $(\const{L},\const{D},\const{R})=(0.396,0.056,\nicefrac{1}{4})$.}
  \vspace{-2ex}
  \begin{tabular}{@{}cc@{}}    
    \toprule
    % Requested file index
    $M=m$ & $\Var{\vect{Q}|M=m}$\\
    \midrule
    % $1$ &  $(0.000, 0.026, 0.020, 0.000, 0.102)$\\
    % $2$ &  $(0.002, 0.037, 0.046, 0.000, 0.001)$\\
    % $3$ &  $(0.002, 0.036, 0.046, 0.000, 0.001)$\\
    % $4$ &  $(0.120, 0.041, 0.045, 0.148, 0.456)$\\
    % $5$ &  $(0.002, 0.038, 0.046, 0.000, 0.001)$\\
    % $6$ &  $(0.000, 0.025, 0.021, 0.000, 0.106)$\\
    % $7$ &  $(0.002, 0.037, 0.046, 0.000, 0.001)$\\
    % $8$ &  $(0.119, 0.040, 0.045, 0.150, 0.463)$\\
    % $9$ &  $(0.002, 0.037, 0.045, 0.000, 0.001)$\\
    % $10$ &  $(0.000, 0.025, 0.020, 0.000, 0.103)$\\   
    $1$ &  $(0.000, 0.000, 0.081, 0.036, 0.001)$\\
    $2$ &  $(0.049, 0.007, 0.024, 0.108, 0.000)$\\
    $3$ &  $(0.041, 0.040, 0.090, 0.004, 0.079)$\\
    $4$ &  $(0.000, 0.000, 0.081, 0.035, 0.001)$\\
    $5$ &  $(0.027, 0.041, 0.000, 0.023, 0.001)$\\
    $6$ &  $(0.027, 0.041, 0.000, 0.023, 0.001)$\\
    $7$ &  $(0.027, 0.041, 0.000, 0.023, 0.001)$\\
    $8$ &  $(0.000, 0.000, 0.080, 0.035, 0.001)$\\
    $9$ &  $(0.041, 0.040, 0.091, 0.005, 0.081)$\\
    $10$ &  $(0.040, 0.039, 0.091, 0.005, 0.081)$\\                    
    \bottomrule
  \end{tabular}
  \label{tab:query-variences_cifar10}
\end{table}
% \end{center}

\section{Conclusion}
\label{sec:conclusion}

% The conclusion goes here.
In this work, we have studied the trade-off between download rate, privacy leakage to the server, and reconstruction distortion at the user for single-server IR schemes. An IR scheme can be seen as an extension of the well-known concept of PIR by allowing for distortion in the retrieval process and relaxing the perfect privacy requirement. An information-theoretical formulation for the trade-off in terms of MI has  been provided in the limit of a large file size. We have shown that generative adversarial models can be successfully applied to design efficient single-server IR schemes. This is in particular beneficial if the data statistics is unknown. The main ingredient is a new optimization approach which combines GANs with additional constraints for the download rate and the desired reconstruction distortion at the user.  We have shown that our proposed GAN-based data-driven approach for a fixed download rate is able to provide a trade-off between distortion for the user and privacy leakage to the server that is close to that of a proposed compression-based scheme for Gaussian data where the source statistics is known. A similar trade-off as for the Gaussian case can be observed if the proposed data-driven approach  is applied to real-world  datasets like MNIST, CIFAR-$10$, and LSUN, for which it significantly outperforms the compression-based scheme. %{\jj \sout{For CIFAR-$10$ and LSUN, the performance of the data-driven approach is notably better than that of the compression-based scheme}.}

\appendices

\section{Proof of Lemma 1}
\label{sec:proof_Lemma1}

%\begin{lemma}
%	\label{lem:timesharing}
%	If the triples $(\const{R}_1,\const{D}_1,\const{L}_1)$ and $(\const{R}_2,\const{D}_2,\const{L}_2)$ are both achievable, then $(\lambda \const{R}_1 + (1-\lambda) \const{R}_2,\lambda \const{D}_1 + (1-\lambda) \const{D}_2,\lambda \const{L}_1 + (1-\lambda) \const{L}_2)$ is achievable, $0 \leq \lambda \leq 1$, for any loss function $f_{\textnormal{Loss}}$. % for which the expected loss $\mathsf{J}$ is concave.
%\end{lemma}
%\begin{proof}

  Assume the schemes that achieve the triples $(\const R_0, \const D_0, \const L_0)$ and $(\const R_1, \const D_1, \const L_1)$ are $\collect{C}_0$ and $\collect{C}_1$, respectively. We denote the queries of $\collect{C}_0$ and $\collect{C}_1$ as $\vect{Q}_0$ and $\vect{Q}_1$, respectively. From $\collect{C}_0$ and $\collect{C}_1$, we construct a new scheme $\collect{C}_{\lambda}$ that achieves the triple $(\lambda \const{R}_1 + (1-\lambda) \const{R}_0,\lambda \const{D}_1 + (1-\lambda) \const{D}_0,\lambda \const{L}_1 + (1-\lambda) \const{L}_0)$, for any $0 \leq \lambda \leq 1$. This is exactly the definition of a convex set. The scheme $\collect{C}_\lambda$ is as follows.
  \begin{itemize}
  \item The user generates a time-sharing RV $K$ according to the distribution
    \[
      P_K(k) = 
      \begin{cases}
        1-\lambda  & \text{if $k = 0$},\\
        \lambda & \text{if $k = 1$}.
      \end{cases}
    \]

  \item Then, the user forms the query $\vect{Q} = (K, \vect{Q}_K)$ and sends it to the server, thus explicitly notifying which of the schemes, $\collect{C}_0$ or $\collect{C}_1$, is used.
  
  \item Finally, the server replies according to $\collect{C}_K$ using the query $\vect{Q}_K$.\footnote{Formally, the scheme $\collect{C}_1$ is now using queries of the form $(1,\vect{Q}_1)$, but a constant prefix does not change the distribution and, thus, the leakage. The same holds for $\collect{C}_0$.}
\end{itemize}
  
The leakage of $\collect{C}_\lambda$ becomes
\begin{IEEEeqnarray*}{rCl}
    % \[
  \rho(P_{\vect Q | M})& = &\rho(\lambda P_{\vect Q_1 | M} + (1-\lambda) P_{\vect Q_0 | M})
  \\
  & \overset{(a)}{\le} &
  \lambda \rho( P_{\vect Q_1 | M} ) + (1-\lambda) \rho (P_{\vect Q_0 | M})
  \\
  & = &\lambda \const L_1 + (1-\lambda) \const L_0,
  % \]%
\end{IEEEeqnarray*}
where $(a)$ holds because of the convexity of $\rho$.

  % where equality in $(a)$ can be achieved by artificially ``worsening'' the scheme $\collect{C}_\lambda$, e.g., by explicitly sending $M$ to the server in some cases (with some probability). 
The proof for the distortion and the rate % of $\collect{C}$
follows in a similar manner (they are not just convex but also linear). Hence, the set of achievable rate-distortion-leakage triples constitutes a convex set.
% \end{proof}

\section{Proof of Theorem~\ref{thm:RDL-function}}
\label{sec:proof-theorem1} 

\textbf{Convexity.} We prove the convexity for $\const{R}(\const{D},\const{L})$ based on the assumption that $\rho(P_{\vect{Q}|M})$ is convex in $P_{\vect{Q}|M}$, i.e., we have
\begin{IEEEeqnarray*}{rCl}
  \IEEEeqnarraymulticol{3}{l}{%
    \const{L}_\lambda\eqdef\rho(\lambda P_{\vect{Q}_1|M}+(1-\lambda)P_{\vect{Q}_0|M})
  }\nonumber\\*% \quad%
  & \leq &\lambda\rho(P_{\vect{Q}_1|M})+(1-\lambda)\rho(P_{\vect{Q}_0|M})\leq\lambda\const{L}_1+(1-\lambda)\const{L}_0,
\end{IEEEeqnarray*}
for $0\leq\lambda\leq 1$, where $\rho(P_{\vect{Q}_0|M}) \leq \const{L}_0$ and $\rho(P_{\vect{Q}_1|M}) \leq \const{L}_1$. Note that since the queries should be generated without knowing any realizations of the retrieved files, it is quite natural to have $\bigMI{X^{[\const{M}]}}{\vect{Q}}=0$. Thus, the objective function in the minimization of \eqref{eq:RDL-function} can be expressed as
\begin{IEEEeqnarray*}{rCl}
  \bigMIcond{X^{[\const{M}]}}{\hat{X}^{[\const{M}]}}{\vect{Q}}& = &\bigMIcond{X^{[\const{M}]}}{\hat{X}^{[\const{M}]}}{\vect{Q}}+\underbrace{\bigMI{X^{[\const{M}]}}{\vect{Q}}}_{=0}
  \\
  & = &\bigMI{X^{[\const{M}]}}{\hat{X}^{[\const{M}]},\vect{Q}}.
\end{IEEEeqnarray*}
Now, let the distributions $P_{\hat{X}^{[\const{M}]}_1,\vect{Q}_1|X^{[\const{M}]}}$ and $P_{\hat{X}^{[\const{M}]}_0,\vect{Q}_0|X^{[\const{M}]}}$ achieve $\const{R}(\const{D}_1,\const{L}_1)$ and $\const{R}(\const{D}_0,\const{L}_0)$, respectively. Given $P_{\hat{X}^{[\const{M}]}_\lambda,\vect{Q}_\lambda|X^{[\const{M}]}}\eqdef\lambda P_{\hat{X}^{[\const{M}]}_1,\vect{Q}_1|X^{[\const{M}]}}+(1-\lambda)P_{\hat{X}^{[\const{M}]}_0,\vect{Q}_0|X^{[\const{M}]}}$ for any $0\leq \lambda\leq 1$, using the convexity of MI in $P_{\hat{X}^{[\const{M}]},\vect{Q}|X^{[\const{M}]}}$ we obtain
\begin{IEEEeqnarray}{rCl}
  \IEEEeqnarraymulticol{3}{l}{%
    \bigMI{X^{[\const{M}]}}{\hat{X}^{[\const{M}]}_\lambda,\vect{Q}_\lambda}}\nonumber\\*% \quad%
  & \leq &\lambda\bigMI{X^{[\const{M}]}}{\hat{X}^{[\const{M}]}_1,\vect{Q}_1}+(1-\lambda)\bigMI{X^{[\const{M}]}}{\hat{X}^{[\const{M}]}_0,\vect{Q}_0}.\IEEEeqnarraynumspace\label{eq:convexity_MI}
\end{IEEEeqnarray}
Further, one can also see that $P_{\vect{Q}_\lambda|M}=\lambda P_{\vect{Q}_1|M}+(1-\lambda)P_{\vect{Q}_0|M}$ and $P_{\hat{X}^{[\const{M}]}_\lambda | X^{[\const{M}]}}=\lambda P_{\hat{X}^{[\const{M}]}_1|X^{[\const{M}]}}+(1-\lambda)P_{\hat{X}^{[\const{M}]}_0|X^{[\const{M}]}}$. From the convexity of $\rho(\cdot)$ and the linearity of the distortion, it is straightforward to see that $\lambda\const{L}_1+(1-\lambda)\const{L}_0\geq \const{L}_\lambda$ and $\lambda\const{D}_1+(1-\lambda)\const{D}_0\geq \const{D}_\lambda$. Hence, from the definition of $\const{R}(\const{D},\const{L})$ we get
\begin{IEEEeqnarray*}{rCl}
  \IEEEeqnarraymulticol{3}{l}{%
    \const{R}(\lambda\const{D}_1+(1-\lambda)\const{D}_0,\lambda\const{L}_1+(1-\lambda)\const{L}_0)}\nonumber\\*\quad%
  & \stackrel{(a)}{\leq} &\const{R}(\const{D}_\lambda,\const{L}_\lambda)
  \leq\bigMI{X^{[\const{M}]}}{\hat{X}^{[\const{M}]}_\lambda,\vect{Q}_\lambda}
  \\
  & \stackrel{(b)}{\leq} &\lambda\const{R}(\const{D}_1,\const{L}_1)+(1-\lambda)\const{R}(\const{D}_0,\const{L}_0).
\end{IEEEeqnarray*}
where $(a)$ holds since the minimization is taken over a smaller constrained set, and $(b)$ follows from \eqref{eq:convexity_MI}. This then completes the proof of convexity of $\const{R}(\const{D},\const{L})$.

In the following, we show that for the case of memoryless vector sources $\vect{X}^{[\const{M}]}$, % i.e., each element of $\vect{X}^{(m)}$ is chosen independently and uniformly at random from $\mathbb{F}$,
the converse bound for the achievable rate $\const{R}$ is indeed $\const{R}(\const{D},\const{L})$.

Consider an i.i.d.\ sequence $\bigl\{X_i^{[\const{M}]}\bigr\}_{i=1}^{\beta}=\bigl\{X^{(1)}_i,\ldots,X^{(\const{M})}_i\bigr\}_{i=1}^{\beta}$, where each element $X_i^{[\const{M}]}$ is distributed according to a prototype PMF $P_{X^{(1)},\ldots,X^{(\const{M})}}$. The answer encoder $f_{\textnormal{A}}$ takes the input sequence $\bigl\{X^{[\const{M}]}_i\bigr\}_{i=1}^\beta$ and the generated query $\vect{Q}$ to construct the codewords that are indexed by $\set{A}=\{1,2,\ldots,2^{\beta\const{R}}\}$. The reconstruction decoder $f_{\textnormal{\^{X}}}$ outputs an estimate $\bigl\{\hat{X}^{[\const{M}]}_i\bigr\}_{i=1}^\beta$ for $\hat{X}^{[\const{M}]}_i=\bigl\{\hat{X}^{(m)}_i\bigr\}_{m=1}^{\const{M}}$ using the answer $\vect{A}$,\footnote{Here, without loss of generality, we assume that the reconstruction, denoted by  $\hat{\vect{X}}^{[\const{M}]}$, has the same dimensions as $\vect{X}^{[\const{M}]}$. However, the user can choose the desired file $\vect{X}^{(M)}$ to retrieve, i.e., the $M$-th entry of $\hat{\vect{X}}^{[\const{M}]}$.} requested index $M$, and the generated query $\vect{Q}$ at the user side. A feasible scheme should satisfy 
\begin{IEEEeqnarray*}{c}
  \frac{1}{\beta}\sum_{i=1}^{\beta}\E[M,\vect{Q},\vect{X}^{[\const{M}]}]{d_i\bigl(X^{(M)}_i,\hat{X}_i^{(M)}\bigr)}\leq\const{D},\,\rho\bigl(P_{\vect{Q}|M}\bigr)\leq \const{L}.\IEEEeqnarraynumspace
\end{IEEEeqnarray*}
Hence, using the fact that the average code length over a source code is bounded from below by the entropy of the source, e.g., see~\cite[Thm.~5.4.1]{CoverThomas06_1}, we have 
\begin{IEEEeqnarray*}{rCl}
  \beta\const{R}& \geq &\eHP{\vect{A}|\vect{Q}}\\
  &\stackrel{(a)}{=}&\eHP{\vect{A}|\vect{Q}}-\bigHP{\vect{A}|\bigl\{X^{[\const{M}]}_i\bigr\}_{i=1}^\beta,\vect{Q}}
  \\
  & = &\bigMIcond{\bigl\{X^{[\const{M}]}_i\bigr\}_{i=1}^\beta}{\vect{A}}{\vect{Q}}
  \\
  & \stackrel{(b)}{\geq} &\bigMIcond{\bigl\{X^{[\const{M}]}_i\bigr\}_{i=1}^\beta}{\bigl\{\hat{X}^{[\const{M}]}_i\bigr\}_{i=1}^\beta}{\vect{Q}}
  \\
  & \stackrel{(c)}{\geq} &\sum_{i=1}^\beta\bigHPcond{X^{[\const{M}]}_i}{\vect{Q}}-\sum_{i=1}^\beta\bigHPcond{X^{[\const{M}]}_i}{\hat{X}^{[\const{M}]}_i,\vect{Q}}
  \\
  & = &\sum_{i=1}^\beta\MIcond{X^{[\const{M}]}_i}{\hat{X}^{[\const{M}]}_i}{\vect{Q}},
  \\
  & \stackrel{(d)}{\geq} &\sum_{i=1}^\beta\const{R}\Bigl(\BigE[M,\vect{Q},\vect{X}^{[\const{M}]}]{d_i\bigl(X^{(M)}_i,\hat{X}_i^{(M)}\bigr)},\rho(P_{\vect{Q}|M})\Bigr)
  \\
  & \stackrel{(e)}{\geq} &\beta\const{R}\biggl(\frac{1}{\beta}\sum_{i=1}^\beta\BigE[M,\vect{Q},\vect{X}^{[\const{M}]}]{d_i\bigl(X^{(M)}_i,\hat{X}_i^{(M)}\bigr)},\rho(P_{\vect{Q}|M})\biggr)\IEEEeqnarraynumspace
  \\  
  & \stackrel{(f)}{\geq} &\beta\const{R}(\const{D},\const{L}),
\end{IEEEeqnarray*}
where $(a)$ holds since the answer $\vect{A}$ is a function of $\bigl\{X^{[\const{M}]}_i\bigr\}_{i=1}^\beta$ and $\vect{Q}$; $(b)$ follows by the data processing inequality; $(c)$ can be verified by using the chain rule of entropy and the fact that conditioning reduces entropy; $(d)$ is from the definition of $\const{R}(\const{D},\const{L})$ in~\eqref{eq:RDL-function} and the fact that all $d_i$ are equal; $(e)$ holds because $\const{R}(\const{D},\const{L})$ is convex; and $(f)$ is from the fact that $\const{R}(\const{D},\const{L})$ is nonincreasing and $\nicefrac{1}{\beta}\sum_{i=1}^{\beta} \bigE[M,\vect{Q},\vect{X}^{[\const{M}]}]{d_i\bigl(X_i^{(M)},\hat{X}_i^{(\const{M})}\bigr)}\leq\const{D}$, $\rho(P_{\vect{Q}|M})\leq\const{L}$. Hence, for any scheme that satisfies the distortion constraint $\bigE[M,\vect{Q},\vect{X}^{[\const{M}]}]{d\bigl(\vect{X}^{(M)},\hat{\vect{X}}^{(\const{M})}\bigr)}\leq\const{D}$ and the leakage constraint $\rho(P_{\vect{Q}|M})\leq\const{L}$, the rate should be larger than or equal to $\const{R}(\const{D},\const{L})$. This completes the converse proof.

\section{Achievable Schemes From Section~\ref{sec:theoretical_approach}}
\label{sec:achievable_schemes}

\subsection{Compression-Based Scheme for Gaussian Data}
\label{sec:compression-based-scheme_Gaussian}

As a reference for synthetic Gaussian data (see~\cref{fig:quant-schemes}(a)), we used the compression-based scheme outlined in Section~\ref{sec:theoretical_approach}. As explained in Section~\ref{sec:theoretical_approach}, the scheme reduces to quantization of random vectors drawn from an $\const N \beta$-dimensional multivariate Gaussian distribution for $\const N=1,2,\dotsc,\const M$ and different numbers of quantization vectors (i.e., different rates). We used the generalized Lloyd algorithm, also known as the Linde-Buzo-Gray (or LBG) algorithm~\cite{Lloyd82_1,LindeBuzoGray80_1}. Note that this algorithm is closely related to $k$-means clustering in unsupervised learning~\cite{Levrard18_1}.
The authors argue in \cite{LindeBuzoGray80_1} that although their approach is not guaranteed to provide the best quantizer, its results are nearly optimal for a wide class of distributions.

We briefly explain the steps of finding the quantization vectors (or levels) below and refer the interested reader to \cite{LindeBuzoGray80_1} for further details. When quantizing a Gaussian vector into $\const r$ bits, the number of quantization vectors is $\const k = 2^{\const r}$. For each $\const N$, we modify the original Gaussian training dataset with $\const n$ samples, described in \cref{sec_numerical_results}, into a dataset of size $\const m = \binom{\const M}{\const N}\const n$ by first shifting each file by its sample mean (to ensure the same distribution for each file) and then generating all $\const N$-subsets of $\const M$ files for each sample.\footnote{This shifting is without loss of generality since the distortion function $d(\cdot,\cdot)$ is translation invariant.} The resulting  dataset $\bigl\{ \vect z(l)=\bigl(\vect z^{(1)}(l),\ldots,\vect z^{(\const{N})}(l)\bigr)\bigr\}_{l=1}^{\const m}$ of $\const N \beta$-dimensional vectors is used to find quantization vectors as follows.
	%Given a large sample of $\const n$ vectors {\lin $\{ \vect{x}(l)\}_{l=1}^{\const n}$} from an $\const N \beta$-dimensional Gaussian dataset, one proceeds as follows.

\begin{enumerate}
%	\item Generate a large sample of $\const n$ vectors {\lin $\{ \vect{x}(l)\}_{l=1}^{\const n}$} from the desired {\eirik multivariate} Gaussian distribution.%, where $\const n$ denotes the number of sample vectors.
	\item Randomly select an initial set of quantization vectors $\bigl\{\vect q_{j} = \bigl( \vect q_j^{(1)},\ldots,  \vect q_j^{(\const N)}\bigr)\bigr\}_{j=1}^{\const k}$.  
	\item Continue iterating as follows.
	\begin{enumerate}
		\item Each sample vector $\vect{z}(l)$ is quantized to (approximated by) its closest quantization vector (its ``nearest neighbor'') $\vect{q}_{j^*(l)}$, where 
		%\[
			%d(\vect x^{(i)}, \vect q^{(j_i^*)}) = \min_j d(\vect x^{(i)}, \vect q^{(j)}).
			$j^*(l) = \argmin_j \nicefrac{1}{\const N} \sum_{i=1}^{\const N} d\bigl(\vect{z}^{(i)}(l), \vect{q}_j^{(i)}\bigr)$. %\footnote{\lin Here, with some abuse of notation, the distortion measure $d(\cdot,\cdot)$ is {\jj $\nicefrac{\sum_{i=1}^{\const N \beta} d_i(z_i,q_{i})}{\const N \beta}$}.}}
%				, where $h$ is the length of the input vector $\vect{x}$.}}
		%\]
		 This gives a mean quantization error (or distortion) among all the sample vectors of  
		 %\[
		 	$\const{D} = \nicefrac{1}{(\const{m} \cdot \const N)} \sum_{l=1}^{\const{m}}  \sum_{i=1}^{\const N} d\bigl(\vect{z}^{(i)}(l), \vect q_{j^*(l)}^{(i)}\bigr)$. 
		 %\]
		\item Each quantization vector is updated to the mean of the sample vectors as %follows,
		%\[
			$\vect q_{j} \leftarrow \nicefrac{1}{|\set Z_j|} \sum_{\vect z \in \set Z_j} \vect z$,
		%\]
		where the set of sample vectors that are quantized to $\vect q_{j}$ is denoted by $\set Z_j$ (``neighborhood'' of $\vect q_{j}$).
	\end{enumerate}
	\item The algorithm stops when the mean quantization error between two consecutive iterations changes less than a predefined threshold.
\end{enumerate}

Since the mean quantization error (or distortion) $\const{D}$ is
nonnegative and nonincreasing between iterations, the algorithm is
guaranteed to converge. Since it is a randomized algorithm, we ran it
multiple times and chose the quantizer that produced the smallest mean
quantization error. Obtained quantization vectors are tested on the test dataset, shifted by the sample mean of the training dataset. 
%Both uniformly picked random vectors (on some wide enough region) as
%well as randomly %picked vectors from the sample set were used for
%the initial set of quantization %vectors in Step 2.  Both methods
%gave basically the same results. 
As a final remark, we remind the reader that our goal here is not to find the best quantization possible, but only to have a good enough approximation.

Additionally, after the achievable points are calculated, we make sure they all satisfy the convexity property (cf.~\cref{lem:timesharing}). More precisely, if the triples $(\const R_0, \const D_0, \const L_0)$, $(\const R_1, \const D_1, \const L_1)$, and $(\lambda \const{R}_1 + (1-\lambda) \const{R}_0,
 \const D_2,\lambda \const{L}_1 + (1-\lambda) \const{L}_0)$ are achievable for some $0 < \lambda < 1$, we update
% \[
 $\const D_2 \leftarrow \min (\const D_2, \lambda \const D_1 + (1-\lambda) \const D_0)$.
% \]

\subsection{Shannon's Scheme for Gaussian Data}
\label{sec:shannon_GaussianData}

This section briefly outlines  Shannon's scheme for Gaussian data (plotted in \cref{fig:quant-schemes}(a)). Following its description in Section~\ref{sec:theoretical_approach}, the server compresses $\const{N}$, $\const{N}\in[\const{M}]$, independent files, each consisting of $\beta$ independent Gaussian RVs, together. It is known from the rate-distortion theory for multi-dimensional sources \cite{Gray73_1}, as the file size $\beta\to\infty$, that a download rate $\const{R}= \const{N} R_{\textnormal{G}}(\const{D})$ is achievable with distortion $\const{D}$. Here, $R_{\textnormal{G}}(\const{D})=\max\{\frac{1}{2}\log{\bigl(\nicefrac{\sigma^2}{\const{D}}\bigr)},0\}$ for a Gaussian RV distributed according to $\Gaussian{\mu}{\sigma^2}$ %, where $(\cdot)^+\eqdef\max\{\frac{1}{2}\log{\bigl(\nicefrac{\sigma^2}{\const{D}}\bigr)},0\}$~
\cite[Ch.~10]{CoverThomas06_1}. Similar to Appendix~\ref{sec:compression-based-scheme_Gaussian}, we further use a convexifying approach to obtain an achievable scheme for any hard decision leakage (or accuracy) $\nicefrac{1}{\const{M}} \leq \const{L} \leq 1$. In the following, we brief describe the scheme. First, we select two accuracy values, say $\const{L}_0$ and $\const{L}_1$, $\const{L}_0, \const{L}_1\in\{1,\nicefrac{1}{2},\ldots,\nicefrac{1}{\const{M}}\}$. Applying the Gaussian rate-distortion function of $R_{\textnormal{G}}(\const{D})=\frac{1}{2}\log{\bigl(\nicefrac{\sigma^2}{\const{D}}\bigr)}$, $\const{D}\in (0,\sigma^2]$, to the previously outlined scheme, one can obtain two achievable distortions $D_\textnormal{G}(\const{R}_0,\const{L}_0)=\sigma^2\cdot 2^{-2\const{R}_0\const{L}_0}$ and $D_\textnormal{G}(\const{R}_1,\const{L}_1)=\sigma^2\cdot 2^{-2\const{R}_1\const{L}_1}$. Next, the desired accuracy $\const{L}\in[\const{L}_0,\const{L}_1]$ is selected to be $\const{L}=\lambda\const{L}_1+(1-\lambda)\const{L}_0$, and the goal is to determine the minimum possible linear combination of $D_{\textnormal{G}}(\const{R}_0,\const{L}_0)$ and $D_{\textnormal{G}}(\const{R}_1,\const{L}_1)$ subject to $\const{R}=\lambda\const{R}_1+(1-\lambda)\const{R}_0$, i.e.,
\begin{IEEEeqnarray}{rCl}
  &&\min_{\substack{\inv{\const{L}}_0,\inv{\const{L}}_1\in[\const{M}]\\\const{L}=\lambda\const{L}_1+(1-\lambda)\const{L}_0}}\min_{\const{R}=\lambda\const{R}_1+(1-\lambda)\const{R}_0}\bigl[\lambda D_{\textnormal{G}}(\const{R}_1,\const{L}_1)\nonumber\\
  &&\hspace*{4.5cm} +\>(1-\lambda)D_{\textnormal{G}}(\const{R}_0,\const{L}_0) \bigr].\IEEEeqnarraynumspace\label{eq:Dcvx_RL}
\end{IEEEeqnarray}

  This then gives the performance of the synthetic Gaussian dataset in Fig.~\ref{fig:quant-schemes}(a). As a final remark, we also numerically evaluate the values of $\const{R}(\const{D},\const{L})$ for the Gaussian dataset. For instance, for $\const{D}=1.875$ and $\const{L}=0.6$, we obtain $\const{R}(\const{D},\const{L})\approx 2.035$, while~\eqref{eq:Dcvx_RL} gives $1.875$ for $(\const{R},\const{L})=(2.0,0.6)$, confirming 
 % 
  %that the performance of our proposed Shannon's scheme is optimal.
%
that our proposed Shannon's scheme performs very close to the information-theoretical optimum. %for the i.i.d.\ Gaussian case

\ifthenelse{\boolean{short_version}}{}{% 
\begin{table*}[tb!]
	\centering
	\caption{Neural network (NN) architectures used for training for the synthetic Gaussian dataset. FC stands for a fully connected layer. Size is the number of neurons or input size. As activation function we used the scaled exponential linear unit (SeLU)~\cite{Klambauer-etal17_1}, softmax, and sigmoid. We added a skip connection between the first layer and the fourth layer of the query generator network to improve the gradient propagation for the training~\cite{He-etal18_1}.}% There is a skip connection between the first layer and the fourth layer of the query generator network.
	\vspace{-2ex}
	\begin{tabular}{@{}lp{0.8\textwidth}@{}}
		\toprule
		NN & Detailed architecture (consecutive layers' sizes and types) \\
		\midrule
		Query generator, $f_{\textnormal{Q}}$   & $8$ (Input), $8$ (FC, SeLU), $8$ (FC, SeLU), $8$ (FC, SeLU), $8$ (FC, SeLU), $4$ (FC, SeLU)
		\\
		Answer generation, $f_{\textnormal{A}}$ & $4 + 3 \times 4 \times 1$ (Input), $16$ (Flatten), $256$ (FC, SeLU), $256$ (FC, SeLU), $256$ (FC, SeLU), $256$ (FC, SeLU), $256$ (FC, SeLU), $256$ (FC, SeLU),
		                                          $256$ (FC, SeLU), $256$ (FC, SeLU), $256$ (FC, SeLU), Answer dim (FC, sigmoid)
		\\
		Decoder, $f_{\textnormal{\^X}}$         & Answer dim $+\;8$ (Input), $256$ (FC, SeLU), $256$ (FC, SeLU), $256$ (FC, SeLU), $256$ (FC, SeLU), $256$ (FC, SeLU), $256$ (FC, SeLU), $256$ (FC, SeLU),
		                                          $256$ (FC, SeLU), $3$ (FC)
		\\
		Adversary, $f_{\textnormal{\^M}}$       & $4$ (Input), $64$ (FC, SeLU), $64$ (FC, SeLU), $64$ (FC, SeLU), $64$ (FC, SeLU), $64$ (FC, SeLU), $64$ (FC, SeLU), $4$ (FC, softmax)
		\\
		\bottomrule
	\end{tabular}
	\label{tab:table_GaussianNNarch}
\end{table*}

}

\subsection{Compression-Based Scheme for the MNIST Dataset}
\label{app:mnist-compression-based}

The scheme is based on the ideas of \cref{sec:theoretical_approach} and reduces to choosing a lossy compression method for a single image (a single compressor for all digits is considered).  %We remark here that we did not try to find an optimal scheme but some good enough one in order to have a benchmark.
First, we applied a grayscale compression similar to the instruments used by the JPEG standard. We refer the interested reader to \cite[Sec.~8.2]{Bocharova10_1} for more details and other techniques for image compression. %Here, we outline  ideas we tried.
The main ingredient of the JPEG standard is a two-dimensional discrete cosine transform. In the standard, an image is split into $8 \times 8$ blocks and the transform is applied to each block independently. Since the images from the MNIST dataset are of size $28 \times 28$, we applied the transform to $7 \times 7$ blocks. In this way, an image is split into $16$ blocks. Next, the resulting values were quantized by a uniform scalar quantizer. For each block, the top-left coefficients play the most important role as they grasp the low-frequency contents of the block. Thus, they are  assigned more bits from the available pool of bits (defined by the desired rate). Finally, these quantized coefficients were encoded with a combination of run-length encoding and a two-dimensional  Huffman code.

Since the MNIST images have a small size, the entropy coding techniques do not provide any compression benefits. For example, the usually ignored overhead of storing the Huffman coding table requires too many bits in our case. Also, the run-length encoder performs poorly as there are not many repetitive values,  increasing the size of an MNIST image on many occasions. Therefore, we turned to a much simpler scalar quantization of the original grayscale images.% where each image is split into blocks. For each block, the average of values of its pixels was quantized and stored. This method also allows for easier control of the desired rate.

Let us describe this method in more details. Assume we split a $28 \times 28$ pixels image into rectangular blocks of the same size $h \times w$ each and all values in each block are substituted by the mean value of the block forming one large ``pixel''. If we allocate $\const{r}$ bits to storing these mean values, each of them can be quantized to $2^\const{r}$ quantization points on $[0, 255]$ either uniformly (U) or nonuniformly (NU) spread. Such a scheme requires
	$
		\nicefrac{28}{h} \cdot \nicefrac{28}{w} \cdot \const{r}
	$
	bits for encoding one image and thus has a rate of $\nicefrac{\const{r}}{h w}$ bits per pixel. The distortion of the scheme is explicitly calculated from compressing and decompressing the training subset of the MNIST dataset. %We exhaustively checked all the choices of $h$, $w$, $\const{r}$, and the locations of the quantization points (in the NU case) for each target value of the required rate in order to find the best ones. The obtained sets of parameters were further tested on the test subset of the MNIST dataset.
We exhaustively checked all  choices of $h$, $w$, $\const{r}$, and for each choice optimized the locations of the quantization points (in the NU case) using the generalized Lloyd algorithm in order to find the best ones for each target value of the required rate. The obtained sets of parameters were further tested on the test subset of the MNIST dataset.

% \begin{comment}
\ifthenelse{\boolean{short_version}}{}{% 
}
% \end{comment}

\subsection{Compression-Based Scheme for the CIFAR-$10$ and LSUN Datasets}
\label{app:cifar10-compression-based}

The compression-based scheme for CIFAR-$10$ and LSUN is rather similar to the scheme for MNIST,  considering a single compressor for all classes compressing a single image. The only difference being a standard image preprocessing stage that is very common when compressing color images. First, the image is transformed from red, green, and blue channel representation to luminance-chrominance representation, consisting of brightness, hue, and saturation channels. Next, the hue and saturation channels are decimated by a factor of $2$. In other words, four neighboring pixels that form a $2 \times 2$ block are described by four values of brightness, one value of hue, and one value of saturation. These channels are further quantized in a  similar manner as for MNIST. However, nonuniform quantization showed to give no improvements for the CIFAR-$10$ and LSUN datasets (and often is actually worse than uniform quantization). We believe this is because of a much more diverse space of images in these datasets (as opposed to MNIST) and thus, fitting quantization points to the training datasets' block distributions has a negative effect on the rate-distortion curves, when calculated on the test subsets of the datasets. We thus do not present the results of nonuniform quantization.

% \begin{comment}
\ifthenelse{\boolean{short_version}}{}{% 
\begin{table*}[htbp!]
	\centering
	\caption{Neural network architectures used for training for the image datasets. The input pixel value is rescaled between $-1.0$ and $1.0$. FC stands for a fully connected layer. Conv and ConvT stand for a convolutional and a transposed convolutional layer, respectively, and ``st'' is shorthand for stride. Size is the number of neurons or input size. As activation function we used SeLU, hyperbolic tangent (tanh), softmax, and sigmoid. We added a skip connection between the first layer and the fourth layer of the query generator network to improve the gradient propagation for the training~\cite{He-etal18_1}.}% There is a skip connection between the first layer and the fourth layer of the query generator network.
%	\vspace{-5mm}
\vspace{-2ex}
	\begin{tabular}{@{}lp{0.8\textwidth}@{}}
		\toprule
		NN & Detailed architecture (consecutive layers' sizes and types) \\
		\midrule
		\multicolumn{2}{c}{Same for all datasets}\\
		\midrule
		Query generator, $f_{\textnormal{Q}}$            & $20$ (Input), $20$ (FC, SeLU), $20$ (FC, SeLU), $20$ (FC, SeLU), $9$ (FC, SeLU), $7$ (FC, SeLU), $5$ (FC, SeLU) 
		\\
		Answer generation part $1$, $f_{\textnormal{A}}$ & $5$ (Input), $5$ (FC, SeLU), $7$ (FC, SeLU), $9$ (FC, SeLU), $10$ (FC, softmax)
		\\
		Adversary, $f_{\textnormal{\^M}}$                & $5$ (Input), $64$ (FC, SeLU), $64$ (FC, SeLU), $64$ (FC, SeLU), $64$ (FC, SeLU), $64$ (FC, SeLU), $64$ (FC, SeLU), $10$ (FC, softmax)
		\\
		\midrule
		\multicolumn{2}{c}{MNIST dataset}\\
		\midrule
		Answer generation part $2$, $f_{\textnormal{A}}$ & $10 \times 28 \times 28 \times 1$ (Input), $10 \times 26 \times 26 \times 8$ (Conv, SeLU), 
		                                                   $10 \times 12 \times 12 \times 8$ (Conv ($\textnormal{st}=2$), SeLU) , $10 \times 10 \times 10 \times 16$ (Conv, SeLU),
		                                                   $10 \times 4 \times 4 \times 16$ (Conv ($\textnormal{st}=2$), SeLU), $256 \times 10$ (Concatenate $10$ parallel outputs of previous layers)
		\\
		Answer generation part $3$, $f_{\textnormal{A}}$ & $10 + 256 \times 10$ (Input), $256 \times 10$ (Broadcast and elementwise multiply), $2560$ (Flatten), $5$ (Input), $2048$ (FC, SeLU),
		                                                   $1024$ (FC, SeLU), Answer dim (FC, sigmoid)
		\\
		Decoder, $f_{\textnormal{\^X}}$                  & Answer dim $+\;20$ (Input), $512$ (FC, SeLU), $512$ (FC, SeLU), $512$ (FC, SeLU), $4 \times 4 \times 32$ (Unflatten), 
		                                                   $11 \times 11 \times 64$ (ConvT ($\textnormal{st}=2$), SeLU), $25 \times 25 \times 128$ (ConvT ($\textnormal{st}=2$), SeLU),
		                                                   $28 \times 28 \times 1$ (ConvT ($\textnormal{st}=2$), tanh)
		\\
		\midrule
		\multicolumn{2}{c}{CIFAR-$10$ and LSUN datasets, $32 \times 32$ pixels  images}\\
		\midrule
		Answer generation part $2$, $f_{\textnormal{A}}$ & $10 \times 32 \times 32 \times 3$ (Input), $10 \times 30 \times 30 \times 8$ (Conv, SeLU), 
		                                                   $10 \times 14 \times 14 \times 8$ (Conv ($\textnormal{st}=2$), SeLU), $10 \times 12 \times 12 \times 16$ (Conv, SeLU), 
		                                                   $2304 \times 10$ (Concatenate $10$ parallel outputs of previous layers)
		\\
		Answer generation part $3$, $f_{\textnormal{A}}$ & $10 + 2304 \times 10$ (Input), $2304 \times 10$ (Broadcast and elementwise multiply), $23040$ (Flatten), $5$ (Input), $2304$ (FC, SeLU),     
		                                                   $1024$ (FC, SeLU), Answer dim (FC, sigmoid)
		\\
		Decoder, $f_{\textnormal{\^X}}$                  & Answer dim $+\;20$ (Input), $512$ (FC, SeLU), $512$ (FC, SeLU), $512$ (FC, SeLU), $4 \times 4 \times 32$ (Unflatten), 
		                                                   $8 \times 8 \times 16$ (ConvT ($\textnormal{st}=2$), SeLU), $16 \times 16 \times 8$ (ConvT ($\textnormal{st}=2$), SeLU),
		                                                   $32 \times 32 \times 3$ (ConvT ($\textnormal{st}=2$), tanh)
		\\
		\midrule
		\multicolumn{2}{c}{LSUN dataset, $64 \times 64$ pixels images}\\
		\midrule
          Answer generation part $2$, $f_{\textnormal{A}}$ & \leavevmode $10 \times 64 \times 64 \times 3$ (Input), $10 \times 62 \times 62 \times 128$ (Conv, SeLU), $10 \times 60 \times 60 \times 128$ (Conv, SeLU), $10 \times 29 \times 29 \times 128$ (Conv ($\textnormal{st}=2$), SeLU), $10 \times 27 \times 27 \times 256$ (Conv, SeLU), $10 \times 25 \times 25 \times 256$ (Conv, SeLU), $10 \times 12 \times 12 \times 256$ (Conv ($\textnormal{st}=2$), SeLU), $10 \times 10 \times 10 \times 512$ (Conv, SeLU), $10 \times 4 \times 4 \times 96$ (Conv ($\textnormal{st}=2$), SeLU), $1536 \times 10$ (Concatenate $10$ parallel outputs of previous layers)
          \\
		Answer generation part $3$, $f_{\textnormal{A}}$ &\leavevmode $10 + 1536 \times 10$ (Input), $1536 \times 10$ (Broadcast and elementwise multiply), $1536 \times 2$ ($1 \times 1$ Conv), 
		                                                   $16 \times 96 \times 2$ (Permutation and reshape), $3072$ (Flatten), $5$ (Input), $3072$ (FC, SeLU), $1024$ (FC, SeLU), 
		                                                   $768$ (FC, SeLU), $768 \times 1$ (Reshape), $768 \times 26$ ($1 \times 1$ Conv, SeLU), $768 \times 4$ ($1 \times 1$ Conv, sigmoid),
		                                                   Answer dim (Flatten)
		\\
		Decoder, $f_{\textnormal{\^X}}$                  &\leavevmode Answer dim (Input), $768 \times 4$ (Reshape), $768 \times 1$ ($1 \times 1$ Conv, SeLU), $768$ (Flatten), $20$ (Input), $768$ (FC, SeLU),
		                                                   $4 \times 4 \times 48$ (Unflatten), $10 \times 10 \times 1024$ (ConvT ($\textnormal{st}=2$), SeLU), $8 \times 8 \times 512$ (Conv, SeLU), 
		                                                   $18 \times 18 \times 256$ (ConvT ($\textnormal{st}=2$), SeLU), $16 \times 16 \times 256$ (Conv, SeLU), 
		                                                   $34 \times 34 \times 128$ (ConvT ($\textnormal{st}=2$), SeLU), $70 \times 70 \times 64$ (ConvT ($\textnormal{st}=2$), SeLU),
		                                                   $68 \times 68 \times 64$ (Conv, SeLU), $66 \times 66 \times 64$ (Conv, SeLU), $64 \times 64 \times 3$ (ConvT ($\textnormal{st}=2$), tanh)
		\\
		\bottomrule
	\end{tabular}
	\label{tab:img_datasets_arch}
\end{table*}

} 
% \end{comment}

\section{{Learning for the Data-Driven Approach}}
\label{sec:Details_learning_alg}

\ifthenelse{\boolean{short_version}}{}{% 
}
 
As elaborated in \cref{sec:learning}, the learning is done by implementing the iterative algorithm of Algorithm~\ref{alg:training}. % first fixing the download rate $\const{R}$ and then solving the minimax optimization problem in \eqref{eq:training}.
Here,
% \footnote{Also, see, F.~Mentzer,  E.~Agustsson, M.~Tschannen, R.~Timofte, and L.~Van Gool, ``Conditional probability models for deep image compression,'' In Proceedings of the IEEE Conference on Computer Vision and Pattern Recognition (CVPR), 2018.}
the learning rate is selected empirically for each accuracy and distortion level with the RMSprop optimizer, and the number of iterations $\const{T}$ is selected empirically and individually for each dataset and training point. 
 %is $\const{T}=100000$. 
 The number of training samples, denoted by $\const{n}$, for the Gaussian case is set to $1000000$ ($250000$ for each of the $\const M = 4$ files) and to $60000$ ($6000$ for each digit), $50000$ ($5000$ for each class), and $1200000$ ($120000$ for each class) for the MNIST,  CIFAR-$10$, and LSUN datasets, respectively. For the image datasets, we used a minibatch size of $\const{b}=32$, while for the synthetic Gaussian dataset, we used a minibatch size of $\const{b}=2048$.

The architectures of the deep neural networks representing the functions $f_{\textnormal{Q}}$,  $f_{\textnormal{A}}$, $f_{\textnormal{\^{X}}}$, and $f_{\textnormal{\^{M}}}$ used for training for the synthetic Gaussian dataset are detailed in~\ifthenelse{\boolean{short_version}}{\cite[Tab.~IV]{WengYakimenkaLinRosnesKliewer20_1sub}.}{Table~\ref{tab:table_GaussianNNarch}.} %{\other In order to improve the training, uniformly distributed random noise is added to the output of the answer network as proposed in \cite{BlauMichaeli19_1} (see text after Eq.~(55) therein).}
% \footnote{G.~Klambauer, T.~Unterthiner, A.~Mayr, and S.~Hochreiter, ``Self-normalizing neural networks,'' In Proceedings International Conference on Neural Information Processing Systems (NeurIPS), pages 971--980, Long Beach, CA, USA, December 4–9, 2017. MIT Press.} 
In the special case of accuracy $\const{L}=1$, i.e., full leakage, the problem reduces to the classical rate-distortion problem, and to improve performance, we only considered an encoder (for compression of the requested file), corresponding to $f_{\textnormal{A}}$, and a decoder (for decompression at the user side), corresponding to $f_{\textnormal{\^{X}}}$. Both the encoder and the decoder are represented as deep neural networks and trained in the classical way~\cite{KingmaWelling14_1} %
%as outlined in [Goodfellow, Bengio, Courville: ``Deep Learning'', MIT
%Press 2017].
% \footnote{See, D.~P.~Kingma, M.~Welling, ``Auto-encoding variational Bayes,'' In Proceedings International Conference on Learning Representations (ICLR), Banff, AB, Canada, Apr.~14--16, 2014}
(due to space limitations, the actual architectures are not tabulated).

For the MNIST dataset there are $\const{M}=10$ files (there are $10$ digits) and each file is of size $\beta = 28 \times 28  =
784$ symbols, or $784 \times 8 = 6272$ bits  (each picture is of size $28 \times 28$ pixels and each pixel is of size $8$ bits). For the CIFAR-$10$ dataset, there are also $\const{M}=10$ files (there are $10$ classes of files) and each file is of size $\beta = 32 \times 32  = 1024$ symbols, or $1024 \times 8 \times 3 = 24576$ bits  (each picture is of size $32 \times 32$ pixels and each pixel is of size $8$ bits for each of the $3$ color channels). For the LSUN dataset, there are $\const{M}=10$ files and we consider two file sizes: one with images of size $32\times 32$ pixels (for the data-driven approach with image splitting), and the other one with file size $\beta = 64 \times 64 = 4096$ symbols. The architectures of the deep neural networks representing the functions $f_{\textnormal{Q}}$, $f_{\textnormal{A}}$, $f_{\textnormal{\^{X}}}$, and $f_{\textnormal{\^{M}}}$ for the MNIST dataset, the CIFAR-$10$ and LSUN datasets with images of size $32\times 32$ pixels, and the LSUN dataset with images of size $64\times 64$ pixels   are all given in~\ifthenelse{\boolean{short_version}}{\cite[Tab.~V]{WengYakimenkaLinRosnesKliewer20_1sub} and~\cite[Tab.~VI]{WengYakimenkaLinRosnesKliewer20_1sub}}{\cref{tab:img_datasets_arch}}.
% The architectures of the deep neural networks representing the functions $f_{\textnormal{Q}}$, $f_{\textnormal{A}}$, $f_{\textnormal{\^{X}}}$, and $f_{\textnormal{\^{M}}}$ in this case are given in~\ifthenelse{\boolean{short_version}}{\cite[Tab.~V]{WengYakimenkaLinRosnesKliewer20_1sub}.}{Table~\ref{tab:tableCIFAR10arch}.}
Note that in contrast to the Gaussian case, in~\ifthenelse{\boolean{short_version}}{\cite[Tab.~V and Tab.~VI]{WengYakimenkaLinRosnesKliewer20_1sub}}{\cref{tab:img_datasets_arch}} we list descriptions of three neural networks for the answer function $f_{\textnormal{A}}$, labeled as part $1$ to $3$. To produce the final answer, the output of the first network is fed to the third network together with the output from the second network and the queries (which are fed into the fourth/sixth layer of the network). The input to the first network is the queries, while the input to the second network is the set of files. The second network takes the set of files as input and extracts their features.  Note that for the LSUN dataset with $64 \times 64$ pixels images, the input to the query network is fed into the fifth layer of the decoder network, while for the other datasets it is fed into the first layer. As for the Gaussian case,  %{\other 1)} 
training for   MNIST, CIFAR-$10$, and LSUN can be simplified for $\const{L}=1$ as described above %, {\other and 2) in order to improve the training, uniformly distributed random noise is added to the output of the answer network as proposed in \cite{BlauMichaeli19_1}.} 
(again due to space limitations, the actual architectures are not tabulated).

\ifCLASSOPTIONcaptionsoff
  \newpage
\fi

\ifthenelse{\boolean{arxiv_version}}{\IEEEtriggeratref{36}}{\IEEEtriggeratref{34}}
\end{document}